\documentclass[runningheads]{llncs}
\usepackage{graphicx}
\usepackage{xspace}
\usepackage{amsmath}
\usepackage{amssymb}
\usepackage{booktabs}
\usepackage{mathtools}
\usepackage{bbm}
\usepackage{tikz}
\usepackage{comment}
\usepackage{amsmath,amssymb}
\usepackage{color}
\usepackage{graphicx}
\usepackage{amsmath}
\usepackage{amssymb}
\usepackage{booktabs}
\usepackage{mathtools}

\usepackage{subcaption}
\usepackage{bbm}
\graphicspath{{images/}{../images/}}
\usepackage[T1]{fontenc}
\usepackage{lipsum}% http://ctan.org/pkg/lipsum
\usepackage{graphicx}

\usepackage{xcolor}
\usepackage{enumitem}

\newcommand{\parsection}[1]{\vspace{0mm}\noindent\textbf{#1:}}

\usepackage[accsupp]{axessibility}  % Improves PDF readability for those with disabilities.

\usepackage[pagebackref=true,breaklinks=true,letterpaper=true,citecolor=blue,colorlinks,linkcolor=red,bookmarks=false]{hyperref}

\def\wrt{w.r.t.}

\begin{document}
\pagestyle{headings}
\mainmatter

\title{Transform your Smartphone into a DSLR
Camera: Learning the ISP in the Wild} % Replace with your title
%******************

% \begin{comment}
\titlerunning{Learning the ISP in the Wild}
\author{Ardhendu Shekhar Tripathi\inst{1} \and
Martin Danelljan\inst{1} \and
Samarth Shukla\inst{1} \and\\
Radu Timofte\inst{1, 2} \and
Luc Van Gool\inst{1, 3}}
\authorrunning{Tripathi et al.}
\institute{Computer Vision Laboratory, ETH Z\"urich, Switzerland\\
\email{\{ardhendu-shekhar.tripathi,martin.danelljan, samarth.shukla, radu.timofte, 
vangool\}@vision.ee.ethz.ch} 
\and
University of W\"urzburg, Germany
\and
KU Leuven, Belgium
}
% \end{comment}
%%%%%%%%%%%%%%%%%%%%%%%%%%%

\maketitle

\begin{abstract}
We propose a trainable Image Signal Processing (ISP) framework that produces DSLR quality images given RAW images captured by a smartphone. To address the color misalignments between training image pairs, we employ a color-conditional ISP network and optimize a novel parametric color mapping between each input RAW and reference DSLR image. During inference, we predict the target color image by designing a color prediction network with efficient Global Context Transformer modules. The latter effectively leverage global information to learn consistent color and tone mappings. We further propose a robust masked aligned loss to identify and discard regions with inaccurate motion estimation during training. Lastly, we introduce the ISP in the Wild (ISPW) dataset, consisting of weakly paired phone RAW and DSLR sRGB images. We extensively evaluate our method, setting a new state-of-the-art on two datasets.
% \keywords{Color Conditional ISP, Efficient Global Attention, Learning in the Wild, Mate30Canon Dataset}
\end{abstract}

\begin{figure}[t]
  \centering
  \subfloat[Phone RAW\label{fig:intro_1}]{\includegraphics[width=.243\linewidth]{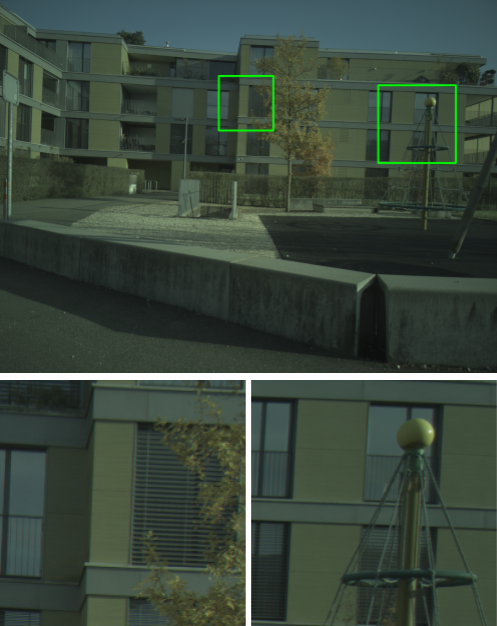}}\hfill%
\subfloat[LiteISP-Net~\cite{liteispnet}.\label{fig:intro_2}]{\includegraphics[width=.243\linewidth]{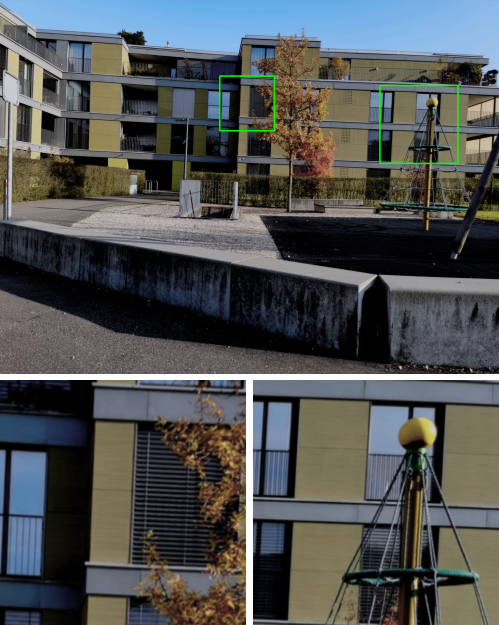}}\hfill
\subfloat[Ours\label{fig:intro_3}]{\includegraphics[width=.243\linewidth]{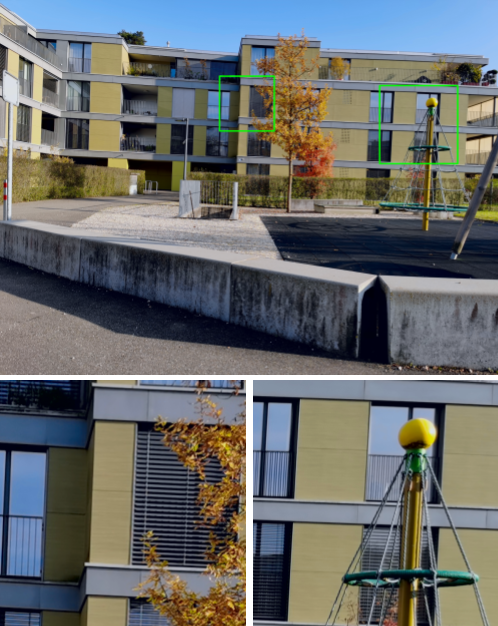}}\hfill
\subfloat[DSLR sRGB\label{fig:PerceiverBlock}]{\includegraphics[width=.243\linewidth]{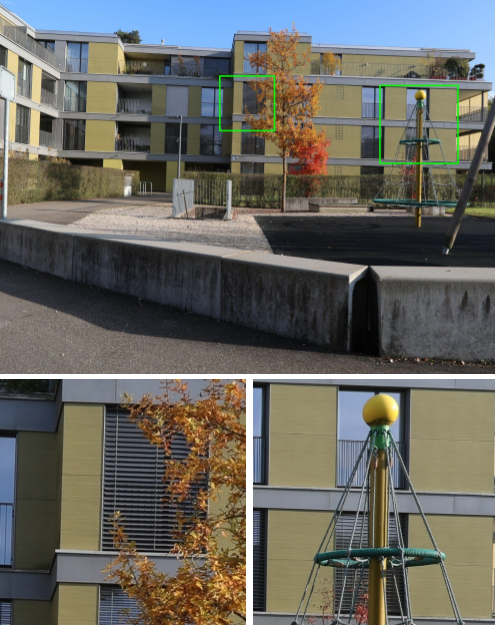}}\vspace{-2mm}
\caption{Our learnable ISP generates a DSLR quality sRGB image from RAW data captured by a smartphone camera. Our approach recovers rich details and produces colors that are more consistent with the DSLR sRGB ground-truth, compared to LiteISPNet (best performing competing method). Shown are the full resolution results on our ISP in the Wild (ISPW) dataset. Best viewed with zoom.}\vspace{-6mm}
\label{fig:intro}
\end{figure}
\vspace{-7mm}
\section{Introduction}
\label{sec:intro}

An Image Signal Processing (ISP) pipeline is characterized by a sequence of low-level vision operations that are performed to convert RAW data from the camera sensor to sRGB images. Each camera has an inherent ISP that is implemented on the device through hand-designed operations. With the advent of mobile photography, smartphones have become the primary source of photo capture due to their portability. However, their strict size constraints enforces small sensor sizes and compact lenses, which inevitably leads to higher sensor noise compared to DSLR cameras. In this work, we therefore strive towards mitigating the hardware constraints in mobile photography by designing a learnable alternative to the ISP pipeline, utilizing DSLR quality sRGB images as reference. 

Compared to standard image enhancement and restoration tasks, learning the ISP mapping introduces new fundamental challenges, which require careful attention. In the paired learning setting, a primary issue is that the color mapping between the input RAW image and the DSLR sRGB image depends on partially unobserved factors, such as camera parameters and the environmental conditions. Further, the image pairs for training, each consisting of a smartphone RAW and a DSLR sRGB image, inevitably contain substantial spatial misalignment that greatly complicate the learning. Despite recent efforts~\cite{ZRR,awnet,liteispnet}, the aforementioned issues remain central in the strive towards a fully learning-based ISP solution.

% In this work, we propose a learnable ISP framework via a soft attention inspired color mapping.

In this work, we propose a learnable ISP framework that can be effectively trained \emph{in the wild}, using only weakly paired DSLR reference images with unknown and varying color and spatial misalignments.
Our approach is composed of an ISP network that maps the input phone RAW to a DSLR quality output. Contrary to much previous works, we further condition the network on a target color image. This allows our ISP network to fully focus on the denoising and demosaicing tasks, without having to guess the unknown color transformation. To allow the target color image to be used during training, we propose a flexible and efficient parametric color mapping. Our color mapping between the input RAW and output DSLR sRGB image is individually optimized for every training image pair. The resulting mapping is then applied to the input RAW image to generate the target color image for conditioning. Importantly, this approach effectively mitigates information leakage from the target ground truth into the network, while achieving a faithful color transformation.

In order to achieve the target color image during inference, we further propose a dedicated target DSLR color prediction network, which solely takes the RAW phone image as input. To predict an accurate target color image, exploiting both local and global cues in an image is essential. While local information capture high-frequency details, global information is important in order to achieve a globally consistent and realistic color mapping across the entire image. We achieve the latter by designing an efficient Global Context Transformer block, which aggregates global color information into a compact latent array through cross-attention operations. This both alleviates the quadratic complexity of standard transformer modules, and importantly enables a variable input size. Finally, we address the problem of misaligned ground-truth by introducing a robust masked aligned objective for training our ISP framework.

To aid in extensive benchmarking and evaluation of RAW-to-sRGB mapping approaches for weakly paired data, we introduce the ISP in the Wild (ISPW) dataset. This dataset comprises of pairs of RAW sensor data from a recent smartphone camera and sRGB images taken from a high-end DSLR camera. Our dataset consists of 200 captured 10+ MegaPixel image pairs, resulting in over 28,000 crops of size $320\times 320$ for training, validation, and test. We perform extensive ablative and state-of-the-art experiments on the Zurich RAW-to-RGB (ZRR) dataset~\cite{ZRR} and our ISPW dataset. Our approach outperforms all previous approaches by a significant margin, setting a new state-of-the-art on both datasets. A visual comparison with the best competing method is provided in Fig.~\ref{fig:intro}.

\noindent\textbf{Contributions:} 
Our main contributions are summarized as: 
\textbf{(i)} We propose a color conditional trainable ISP in the wild.
\textbf{(ii)} We propose a color prediction network that integrates a global-context transformer module for efficient and globally coherent prediction of the target colors.
\textbf{(iii)} We condition on color information from the reference image during training by introducing a flexible parametric color mapping, which is efficiently optimized for a single RAW-sRGB training pair.
\textbf{(iv)} We employ a loss masking strategy for robust learning under alignment errors.
\textbf{(v)} We introduce the ISPW dataset for learning the camera ISP in the wild.
\vspace{-1.5mm}
\section{Related Work}

% Learnable ISP/raw2rgb (focus on learning from misaligned data and colors): 
% - all methods doing this from challenges etc
% - deep burst SR 
% - zoom to learn

% Datasets for ISP:
% - all related datasets
% - what's new in our dataset

% Color mapping works:
% - deep white balancing (look at other afifi's works)
% - other methods??

%  cycle isp, invertible isp (these use same device mapping). Also https://ieeexplore.ieee.org/document/9394803 - cie xyz unprocess net thing from afifi

%-------------------------------------
Despite the successes of deep-learning for low-level vision tasks, its application to camera ISP in the wild has been much less explored. Among the existing methods, CycleISP~\cite{cycleisp} and Invertible-ISP~\cite{invertibleisp} propose a full camera imaging pipeline in the forward and reverse directions. These methods learn the ISP in a well aligned setting, where the RAW-sRGB training pairs originate from the same device. For RAW-to-sRGB mapping in the wild, the goal of the AIM 2020 challenge~\cite{ZRR} on learned image processing pipeline was to map the original low-quality RAW images captured by a phone to a DSLR sRGB image. In particular, the CNN approaches inspired by the Multi-level Wavelet CNNs (MWCNN)~\cite{mwcnn} obtained the best results. Among the MWCNN-based methods both, MW-ISPNet~\cite{ZRR} and AWNet~\cite{awnet} employ different variations of a U-Net for generation of appealing sRGB images. 
% They alleviate the effect of misalignment in training pairs by employing a VGG perceptual~\cite{vggloss} loss term.
% Among the MWCNN-based methods, MW-ISPNet~\cite{ZRR} propose a U-Net based architecture taking advantage of residual channel attention blocks~\cite{rcan}. Further, AWNet~\cite{awnet} also incorporates a global context block to learn the non-local color mapping for the generation of appealing RGB images. Both, MW-ISPNet and AWNet alleviate the effect of misalignment by employing a VGG perceptual~\cite{vggloss} loss term.

More recently, LiteISPNet~\cite{liteispnet} propose an aligned loss by explicitly calculating the optical flow between the predicted DSLR image and the ground truth. The idea of the aligned loss using optical flow in case of misaligned data was first used in DeepBurstSR~\cite{deepburstsr} for burst super-resolution. Prior to DeepBurstSR, other efforts to handle misaligned data include a contextual bilateral loss (CoBi)~\cite{zoomlearn} or primarily relying on a deep perceptual loss function, as in MW-ISPNet~\cite{ZRR} and AWNet~\cite{awnet}.

Another bottleneck for the field has been the dearth of datasets for camera ISP learning and benchmarking. The datasets MIT5K~\cite{fivek}, DND~\cite{dnd}, SIDD~\cite{SIDD} and Zoom-to-Learn~\cite{zoomlearn} capture several images from the same device under different settings. Moreover,~\cite{fivek,dnd,SIDD} collect images in very controlled settings, where accurate alignment is possible. They are therefore unfit for designing approaches for ISP in the wild. Further, DPED~\cite{dped} provides RGB images from different devices but does not contain RAW images and thus cannot be used for our task of designing and training the full ISP pipeline. In contrast, we aim to learn the ISP from a constrained device, i.e.\  smartphone, using high-quality DSLR images. The BurstSR dataset~\cite{deepburstsr} is designed for the burst super-resolution task. Most related is the ZRR dataset~\cite{ZRR}. Our ISPW dataset contains RAW images collected via a more modern smartphone. Additionally, our ISPW dataset contains important meta information, such as the ISO and exposure settings, that can further be exploited by the community for controllable and conditional learning of the RAW-to-sRGB mapping for weakly paired data.

\section{Method}
\begin{figure*}[t]
\centering
\includegraphics[width=0.9\textwidth]{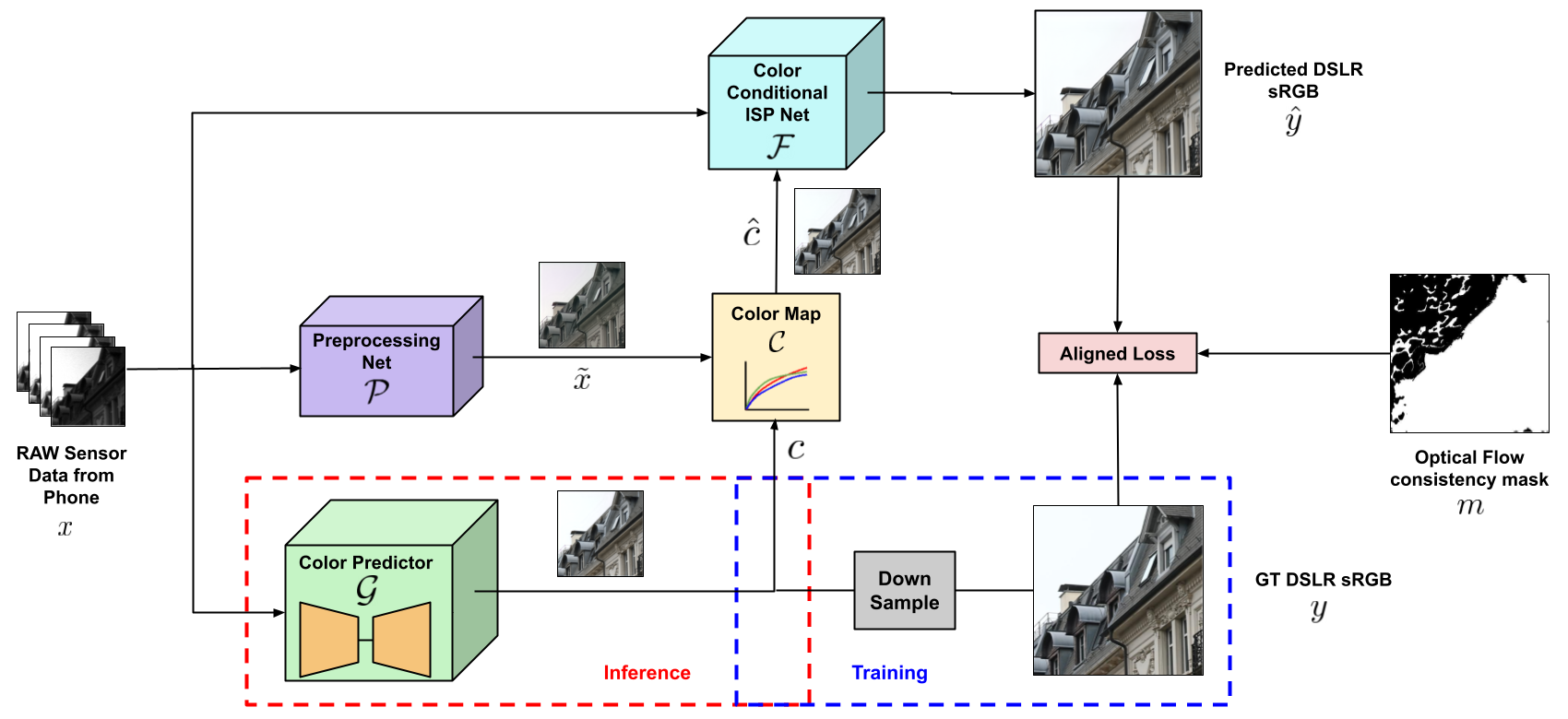}
\caption{An Overview of our learnable ISP framework: We learn a color conditional framework $\mathcal{F}(x, \hat{c})$ for RAW-to-sRGB mapping in the wild (Sec.~\ref{sec:restore}). The estimated target color image $\hat{c}$ is achieved by our color mapping $\hat{c} = \mathcal{C}(x, c)$ (Sec.~\ref{sec:color_map}), which maps the raw input $x$ to the color space of $c$. During training $c$ is given by the downsampled ground truth. During inference, the DSLR-quality color content is predicted by the dedicated global attention based color prediction network $\mathcal{G}(x)$, using only the raw image $x$ as input (Sec.~\ref{sec:colorinference}). 
Finally, for robust learning of the ISP in the presence of even substantial misalignments (see Fig.~\ref{fig:intro}), we propose a masked aligned loss (Sec. \ref{sec:loss}), which is robust to errors in the computed optical flow.}
\label{fig:overview}
\end{figure*}

In this work, we strive towards a fully deep learning based ISP module, which predicts a high-quality sRGB image $y\in\mathbb{R}^{3\times H\times W}$ given the RAW image $x\in\mathbb{R}^{4\times \frac{H}{2}\times \frac{W}{2}}$ captured by a mobile phone camera. Specifically, our aim is to learn such a module from a set of weakly paired training samples $\{(x^k, y^k)\}_k$. Our approach is illustrated in Fig.~\ref{fig:overview}. It is comprised of a color conditional restoration network $\mathcal{F}(x, \hat{c})$ (Sec.~\ref{sec:restore}). The color information $\hat{c}$ is provided by a dedicated color prediction network $\mathcal{G}(x)$ during inference (Sec.~\ref{sec:colorinference}) and by the ground truth DSLR sRGB during training. To avoid the network from cheating during training, we propose a color mapping approach (Sec.~\ref{sec:color_map}) that maps the RAW sensor data to the target DSLR sRGB. During inference, our color mapping module works as a regularizer for our color predictor network in case of spurious inaccurate local colors predicted. Further, there also exists a spatial misalignment between the noisy mobile sensor data and the target DSLR sRGB image. To handle misalignment between the RAW-sRGB pairs, we propose a robust masked aligned loss (Sec.~\ref{sec:loss}) that also takes into account the inaccuracies that are introduced during the alignment operation.  

\subsection{ISP Network}
\label{sec:restore}
As motivated in Sec.~\ref{sec:intro}, there exists an unknown color mapping between the input $x^k$ and the target $y^k$, which further varies between each capture $(x^k, y^k)$ due to changes in the parameters and environment. Modelling the ISP pipeline in the wild as a single feed-forward network $y=\mathcal{F}(x)$ can therefore prove detrimental to the learning of an accurate RAW-to-sRGB mapping as no fixed global color mapping exists. In order to learn effectively the RAW-to-sRGB mapping in these conditions, we propose a network $y=\mathcal{F}(x, \hat{c})$ that is conditioned on the desired output color information $\hat{c}$. During training, the color information is extracted from the RAW-sRGB pair using a flexible parametric formulation, which is detailed in Sec.~\ref{sec:color_map}. This allows us to capture a rich color mapping model from a single training pair $(x^k, y^k)$, while preventing the network $\mathcal{F}$ to cheat. Additionally, our dedicated RAW pre-processing network discussed in Sec.~\ref{sec:preprocess_net} mitigates the ill-effects that noise in the RAW sensor data has on our color mapping estimation module. During inference, the color information $\hat{c}$ is predicted by a dedicated color predictor network $\mathcal{G}(x)$  (Sec.~\ref{sec:colorinference}) and the color mapping module (Sec.~\ref{sec:color_map}). 

\subsection{Color Prediction}
\label{sec:colorinference}
% - Motivate what we want to achieve
% - ...
In this section, we propose a low-resolution reference color prediction network $c = \mathcal{G}(x)$. This network aims to predict a low-resolution image $c$ with the color content and dynamic range of the target DSLR camera. It is then the task of our ISP network $\mathcal{F}$, to predict a detailed high-resolution image, conditioned on this color information. The measured colors and intensities depend on the camera parameters during capture, along with various other environmental factors, such as the properties of the illuminants in the scene. These conditions vary on a capture to capture basis. Hence, a simple feedforward network fails to capture the DSLR sRGB color accurately.

\begin{figure}[t]
  \centering%
  \subfloat[Architecture overview.\label{fig:encode_decode}]{\includegraphics[width=.47\linewidth]{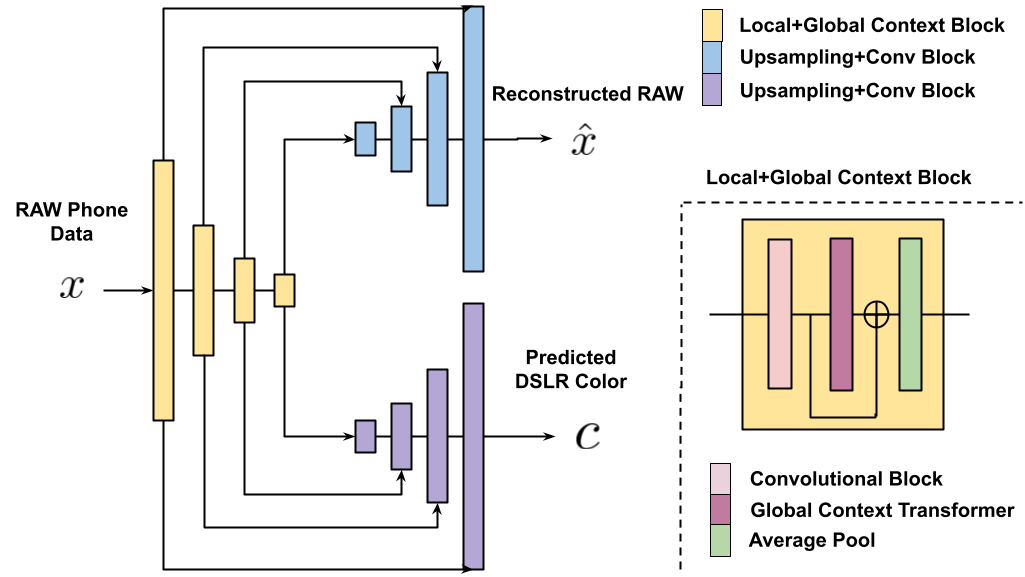}}\hspace{.05\linewidth}%
\subfloat[Global Context Transformer.\label{fig:globalcontext}]{\includegraphics[width=.47\linewidth]{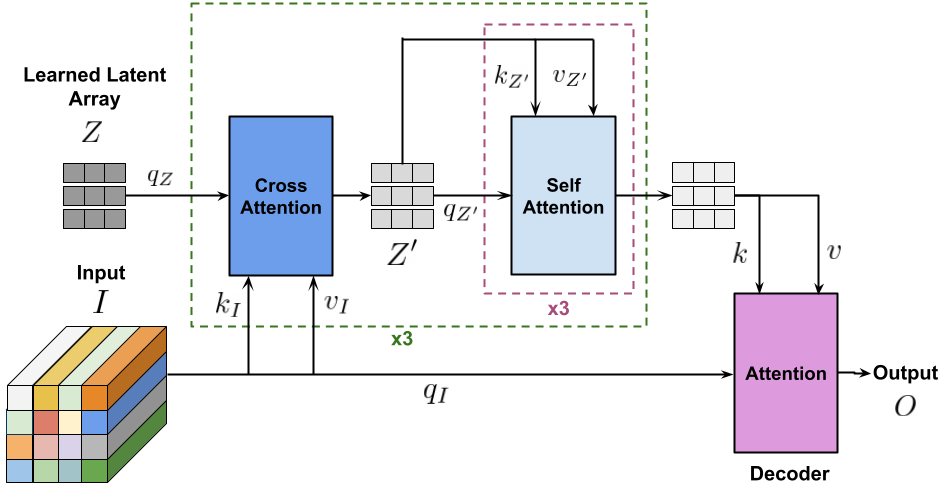}}
\caption{Illustration of the full Color Prediction Network (a) with its Global Context Transformer module (b).}
\label{fig:PerceiverBlock}
\end{figure}

\parsection{Color prediction network}
To circumvent this drawback of feed-forward nets, we design an encoder-decoder based color prediction network (Fig.~\ref{fig:encode_decode}). 
\begin{equation}
\label{eq:colorpred}
    c = \mathcal{G}(x) = D_\text{DSLR}\big(E_\text{phone}(x)\big).
\end{equation}
Here, $D_\text{DSLR}$ is the DSLR decoding network that predicts a low resolution target sRGB color. Predicting the target sRGB colors in low resolution makes the learning easier and leads to a faster convergence. We employ a U-Net inspired architecture (Fig.~\ref{fig:encode_decode}) for our encoder-decoder. This is because U-Net~\cite{U-net} effectively expands the receptive field by integrating pooling operations and exploiting contextual information at different scales using skip connections. Further, a U-Net is relatively insensitive to small misalignments in the image due to the low-resolution of core features, achieved by successive pooling operations. Our U-Net encoder $E_\text{phone}$ exploits local and global cues by integrating a successive convolutional layer and an efficient global context transformer. 

\parsection{Global Context Transformer}
\label{sec:transformer}
For target color prediction, capturing a global context is pivotal since color in one patch of the image can be related to the color in a spatially distant patch of the same image. Hence, attending to different patches in the image may prove beneficial for predicting an accurate target color. Using standard transformers~\cite{attentionisallyouneed} for global attention is a viable option. However, its quadratic computational complexity w.r.t.\ the number of patches in the image/feature map makes it unsuitable for our color prediction network. Furthermore, our network needs to be able to process an image of arbitrary resolution, which brings further challenges to a standard transformer architecture.

We therefore design our Global Context Transformer block by taking inspiration from the Perceiver~\cite{Perceiver,PerceiverIO} architecture. Specifically, we perform cross attention operations between an auxiliary latent space $Z' \in \mathbb{R}^{K\times C}$ and the input feature map $I_l \in \mathbb{H}_l\times \mathbb{W}_l \times \mathbb{D}_l$, followed by self attention layers on $Z'$. Here, $I_l$ is extracted from the U-Net encoder at level $l$. The latent space contains $K$ tokens of dimension $C$, as is initialized by a learned constant array $Z \in \mathbb{R}^{K\times C}$. The majority of the computation thus happens on $Z'$. This reduces the complexity of the attention operations from quadratic to linear in the input size, and crucially enables a variable input image size. 

Fig.~\ref{fig:globalcontext} details the architecture of our global context block. It comprises multiple cross and self-attention layers on the fixed-size auxiliary latent array $Z'$. Hence, decoupling the network depth from the input size. 
Through the global attention operations, the learned latent arrays $Z'$ can encode color transformations. The final decoder module then maps information encapsulated in $Z'$ to the output array $O$ through cross attention with the input query $q_I$. 
We integrate our Global Context Transformer block in the contracting path of our color prediction module after each convolutional block (Fig.~\ref{fig:encode_decode}). This aids in exploiting local cues (convolutional block) as well as global cues (Global-context transformer block) while remaining computationally efficient.

\parsection{Reconstruction branch}
\label{sec:reconstruct}
In addition to the DSLR specific decoder, we also employ a decoder $D_\text{phone}$ for reconstructing the RAW input $x$ such that $\hat{x} = D_\text{phone}(E_\text{phone}(x))$ (Fig.~\ref{fig:encode_decode}). Employing a RAW reconstruction decoder equips our color prediction framework to learn an optimal phone-specific embedding $E_{\text{phone}}(x)$ that encodes various meta-information that was not provided with the RAW data for reconstructing the RAW input $x$. Hence, intuitively our DSLR-specific decoder learns a mapping from the phone ISP to the DSLR ISP.

\subsection{Color Mapping Module}
\label{sec:color_map}
In this section we introduce our approach for estimating the color transformation between the RAW input $x$ and a target color sRGB image $c$. For this, we design a module $\hat{c} = \mathcal{C}(x,c)$ that estimates a color mapping between a single pair $(x, c)$, and applies it to $x$. The result represents the RAW image $x$ transformed according to the target color space in $c$.  Our approach is particularly important during training, when $c$ is derived from the ground-truth image $y$ through downsampling and alignment. It supplies our ISP network, conditioned on $\hat{c}$, with the correct color transformation between the pair $(x, y)$ while preventing information leakage from the ground-truth $y$. During inference, $\mathcal{C}$ works as a regularizer for our color predictor network \eqref{eq:colorpred} in case of spurious inaccurate local colors predicted. 

\parsection{Pre-processing network}
\label{sec:preprocess_net}
Real world training image pairs, apart from being weakly paired in terms of alignment, pose many other challenges. In particular, the RAW sensor data from the phone is prone to noise due to the limited sensor size, along with other interference from the environment. The noise may be signal-dependent or signal-independent. A noisy source image $x$ inhibits the performance of the color mapping significantly. Hence, removing noise from the RAW data is pivotal. In this direction, we design a pre-processing module for removing noise from the RAW data, thereby aiding our color mapping module. 

Our RAW pre-processing network $\mathcal{P}$ aims to retrieve the clean source image $\Tilde{x}$ given a noisy RAW $x$,
\begin{equation}
    \mathcal{P}(x) \coloneqq \Tilde{x} = x'-\eta(x'),\text{ where } x'=\Gamma(x).
    \label{eq:tilda_x}
\end{equation}
% The noise estimation net $\eta$ consists of 3 Residual-in-Residual Dense Blocks (RRDB) ~\cite{RRDB} followed by a $3\times3$ convolutional layer. 
Here, $\eta$ is our noise estimation net and is implemented as a CNN with residual connections. For our framework, $x'$ is a processed version of the mobile RAW sensor data $x$. We obtain $x'$ by neglecting one of the green channels in $x$ and normalizing the resulting 3-channel image between $[0, 1]$ uniformly. To further reduce the non-linearities in the color mapping, we apply a constant approximate gamma correction to obtain the final processed image $x'$. The processing operation $\Gamma(\cdot)$ is detailed in the Appendix (Sec. ~\ref{sec:preproc_net_supp}).

\parsection{Color mapping}
Formulating our color mapping scheme, we define a set of $\mathcal{B}$ equally spaced bins between the range of values in each channel of the source image $\Tilde{x}$ (Eq.~\ref{eq:tilda_x}). The $b^{th}$ bin centroid for color channel $j$ is denoted as $k_b^j$. The goal is to map the image $\Tilde{x}$ to the target color image as,
\begin{equation}
\label{eq:inference_cmap}
\hat{c}^j_i = \sum_{b=1}^\mathcal{B} \hat{w}^j_{ib}(A^j_b \Tilde{x}_i + B^j_b),  
\end{equation}
using a learned affine transformation $A^j_b\Tilde{x}_i+B^j_b$ for each bin $b$. Here, $A^j_b\in\mathbb{R}^{1\times 3}$ and $B^j_b\in\mathbb{R}$ are the parameters of the affine map, while $\Tilde{x}_i\in\mathbb{R}^{3}$ (Eq.~\ref{eq:tilda_x}) denotes the color values at pixel $i$ after the pre-processing network. The result $\hat{c}^j_i$ is the mapped intensity at channel $j$ and location $i$. 
The soft bin assignment weights in~\eqref{eq:inference_cmap} are calculated as
$\hat{w}^j_{ib}=\underset{b}{\text{SoftMax}}(-\|\Tilde{x}^j_i-k^j_b\|^2/T)$, where, $T$ is a temperature parameter. Hence, our color mapping~\eqref{eq:inference_cmap} can be seen as an attention mechanism, with the source image attending to the learned values through the bin centroids. The motivation of learning an affine transformation instead of a fixed numeric value for each bin centroid is providing each bin more expressive power leading to better color mapping even with less number of bins. 

In~\eqref{eq:inference_cmap}, the parameters $(A^j_b , B^j_b)$ of the affine mapping are learned using only a single pair $(\Tilde{x},c)$. This is performed by minimizing the following squared error to the target color value $c^j_i$,
\begin{equation}
\label{eq:min_problem}
    A_b^j, B_b^j = \underset{A, B}{\text{argmin}}\sum_i w^j_{ib}\|A\Tilde{x}_i + B-c^j_i\|_2^2 \,. 
\end{equation} 
Here, the weights $w^j_{ib}$ are calculated as $w^j_{ib}=\underset{i}{\text{SoftMax}}(-\|\Tilde{x}^j_i-k^j_b\|^2/T)$. These set of weights signify how much each target intensity affects the affine transformation learned for each bin centroid. The objective ~\eqref{eq:min_problem} corresponds to a linear least squares problem, which can efficiently be solved in closed form as detailed in the Appendix (Sec. ~\ref{sec:closed_form_supp}).

\subsection{Learning the Camera ISP}
\label{sec:loss}
The RAW-sRGB pairs taken from two different devices are misaligned. The reasons are the different fields of view for both the cameras, parallax, and small motion of objects in the scene. Misalignment in the RAW-sRGB pair makes  training the ISP pipeline difficult. Trying to learn in such a setting produces blurry results and significant color shift (Fig.~\ref{fig:sota}). Hence, a robust loss applicable to the weakly paired setting is pivotal. In this section, we introduce an aligned masked loss for robust learning in a weakly paired setting. We then introduce the objectives for our main ISP network, pre-processing network, and the color prediction network. Lastly, we provide training strategies and details.

\parsection{Alignment}
We calculate aligned losses for learning our color conditional RAW-to-sRGB network in the wild. For alignment, we use the PWC-net~\cite{pwcnet} for computing optical flow. We denote by $c_{x'} = \mathcal{W}(c,f(c, x'))$ the color image $c$ aligned with respect to the processed RAW $x'$ (Sec.~\ref{sec:color_map}). Here, $f(c,x')$ is the optical flow from the color image $c$ to the  processed RAW $x'$. While we found PWC-Net to be robust to substantial color transformations between the input images, we use the processed RAW $x'$ as input as it has a much smaller difference in color and intensity to the reference color image $c$. Further, the loss masking discussed next aids in a more robust loss calculation for inaccurately aligned regions.

\parsection{Loss masking}
Although, employing an aligned $L_1$-loss partially handles the misalignment problem for ISP learning in the wild, the flow estimation itself can introduce errors. In particular, optical flow is often inaccurate in the presence of repeating patterns, occlusions, and homogeneous regions. This leads to an incorrect training signal which degrades the quality of the ISP network. We therefore propose a mask for our loss by identifying regions where the optical flow is inaccurate. Inspired by~\cite{unflow}, we use the forward-backward consistency constraint to filter out regions with inaccurate flow. The optical-flow consistency mask $m$ is set to 1 where the following condition holds true, and otherwise to 0:
\begin{align}
\label{eq:mask}
\begin{split}
\big|f(x', y^\downarrow)+f(x'_{y^\downarrow}, x')\big|^2<\alpha_1\Big(\big|f(x', y^\downarrow)\big|^2 + \big|f(x'_{y^\downarrow}, x')\big|^2\Big) + \alpha_2.
\end{split}
\end{align}
Here, $x'$ is the processed RAW sensor data (Sec.~\ref{sec:preprocess_net}). And, $y^\downarrow$ is the target sRGB image bilinearly downsampled by a factor of 2. And, $x'_{y^\downarrow}$ is $x'$ aligned with ${y^\downarrow}$. Thus, the mask $m$ aids in masking out inaccurately aligned regions.

\parsection{ISP Network Loss}
The masked target sRGB prediction loss is given by:
\begin{gather}
    \hat{y} = \mathcal{F}(x, \hat{c})\text{, where }\hat{c}=\mathcal{C}(\Tilde{x}, c_{\Tilde{x}}) \nonumber\\
    L_\text{pred}(\hat{y}, y)=\|m^\uparrow\odot\big(y_{\hat{y}} - \hat{y}\big)\|_1.
    \label{eq:pred_mask}
\end{gather}
Here, $y_{\hat{y}}$ is the target DSLR sRGB aligned w.r.t.\ the final predicted sRGB $\hat{y}$. We did not see a significant difference in performance
when we align the predicted sRGB $\hat{y}$ w.r.t.\ the target DSLR sRGB for our loss calculation (Sec. ~\ref{sec:ablation_supp} of the Appendix).  Our choice of alignment direction circumvents the need of differentiating through the warping process. During training, the color image $c=y^\downarrow$ is the $2\times$ downsampled ground truth sRGB. Further, $c_{\Tilde{x}}$ is the color image $c$ aligned with $\Tilde{x}$ (Eq.~\ref{eq:tilda_x}).
Lastly, $m^\uparrow$ is the 2$\times$ upsampled mask $m$ via nearest neighbour interpolation. 

\parsection{Pre-processing Network Loss}
The pre-processing net (Sec.~\ref{sec:preprocess_net}) aims at providing a source image that aids our learned parametric color mapping scheme (Sec.~\ref{sec:color_map}) and denoising the processed RAW $x'$ (Sec.~\ref{sec:preprocess_net}). Motivated by this, we design loss for our pre-processing net $\mathcal{P}$ as,
\begin{gather}
    L_\text{map}(\mathcal{C}(\Tilde{x}, c_{x'}), c_{x'})=  \|m\odot(\mathcal{C}(\Tilde{x}, c_{x'}) - c_{x'})\|_1\text{, and} \nonumber\\
    L_\text{constraint}(x', \Tilde{x})=\|b*x' - b*\Tilde{x}\|_1.
\label{eq:preprocess_loss}
\end{gather}
Here, $\Tilde{x}$ is the output of our Pre-processing Net (Eq.~\ref{eq:tilda_x}) and $b$ is a predefined blurring kernel. The loss $L_\text{constraint}$ constrains $\mathcal{P}$ to keep the color of $x'$. The color image $c=y^\downarrow$ is the $2\times$ downsampled ground truth sRGB. And, $c_{x'}$ is the color image $c$ aligned with $x'$. These set of losses aid the pre-processing network in not only denoising the RAW sensor data but also allows for the network to be flexible enough to learn a color space where the color mapping (Sec.~\ref{sec:color_map}) is optimal. 

\parsection{Color Prediction Network Loss} 
To train our target color prediction network (Sec.~\ref{sec:colorinference}), we employ a color prediction loss on the predicted low resolution target color image $\hat{y}^\text{clr}=\mathcal{G}(x)$ and a reconstruction loss on the reconstructed RAW sensor data $\hat{x}$,
\begin{gather}
    L^\text{clr}_\text{pred}(\hat{y}^\text{clr}, c_{x'})=\Big\|m\odot\big(\hat{y}^\text{clr} - c_{x'}\big)\Big\|_1  \nonumber\\
    L_\text{reconstruct}(\hat{x}, x)=\|x - \hat{x}\|_1
\label{eq:clr_pred_loss}
\end{gather}
Here, $c_{x'}=y^\downarrow_{x'}$ is the $2\times$ downsampled ground truth sRGB aligned with $x'$. Hence, $c_{x'}$ serves as the target color image for training our color prediction network in the loss $L^\text{clr}_\text{pred}$. The reconstruction loss $L_\text{reconstruct}$ further encourages the encoder $E_\text{phone}(x)$ to preserve important image details. 

\parsection{Training}
Thanks to the independent objectives, we can train our color conditional ISP network $\mathcal{F}$ and the color prediction network $\mathcal{G}$ separately. This allows use of larger batch sizes and reduced training times significantly. A comparative study with the joint fine-tuning of both the networks is provided in Sec. ~\ref{sec:ablation_supp} of the Appendix. The final training loss for $\mathcal{F}$ is given by~\eqref{eq:pred_mask} and~\eqref{eq:preprocess_loss}. The loss for the color prediction net $\mathcal{G}$ is given by~\eqref{eq:clr_pred_loss}. Each batch for training both, $\mathcal{F}$ and $\mathcal{G}$ comprises 16 pairs of randomly sampled RAW phone images $x\in\mathbb{R}^{4\times 80 \times 80}$ and DSLR sRGB images $y\in\mathbb{R}^{3\times160\times160}$. During training, we augment the data by applying random flips and $90\deg$ rotations. 
To increase the robustness of our color conditional ISP network $\mathcal{F}$, we employ color augmentations on the ground truth DSLR sRGB during training. Specifically, we randomly jitter the hue, saturation, brightness and contrast in a range $[-0.2,0.2]$.

The blurring kernel $b$ in~\eqref{eq:preprocess_loss} is a $9\times 9$ Gaussian with the standard deviation in each of the dimension set to $2$. The constants $\alpha_1$ and $\alpha_2$ for computing $m$ are set to $0.01$ and $0.5$, respectively. The number of bins $\mathcal{B}$ in our color mapping~\ref{sec:color_map} is set to $15$ and the temperature parameter $T=(1/\mathcal{B})^2$. Finally, to handle vignetting (dark corners) that occurs in RAW sensor data, we append the RAW data with a pixel-wise function of 2D coordinate map for the inputs to our pre-processing net $\mathcal{P}$ and the color prediction net $\mathcal{G}$. We use the ADAM algorithm~\cite{adam} as optimizer with $\beta_1=0.9$ and $\beta_2=0.99$. The initial learning rate for training both our networks is set to $2e-4$ which is halved at $50\%$, $75\%$, $90\%$ and $95\%$ of the total number of epochs respectively. The networks are trained separately for $100$ epochs on a Nvidia V100 GPU. The training time for our $\mathcal{F}$ and $\mathcal{G}$ nets was $27$ hours and $22$ hours, respectively. Other implementation and architecture details are provided in the Appendix. The code would be released upon publication.
\section{Dataset}
\label{sec:dataset}
% To train and validate our Color Conditional RAW-to-sRGB network in the real world, we require a diverse, high quality dataset. To the best of our knowledge, the ZRR dataset \cite{} is the only dataset that exists for RAW-to-sRGB mapping in the wild. 
We propose the ISP in the Wild (ISPW) dataset for learning the camera ISP in the wild. The ISPW dataset consists of a set of 200 high-resolution captures from a Canon 5D Mark IV DSLR camera (with a lens of focal length 24mm) and a Huawei Mate 30 Pro mobile phone. Each capture comprises of the RAW sensor data from the mobile phone (4$\times$1368$\times$1824) and 3 sRGB DSLR images (3$\times$4480$\times$6720) of the same scene taken at different exposure settings (EV values: -1, 0 and 1). All DSLR images were captured with an ISO of 100 for more detail and less noise. Further a small aperture of F18 was used for a large depth of field. The dataset was collected over several weeks in a variety of places and in various illumination and weather conditions to ensure diversity of samples. During the capture, both the devices were mounted on a tripod using a custom made rig to ensure no blur due to camera motion. Collection was focused on predominately static scenes in order to ease the alignment between the two cameras. However, small motion is inevitable in most settings, and thus need to be handled by our data processing and robust learning objectives. We split the ISPW dataset into 160, 20, and 20 high-resolution captures for training, validation, and test, respectively. Our ISPW dataset will be released upon publication. We believe that it can serve as an important benchmarking and training set for RAW-to-sRGB mapping in the wild.

\parsection{Data processing}: 
We describe the pre-processing pipeline for our ISPW data here. We consider the DSLR image taken at EV value 0 as the target DSLR sRGB in this work. We first crop out the matching field of view from the phone and the DSLR high-resolution captures using SIFT~\cite{sift} and RANSAC~\cite{ransac}. Crops of size 320$\times$320 are then extracted in a sliding manner (stride of 160) from both, the DSLR sRGB and the phone sRGB (obtained using the phone ISP). Local alignment is performed by estimating the homography between two crops. The corresponding 4-channel RAW crop from the phone of size $160\times160$ is extracted using the coordinates of the $320\times320$ phone sRGB crop and paired with the DSLR sRGB crop. In order to filter out crops with extreme scene mismatch, we discard the RAW-sRGB pairs which have a normalized cross correlation of less than 0.5 between them.

%(this threshold is 0.9 for the ZRR dataset ~\cite{ZRR}). 
\vspace{-1mm}
\section{Experiments}

Here, we perform extensive experiments to validate our approach for RAW-to-sRGB mapping in the wild. We evaluate our approach on the test sets of the ZRR  dataset~\cite{ZRR} and our ISPW dataset (Sec.~\ref{sec:dataset}). 
The methods are compared in terms of the widely used PSNR and SSIM~\cite{ssim} metrics. For a fair comparison, we align the ground truth DSLR sRGB with the phone RAW for the computation of PSNR and SSIM metrics. For additional qualitative results and analysis, refer to the Appendix (Sec. ~\ref{sec:ablation_supp}). 

\subsection{Ablative Analysis of the Color Mapping}
\label{sec:ablation_color_map}
In this section, we study the effectiveness of our color mapping scheme (Sec.~\ref{sec:color_map}) compared to other alternatives. The results on the ZRR dataset are reported in Tab.~\ref{tab:color_map}.
% like camera parameters and external environmental conditions

\textbf{NoColorPred:} As a baseline for evaluating our color mapping scheme, we train $\mathcal{F}(x, \hat{c})$ with the color information $\hat{c}$ set to $0$. This implies a simple feed-forward network setting. We do not include the color mapping module $\mathcal{C}$ in this version. NoColorPred achieves a PSNR of 21.27 dB and a SSIM of 0.844. This variation learns average average and dull colors and is not able to account for various factors on which the color in an image depends. \textbf{ColorBlur:} Next, as in CycleISP~\cite{cycleisp}, we train $\mathcal{F}(x,\hat{c})$ where the target color $\hat{c}=z*y^\downarrow_{x'}$ is achieved by blurring the 2x downsampled target DSLR sRGB (aligned with $x'$) with a Gaussian kernel $z$ during training. At inference, we apply the same blurring to our predicted target color $\hat{c}=z*\mathcal{G}(x)$. As in NoColorPred, we do not include the color mapping module $\mathcal{C}$ in this version. ColorBlur achieves a gain of 2.16 dB in PSNR over NoColorPred. Although being better than NoColorPred, ColorBlur fails to capture the sudden changes of color in the image contour. 

% (this is different from $b$ in Eq. ~\ref{eq:preprocess_loss})

% Thus, showing that conditioning $\mathcal{F}$ on $c$ is beneficial for learning an accurate ISP in the wild. 

We further evaluate different versions of the color mapping scheme $\mathcal{C}$.
\textbf{LinearMap:} First, we consider learning a $3\times3$ global color correction matrix between the processed RAW $x'$ and the color $c$ for each training pair, as in \cite{deepburstsr}. LinearMap produces inaccurately colored images specially in terms of the contrast, since it cannot represent more complex color transformations and tone curves.
% LinearMap produces inaccurately colored images specially in terms of the contrast due to the non-local addressing of the problem. 
% specially in terms of contrast
% During training, the color $c$ is set to the downsampled ground truth and during inference $c=\mathcal{G}(x)$. 
% $c=y^\downarrow_{x'}$
%LinearMap achieves a PSNR of 21.89 dB and a SSIM of 0.834.
\textbf{ConstValMap:} Here, we use a simplified version of our approach (Sec.~\ref{sec:color_map}) as $\mathcal{C}$ by using fixed values for each bin instead of the affine mapping learned in Sec.~\ref{sec:color_map}. Channel dependence is not exploited in this version for calculating the values. This achieves a substantial improvement of 0.76 dB in PSNR over LinearMap. Thus, proving the utility of using a more flexible color mapping formulation.
\textbf{AffineMapIndep:} Setting $\mathcal{C}$ to our color mapping scheme (Sec.~\ref{sec:color_map}) but without any channel dependence boosts the PSNR by a further 1.13 dB over ConstValMap. Increasing the expressive power of each bin by predicting an affine transform instead of a constant is thus pivotal for better performance of our color conditional RAW-to-sRGB mapping. \textbf{AffineMapDep:} Here, $\mathcal{C}$ is set to our full formulation discussed in Sec.~\ref{sec:color_map}. Thus, exploiting channel dependence in $\mathcal{C}$ is beneficial as quantified by the PSNR increase of 0.63 dB \wrt AffineMapIndep. \textbf{+Preprocess:} Finally, we add our pre-processing network $\mathcal{P}$ (Sec.~\ref{sec:preprocess_net}) to the AffineMapDep version. This gives an impressive boost of 0.83 dB in PSNR over AffineMapDep hence, validating the need to remove noise and pre-process the phone RAW before color mapping. 

\begin{table}[t]
\caption{Ablative study of our color mapping scheme (Sec.~\ref{sec:color_map}) on the ZRR dataset.}
%\vspace{-2mm}%
\label{tab:color_map}%
\centering%
	\resizebox{0.95\linewidth}{!}{%
\begin{tabular}{lccccccc}
	\toprule
	&\textbf{NoColorPred}&\textbf{ColorBlur}&\textbf{LinearMap}&\textbf{ConstValMap}&\textbf{AffineMapIndep}&\textbf{AffineMapDep}&\textbf{+Preprocess}
	\\\midrule
\textbf{PSNR$\uparrow$}&21.27&23.43&21.89&22.65&23.78&24.41&25.24\\
	\textbf{SSIM$\uparrow$}&0.844&0.857&0.832&0.859&0.861&0.873&0.879\\\bottomrule
\end{tabular}}
%\vspace{-2mm}
\end{table}

\begin{table}[t]
\centering%
\begin{minipage}[t]{0.4\linewidth}%
\caption{Ablative study of our loss (Sec.~\ref{sec:loss}) on the ZRR dataset.}
\label{tab:loss}
\centering%
	\resizebox{\linewidth}{!}{%
\begin{tabular}{lccc}
	\toprule
	&\textbf{NoAlign}&\textbf{+AlignedLoss}&\textbf{+Mask}
	\\\midrule
    \textbf{PSNR$\uparrow$}&20.56&24.62&25.24\\
	\textbf{SSIM$\uparrow$}&0.785&0.867&0.879\\\bottomrule
\end{tabular}}
\end{minipage}\hfill%
\begin{minipage}[t]{0.57\linewidth}%
\caption{Ablative study of our color prediction network (Sec.~\ref{sec:colorinference}) on the ZRR dataset}%
\label{tab:color_pred}%
\centering%
	\resizebox{\linewidth}{!}{%
\begin{tabular}{lcccc}
	\toprule
	&\textbf{NoColorPred}&\textbf{+U-Net}&\textbf{+Reconstruct}&\textbf{+GlobalContext}
	\\\midrule
    \textbf{PSNR$\uparrow$}&21.27&24.09&24.43&25.24\\
	\textbf{SSIM$\uparrow$}&0.844&0.865&0.871&0.879\\\bottomrule
\end{tabular}}
\end{minipage}
\end{table}

\subsection{Ablative Study of the Training Loss}
\label{sec:ablation_loss}
Here, we study the effect of our masked aligned loss (Sec.~\ref{sec:loss}). The results on the ZRR dataset are reported in Tab.~\ref{tab:loss}. See Appendix for a visual comparison.

\textbf{NoAlign:} As a baseline for ablating our loss, we employ an unaligned $L_1$-loss for all our objectives (Eq.~\eqref{eq:pred_mask},~\eqref{eq:preprocess_loss} and~\eqref{eq:clr_pred_loss}). The mask $m$ is set to 1 at all locations. %NoAlign produces blurry results due to the misalignment between the phone RAW and the corresponding DSLR sRGB during training. Employing a non-aligned loss with no masking gives a PSNR of 20.56 dB and a SSIM score of 0.788.
\textbf{+AlignedLoss} Further, employing alignment before the loss calculation leads to more crisp predictions, giving a large improvement of 4.06 dB in PSNR and a relative gain of 10.4$\%$ in SSIM. Although improving the results, the prediction lacks detail and is characterized by a noticeable color shift. This is due to the inaccuracies in optical flow computations that may occur due to occlusions and homogenous regions. \textbf{+Mask} Finally, our masking strategy using Eq.~\eqref{eq:mask} leads to a significant gain of 0.62 dB in PSNR. (+Mask) produces a more detailed output with colors consistent with the target DSLR sRGB. This shows that accurate supervision using our masked loss during training is beneficial to our DSLR sRGB restoration network.

% \begin{table}
% \caption{Ablative study for our loss (Sec. ~\ref{sec:loss}) on the ZRR dataset}
% \label{tab:loss}
% \centering
% 	\resizebox{0.4\linewidth}{!}{%
% \begin{tabular}{lccc}
% 	\toprule
% 	&\textbf{NoAlign}&\textbf{+AlignedLoss}&\textbf{+Mask}
% 	\\\midrule
%     \textbf{PSNR$\uparrow$}&20.56&24.62&25.24\\
% 	\textbf{SSIM$\uparrow$}&0.785&0.867&0.879\\\bottomrule
% \end{tabular}}
% \end{table}

\subsection{Ablative Study of the Color Prediction Network}
\label{sec:ablation_color_pred}
Next, we study the effect of our color prediction module (Sec.~\ref{sec:colorinference}). The results on the ZRR dataset are reported in Tab.~\ref{tab:color_pred}.

% Additional analysis (suppl):
% - depth
% - num latents
\textbf{NoColorPred:}
This is the same baseline as in Sec.~\ref{sec:ablation_color_map}, which employs no explicit color prediction or conditioning. \textbf{U-Net:} Integrating a low resolution U-Net based color predictor without the reconstruction branch or global context transformer leads to an impressive gain of 2.82 dB over NoColorPred. This demonstrates the effectiveness of conditioning $\mathcal{F}$ on the color image for robust ISP learning and prediction. \textbf{+Reconstruct:} Further, integrating a reconstruction branch in our color predictor helps $\mathcal{G}(x)$ in learning a more informative encoding $E_\text{phone}(x)$, leading to a 0.34 dB increase in PSNR. Thus, +Reconstruct facilitates our encoder in the color predictor module to encapsulate all the information into the encoding that is necessary for accurate color prediction. \textbf{+GlobalContext:} Finally, integrating the global context transformer (Sec. ~\ref{sec:transformer}) in our U-Net color predictor $\mathcal{G}(x)$ provides our color conditional ISP net $\mathcal{F}(x, \hat{c})$ with a substantial gain of 0.81 dB. This clearly demonstrates the importance of exploiting global information in predicting coherent colors.

\begin{figure*}[t]
\newcommand{\wid}{0.17\textwidth}
    \centering
    \subfloat[MWISPNet]{
    \begin{minipage}[b]{\wid}
        \includegraphics[width=\linewidth]{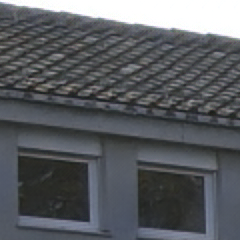}
        \includegraphics[width=\linewidth]{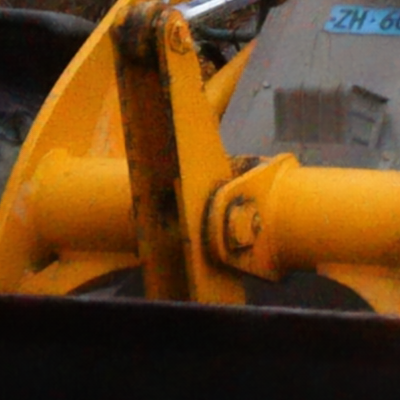}
        % \label{fig:res_mwnet2}
    \end{minipage}}
    \subfloat[AWNet]{
    \begin{minipage}[b]{\wid}
        \includegraphics[width=\linewidth]{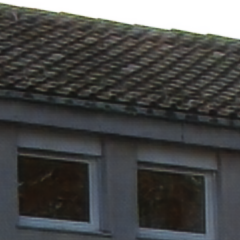}
        \includegraphics[width=\linewidth]{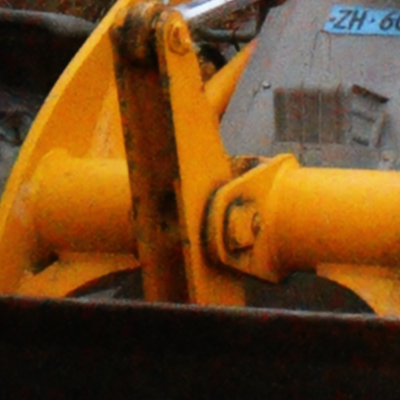}
        % \label{fig:res_awnet12}
    \end{minipage}}
    \subfloat[LiteISPNet]{
    \begin{minipage}[b]{\wid}
        \includegraphics[width=\linewidth]{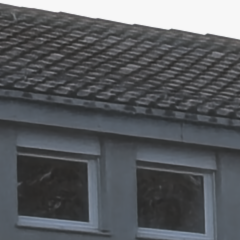}
        \includegraphics[width=\linewidth]{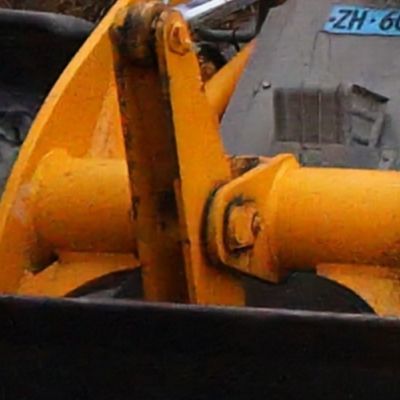}
        % \label{fig:res_liteisp2}
    \end{minipage}}
    \subfloat[Ours]{
    \begin{minipage}[b]{\wid}
        \includegraphics[width=\linewidth]{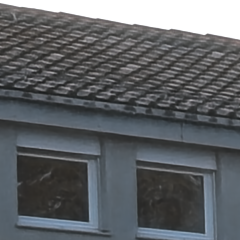}
        \includegraphics[width=\linewidth]{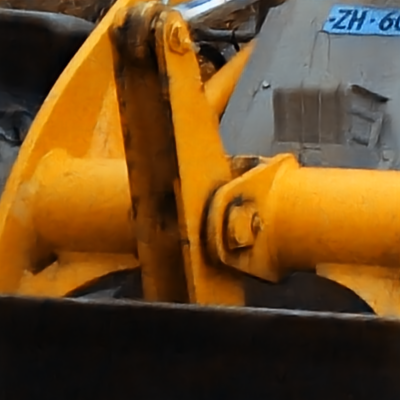}
        % \label{fig:res_ours2}
    \end{minipage}}
    \subfloat[GT]{
    \begin{minipage}[b]{\wid}
        \includegraphics[width=\linewidth]{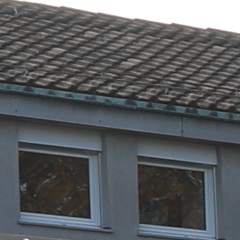}
        \includegraphics[width=\linewidth]{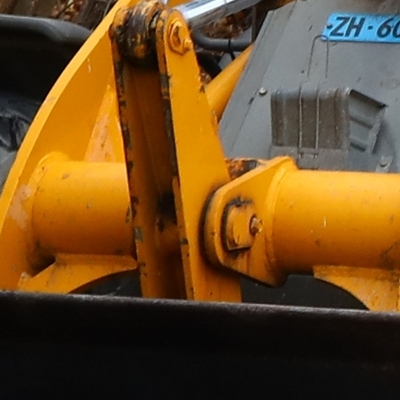}
        % \label{fig:res_gt2}
    \end{minipage}}
    \caption{Visual results for state-of-the-art comparison on our ISPW dataset (first row) and the ZRR dataset (second row). Best viewed with zoom.}\label{fig:sota}
\end{figure*}

\subsection{State-of-the-Art Comparison}

In this section, we compare our color conditional ISP network with state-of-the-art methods for RAW-to-sRGB mapping, namely PyNet ~\cite{PyNet}, MW-ISPNet~\cite{ZRR}, AWNet~\cite{awnet} and LiteISPNet~\cite{liteispnet}. We evaluate on the test splits of the ZRR dataset~\cite{ZRR} and our ISPW dataset (Sec.~\ref{sec:dataset}). Among these methods, MW-ISPNet, AWNet and LiteISPNet employ discrete wavelet transforms for incorporating global context. To deal with misalignments, MW-ISPNet, AWNet and PyNet incorporate the VGG perceptual loss~\cite{vggloss}, while LiteISPNet employs an aligned loss using optical flow computation~\cite{pwcnet}.

\begin{table}[!t]
 \caption{State-of-the-Art comparison on the ZRR~\cite{ZRR} and our ISPW datasets.}
 \label{tab:sota}
 \centering%
 %  \caption{Ablative study: Results are reported in terms of accuracy (\%) with 95\% confidence interval.}
	\resizebox{0.85\columnwidth}{!}{%
 \begin{tabular}{lccc@{~~~~~~}ccc}
	\toprule
	&\multicolumn{3}{c}{\textbf{ZRR Dataset}}&\multicolumn{3}{c}{\textbf{ISPW Dataset}}\\
 \cline{2-4} \cline{5-7}
	&\textbf{PSNR$\uparrow$}&\textbf{SSIM$\uparrow$}&\textbf{Time(ms)}&\textbf{PSNR$\uparrow$}&\textbf{SSIM$\uparrow$}&\textbf{Time(ms)}\\\midrule
	\textbf{PyNet}~\cite{PyNet}&22.73&0.845&62.7&-&-&-\\
	\textbf{MW-ISPNet}~\cite{ZRR}&23.13&0.849&111.3&22.43&0.746&99.4\\
	\textbf{AWNet}~\cite{awnet}&23.52&0.855&63.4&23.10&0.787&50.8\\
	\textbf{LiteISPNet}~\cite{liteispnet}&23.81&0.873&23.3&23.51&0.809&17.2\\
	\textbf{Ours}&\textbf{25.24}&\textbf{0.879}&67.6&\textbf{25.05}&\textbf{0.821}&55.7\\
 \bottomrule
\end{tabular}}
\end{table}
Table~\ref{tab:sota} lists the quantitative results on the test split of the ZRR dataset that contains 1203 RAW-sRGB crop pairs of size 448$\times$448. Our method outperforms all previous approaches by a significant margin, achieving a gain of 1.43 dB PSNR compared to the second best method: the very recent LiteISPNet. We then run the best performing methods on the test split of our ISPW dataset, that contains 3023 RAW-sRGB crop pairs of size 320$\times$320. For a fair comparison, all the methods were retrained on our dataset using apt train settings. The performance gap between our color conditional ISP network and other methods is more stark for the ISPW dataset, with our approach achieving a PSNR 1.54 dB higher than the second best LiteISPNet. 

Figure~\ref{fig:sota} shows the visual results for our color conditional ISP compared to the top three performing methods. Compared to our approach, all the other three methods fail to capture the accurate color of the target DSLR sRGB. Moreover, the results for MW-ISPNet and AWNet are blurry due to their inability to handle misalignment well. On the other hand, although LiteISPNet employs an aligned loss, it fails to account for inconsistent flow computations hence leading to significant color shift and loss of detail. Conversely, our approach produces crisp DSLR-like sRGB predictions with accurate colors, thus proving the utility of our global attention based color predictor paired with our masked aligned loss. The blur and color shift effect is more intense for all other methods on our dataset that contains misaligned RAW-sRGB pairs.  Finally, we calculate the average inference time per image for our method on both the datasets. We achieve an average per image inference times of 67.6 ms and 55.7 ms, respectively on the sRGB images of sizes 448$\times$448 (ZRR dataset) and 320$\times$320 (ISPW dataset).
\section{Conclusion}
We address the problem of mapping RAW sensor data from a phone to a high quality DSLR image by modelling it as a conditional ISP framework on the target color. To aid our color conditional ISP net during inference, we propose a novel encoder-decoder based color predictor that encapsulates an efficient global attention module. A flexible parametric color mapping scheme from RAW to the target color is integrated for a robust training and inference. Finally, we propose a masked aligned loss for filtering out regions with inconsistent optical flow during aligned loss calculations.  We perform experiments on the ZRR dataset and our ISPW dataset, setting a new state-of-the-art on both the datasets. 

\section*{Acknowledgements}
This work was supported by the ETH Z\"urich Fund (OK), a Huawei Technologies Oy (Finland) project and the Alexander von Humboldt Foundation.
\clearpage
\setcounter{section}{0}
\renewcommand{\thesection}{\Alph{section}}

\begin{center}
	\textbf{\large Appendix}
\end{center}
%%%%%%%%%%%%%%%%%%%%%%%%%%%%%%%%%%%%%%%%CVPR stuff
In the appendix, we present details such as the network architecture for each of the components in our architecture. We also provide additional full-resolution results for our approach. Further, we provide additional ablations and some more qualitative results. Concretely:

\begin{itemize}
    \item We provide the closed-form solution for the minimization problem stated for our color mapping (Sec. ~\ref{sec:color_map}) (Sec.~\ref{sec:closed_form_supp}).
    \item We provide details about the network architecture and some other important details for all the components in our framework (Sec.~\ref{sec:rch_supp}).
    \item We provide some additional ablations and qualitative results for the ablations stated in the manuscript (Sec.~\ref{sec:ablation_supp}).
    \item We provide some additional full resolution results for our approach (Sec.~\ref{sec:full_res_results_supp}).
    \item We provide some more qualitative results for state-of-the-art comparisons of our method with other approaches (Sec.~\ref{sec:sota_supp}).
    \item We provide some example captures from our ISPW dataset (Sec.~\ref{sec:dataset_supp}).
    \item We visualize the intermediate results for our ISP in the wild pipeline (Sec.~\ref{sec:intermediate_results_supp}).
    \item We provide some additional experiments for our approach (Sec.~\ref{sec:additional_exp}).
\end{itemize}

\section{Color Mapping}
\label{sec:closed_form_supp}
Here, we present the closed form solution to the minimization problem for learning the affine transformation for each bin centroid in our color mapping scheme (Sec. ~\ref{sec:color_map}) stated in equation ~\ref{eq:min_problem}. We define $V^{j}_b\in\mathbb{R}^{4\times 1}$ as the affine transform calculated for bin centroid $b$ and channel $j$.  $V^{j}_b$ is a column vector of length 4 that contains $A^{j}_b\in\mathbb{R}^{3\times 1}$ as the first 3 elements and $B^{j}_b\in\mathbb{R}$ as the last element. Using pseudo-inverses:

% \begin{equation}
%     V^{j}_b = (X^TX + \lambda I)^{-1}X^Ty^j
% \end{equation}
\begin{equation}
    V^{j}_b = (\Tilde{X}^T\Tilde{X})^{-1}\Tilde{X}^Tc^j
\end{equation}

Here, $\Tilde{X}\in\mathbb{R}^{N\times 4}$, where $N$ is the total number of pixels in $\Tilde{x}$ which is the output of our pre-processing network $\mathcal{P}$ (Sec. ~\ref{sec:color_map}). The $i^{th}$ row of $\Tilde{X}$, $\Tilde{X}_i=\sqrt{w^j_{ib}}[\Tilde{x}^{1}_i\text{ }\Tilde{x}^{2}_i\text{ }\Tilde{x}^{3}_i\text{ }1]$. And $c^j\in\mathbb{R}^{N\times 1}$ are the intensity values of the $j^{th}$ channel in the target color image $c$. Note that the color image $c$ is given by the downsampled target DSLR sRGB during training and during inference, $c=\mathcal{G}(x)$ is given by our color prediction network (Sec. ~\ref{sec:colorinference}). Further, $\Tilde{x}^1_i$, $\Tilde{x}^2_i$ and $\Tilde{x}^3_i$ are the intensity values of the red, green and blue channels, respectively at the $i^{th}$ location in the pre-processed source image $\Tilde{x}$ (Sec. ~\ref{sec:color_map}) .  The weights $w^j_{ib}$ are calculated as in Sec. ~\ref{sec:color_map}. 

\section{Network Architecture and Other Details}
\label{sec:rch_supp}
In this section, we provide the network architectures for each of the components proposed in our ISP Net. 

\subsection{The Color Conditional ISP Network}
Here, we discuss the architecture for our color conditional RAW-to-sRGB network. Our DSLR sRGB network $\mathcal{F}(x, \hat{c})$ is conditioned on the color $\hat{c}$. Hence, it takes a 7-channel input which we get by concatenating the 4-channel phone RAW $x$ and the 3-channel color $\hat{c}$ in the channel dimension. Our restoration net $\mathcal{F}$ comprises of a convolutional layer followed by 8 Residual-in-Residual Dense Blocks (RRDB)~\cite{esrgan}. The resulting feature map is 2x up-scaled using an upconv layer. Our upconv layer applies a convolution followed by a leakyReLu to the 2x up-scaled feature map from the RRDB layer via nearest-neighbour interpolation.  

\begin{figure}[h]
\centering
\includegraphics[width=1.0\columnwidth]{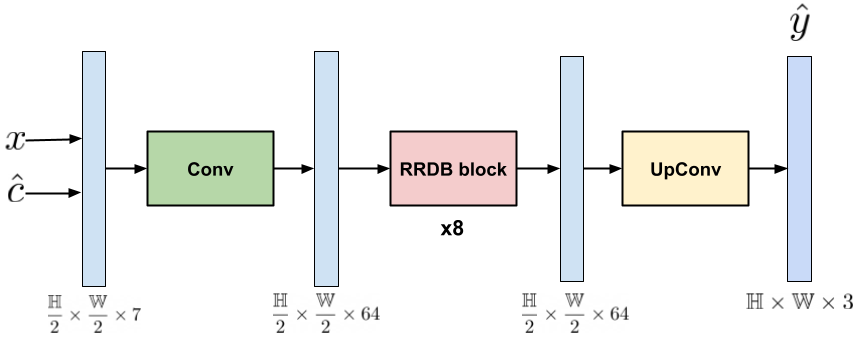}
\caption{Our color conditional DSLR sRGB restoration network $\mathcal{F}$.}
\label{fig:F_arch}
\end{figure}

\subsection{The Pre-processing Network}
\label{sec:preproc_net_supp}
\begin{figure}[h]
\centering
\includegraphics[width=1.0\columnwidth]{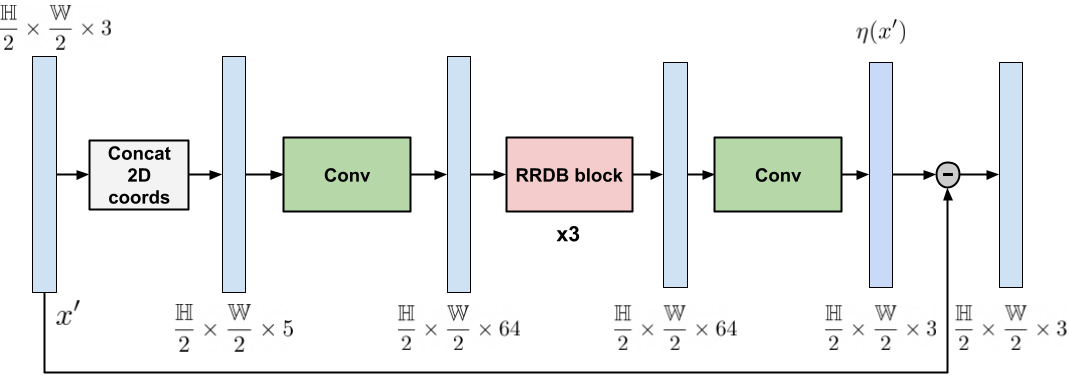}
\caption{Our RAW pre-processing network $\mathcal{P}$.}
\label{fig:P_arch}
\end{figure}

Here, we state the architecture for our pre-processing net $\mathcal{P}$. The pre-processing net $\mathcal{P}$ comprises of a noise estimation module $\eta$. The architecture for our pre-processing network $\mathcal{P}$ is shown in Fig.~\ref{fig:P_arch}. It is important to note that 2-channel 2D positional coordinates are concatenated in the channel dimension to the 3-channel processed RAW $x'$ to mitigate the effects of vignetting that is a common phenomenon in RAW data. 

The processed phone RAW $x'=\Gamma(x)$ is a rough visualization of the RAW data $x$. We define the operation $\Gamma(x)$, henceforth. To get $x'$, we first  neglect one of the green channels in $x$ and then normalize the resulting 3-channel image between $[0, 1]$. Further, we apply a constant approximate gamma correction to the final processed image $x'$. The scaling and gamma correction operations can be listed as: 

\begin{align}
\label{eq:process}
    x'^1 &\coloneqq \big(x^1/\text{max}(x^1_{max}, 1/2.5)\big)^\frac{1}{2.2} \\
    x'^2 &\coloneqq \big(x^3/\text{max}(x^3_{max}, 1)\big)^\frac{1}{2.2}\\
    x'^3 &\coloneqq \big(x^4/\text{max}(x^4_{max}, 1/1.4)\big)^\frac{1}{2.2}.
\end{align}
The above operations encompass the functional $\Gamma(x)$. Here, $x'^1$, $x'^2$ and $x'^3$ are the red, green and blue channels, respectively of $x'$. And, $x^1_{max}$, $x^3_{max}$ and $x^4_{max}$ are the max values in the red, green (one of the green) and blue channels, respectively of the RAW $x$. The specific scaling factors in the above mentioned power law were arrived by quantitative evaluation of the data. Further, the gamma correction factor of 1/2.2 is a commonly used value in imaging systems. 

\subsection{The Color Prediction Network}
\begin{figure}[h]
\centering
\includegraphics[width=1.0\columnwidth]{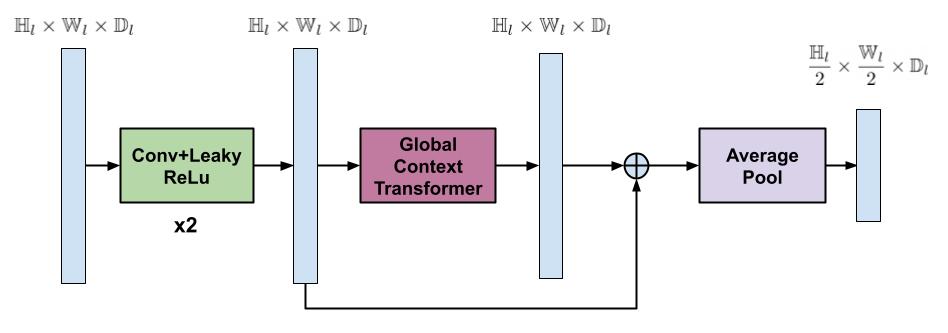}
\caption{The encoder blocks in the contracting path of our DSLR color predictor $\mathcal{G}$.}
\label{fig:encoder_supp}
\end{figure}

\parsection{Encoder block}
Figure~\ref{fig:encoder_supp} shows the architecture at each of the levels in the contracting path of our U-Net. Each of these modules comprises of 2 convolutional layers comprising of successive convolution and leakyReLu activations. The convolutional layer is followed by an efficient Global . A skip connection between the input and output of the Global Context Transformer  makes the learning more stable and efficient. The resulting feature map is then average pooled and passed on to the next contracting level. The number of input channels at level $l$ is given by $\mathbb{D}_l=64\times2^l$ where $l\in\{1, 2, 3\}$. For level $l=0$, $\mathbb{D}_l=6$ \textit{i.e.} the phone RAW data is concatenated with the 2D positional coordinates  to mitigate vignetting that is a common in RAW sensor data. For the Global Context Transformer, the learned latent vector $Z_l\in\mathbb{R}^{\frac{1024}{2^l}\times2^{l+7}}$ at level $l$ of the contracting path. Fixing the size of the latent vectors limits the computational complexity for attention to linear in the input instead of quadratic. The number of levels in both, the contracting and expanding path's is set to 4.

\begin{figure}[h]
\centering
\includegraphics[width=1.0\columnwidth]{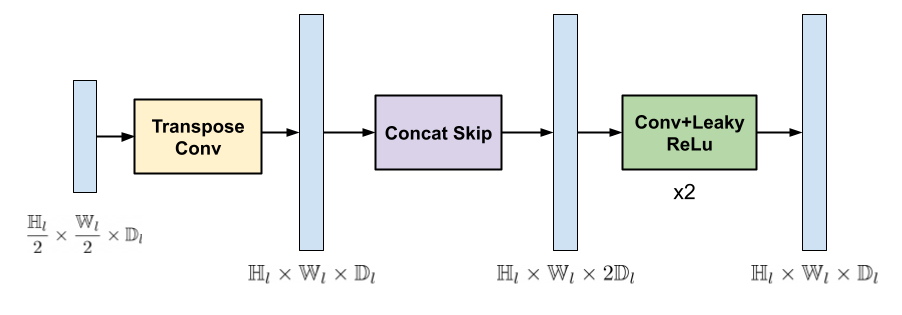}
\caption{The decoder blocks in the expanding path of our color predictor $\mathcal{G}$.}
\label{fig:decoder_supp}
\end{figure}

\parsection{Decoder block}
Figure ~\ref{fig:decoder_supp} shows the architecture at each of the levels in the expanding paths (both our decoders). Each of these modules comprises of a transposed2D convolution with kernel size=2 and the stride=2. This is followed by concatenating the features from the corresponding level in the contracting path. The resulting feature map is finally passed through a couple of convolutional layers comprising of successive convolutions and leakyReLu activations.

As a final layer, our RAW reconstruction decoder applies an extra $3\times3$ convolution to the output of the respective U-Net decoder branch. And, the DSLR color predictor branch employs a RRDB block to the output of the respective decoding branch.  

\section{Detailed Ablative Experiments}
\label{sec:ablation_supp}
In this section, we provide additional ablations for our approach and provide qualitative results for the ablations discussed in the manuscript (Sec. ~\ref{sec:ablation}).

\subsection{Additional Ablations}
In addition to the ablations provided in the manuscript, here we provide some more ablations on the test set of the ZRR dataset. The evaluation criteria remains the same as in the manuscript. 

\begin{table}[h]
 \centering%
 \caption{Impact of joint fine-tuning of our model components $\mathcal{F}$ and $\mathcal{G}$, starting from the independent training used in the paper. Results listed on the ZRR dataset.}
	\resizebox{0.6\columnwidth}{!}{%
 \begin{tabular}{lc@{~~~~~}c}
	\toprule
 	&\textbf{Independent\ Train}&\textbf{Joint Fine-tuning}\\\midrule
	\textbf{PSNR$\uparrow$}&25.24&25.27\\
	\textbf{SSIM$\uparrow$}&0.879&0.883\\
 \bottomrule
\end{tabular}}
 \label{tab:indep_joint_supp}
\end{table}

\parsection{Impact of joint fine-tuning of our model components $\mathcal{F}$ and $\mathcal{G}$, starting from the independent training} Here, we do a comparative study of the independent training of our ISP network $\mathcal{F}$ and Color Prediction $\mathcal{G}$ versus the joint fine-tuning of $\mathcal{F}$ and $\mathcal{G}$. Training $\mathcal{F}$ and $\mathcal{G}$ independently allows us to use larger batch sizes, hence faster convergence of the training. We investigate joint fine-tuning of both, our ISP net $\mathcal{F}$ and the Color Prediction net $\mathcal{G}$ by starting from the independently pretrained $\mathcal{F}$ and $\mathcal{G}$ models. The batch size is reduced to 8 (versus 16 when we train $\mathcal{F}$ and $\mathcal{G}$ independently). Table \ref{tab:indep_joint_supp} shows the effect of this joint fine-tuning compared to independent training of our $\mathcal{F}$ and $\mathcal{G}$ on the ZRR dataset. It is evident from Tab.\ref{tab:indep_joint_supp} that the improvement is negligible when we jointly fine-tune our ISP net $\mathcal{F}$ and our color predictor $\mathcal{G}$. Thus, justifying our choice of independently training $\mathcal{F}$ and $\mathcal{G}$.   

\begin{table}[t]
 \centering%
 \caption{Impact of different alignment strategies for ISP Network Loss computation (Eq. ~\ref{eq:pred_mask} of the manuscript). Results listed on the ZRR dataset.}
	\resizebox{0.8\columnwidth}{!}{%
 \begin{tabular}{lc@{~~~~~}c@{~~~~~}c}
	\toprule
 	&\textbf{Align GT}&\textbf{Align GT }&\textbf{Align Prediction}\\
	&\textbf{with RAW}&\textbf{with Prediction}&\textbf{with GT}\\\midrule
	\textbf{PSNR$\uparrow$}&25.09&25.24&25.26\\
	\textbf{SSIM$\uparrow$}&0.874&0.879&0.881\\
	\textbf{Training time (hrs)$\downarrow$}&26.0&26.8&29.2\\
 \bottomrule
\end{tabular}}
 \label{tab:align_loss_supp}
\end{table}

\parsection{Impact of different alignment strategies for ISP Network loss computation}
Next, we analyze the different alignment strategies in our ISP Network Loss (Eq. ~\ref{eq:pred_mask} of the manuscript). First, we report results for align the DSLR sRGB with the phone RAW (Align GT with RAW) for ISP Network Loss calculation. We observe a drop in performance compared to the case where we align the DSLR sRGB with the ISP Net prediction (Align GT with Prediction). This drop can be explained by the fact that aligning the DSLR sRGB with the RAW involves estimation of the optical flow in a low resolution (downsampled DSLR sRGB aligned with $x'$) and then upscaled (via bilinear interpolation) by a factor of 2. This introduces some warping inaccuracies and hence, the drop in performance. On the other hand, aligning the ISP Net prediction with the DSLR sRGB (Align Prediction with GT) gives a very slight improvement in terms of the PSNR while increasing the training time of the ISP Net $\mathcal{F}$ by almost 10\% because this alignment strategy involves differentiating
through the warping process. Hence, we align the DSLR sRGB with the ISP Net prediction for the ISP Network Loss calculation.  

We also time each of our training iterations (with a batch size of 16). Computation of the optical flow and warping in each training step is not the bottleneck: only 11\% of the time in a training iteration (2.6s). The forward time was found to be $~1.1s$, while the backward time was $~0.9s$. The total loss calculation takes $~0.6s$ (this also encompasses the optical flow). It is important to note that the timings are a bit inflated because of the time() function usage in python.

\begin{table}[h]
\centering
\caption{Additional ablative study for our color mapping scheme - unlike the ablation provided in the manuscript (Tab. ~\ref{tab:ablation1} of the manuscript), we feed in directly the color $\hat{c}=\mathcal{G}(x)$ into $\mathcal{F}$ without the color mapping $\mathcal{C}$ during inference. Results listed on the ZRR dataset.}
	\resizebox{0.4\linewidth}{!}{%
\begin{tabular}{lcc}
	\toprule
	&\textbf{PSNR$\uparrow$}&\textbf{SSIM$\uparrow$}
	\\\midrule
\textbf{NoColorPred}&21.27&0.844\\
	\textbf{ColorBlur}&23.43&0.857\\
	\textbf{LinearMap}&22.16&0.839\\
	\textbf{ConstValMap}&22.96&0.860\\
	\textbf{AffineMapIndep}&23.90&0.863\\
	\textbf{AffineMapDep}&24.46&0.873\\
	\textbf{+Preprocess}&25.19&0.878\\\bottomrule
\end{tabular}}
\label{tab:ablation1_supp}
\end{table}

\parsection{Effect of color mapping during inference}
We additionally ablate the use of our color mapping scheme $\mathcal{C}$ at inference for our approach. In table ~\ref{tab:ablation1} of the manuscript, we provided the ablation for various color mapping schemes. Here, we provide an additional ablation (Tab.~\ref{tab:ablation1_supp}) where unlike in the manuscript, we feed in directly the color $\hat{c}=\mathcal{G}(x)$ into $\mathcal{F}$ without the color mapping $\mathcal{C}$ during inference. For each of the ablations the corresponding network is still trained with the respective color mapping scheme. From Tab.~\ref{tab:ablation1_supp}, it is evident that for the less powerful color mapping schemes, it is better to directly feed in the the color image $\hat{c}=\mathcal{G}(x)$ into our color conditional restoration network $\mathcal{F}$. On the other hand, we observe that using a powerful and a more flexible color mapping scheme like ours is beneficial during inference giving a boost of 0.05 in PSNR over the case where we do not employ the color mapping at inference (Tab.~\ref{tab:ablation1_supp}). Hence, in our final architecture we apply our color mapping from Pre-processed RAW $\Tilde{x}$ to the predicted color $c$ by our color prediction net $\mathcal{G}$ during inference. This provides an additional regularization for spurious local colors that may occur in $c$.         

\begin{table}[h]
\centering
\caption{Influence of using a processed RAW $x'$ in place of a 3-channel version of $x$ (by neglecting one of the green channels) for our color mapping and pre-processing network. Results listed on the ZRR dataset.}
	\resizebox{0.3\linewidth}{!}{%
\begin{tabular}{lcc}
	\toprule
	&\textbf{PSNR$\uparrow$}&\textbf{SSIM$\uparrow$}
	\\\midrule
\textbf{Ours-RAW}&24.97&0.875\\
	\textbf{Ours}&25.24&0.879\\
	\bottomrule
\end{tabular}}
\label{tab:ablation2_supp}
\end{table}

\parsection{Effect of using $x'$ instead of a 3-channel version (by neglecting one of the green channels) of the RAW $x$ in our framework}
Here, we provide an ablation for the utility of using the processed RAW $x'$ (Eq.~\eqref{eq:process}) instead of a 3-channel version of $x$ (by neglecting a green channel) in our color mapping $\mathcal{C}$ and our pre-processing network $\mathcal{P}$. Table~\ref{tab:ablation2_supp} shows that using a processed RAW $x'$ (Ours) aids both, our color mapping $\mathcal{C}$ and our pre-processing net $\mathcal{P}$. Hence, achieving an improvement in PSNR by 0.27 dB in comparison to the version where we use the RAW $x$ (Ours-RAW).   

\begin{table}[h]
\centering
\caption{Ablative study for exploiting the 2D positional coordinates of the RAW to counter vignetting. Results listed on the ZRR dataset.}
	\resizebox{0.4\linewidth}{!}{%
\begin{tabular}{lcc}
	\toprule
	&\textbf{PSNR$\uparrow$}&\textbf{SSIM$\uparrow$}
	\\\midrule
\textbf{Ours-No2DCoords}&25.07&0.877\\
	\textbf{Ours}&25.24&0.879\\
	\bottomrule
\end{tabular}}
\label{tab:ablation3_supp}
\end{table}

\parsection{Effect of concatenating the 2D positional coordinates to the input RAW for our pre-processing network and the color predictor}
Table ~\ref{tab:ablation3_supp} shows that using the 2D positional coordinates in our pre-processing network and the color predictor provides us an improvement of 0.17 dB in PSNR over Ours-No2DCoords where we do not concatenate the 2D positional information to the raw input in the pre-processing network $\mathcal{P}$ and our color predictor $\mathcal{G}$. It is important to note that we found concatenating the positional information only in $\mathcal{P}$ and $\mathcal{G}$ to be beneficial. We believe that this is due to the fact that our color conditional restoration net $\mathcal{F}$ is very efficient in exploiting the color information $c$ provided by the color predictor $\mathcal{G}$. 

\subsection{Color Mapping}
\begin{figure*}[h]
\captionsetup[subfigure]{labelformat=empty}
\newcommand{\wid}{0.24\linewidth}
    \centering
    \begin{subfigure}[b]{\wid}
        \includegraphics[width=\textwidth]{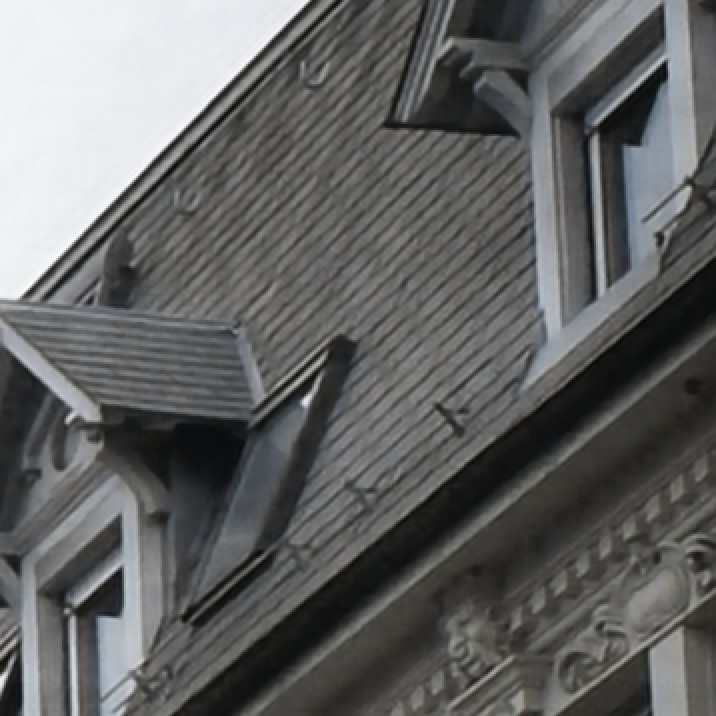}
        \caption{1a) NoColorPred}
        % \label{fig:res_mwnet2}
    \end{subfigure}
    \begin{subfigure}[b]{\wid}
        \includegraphics[width=\textwidth]{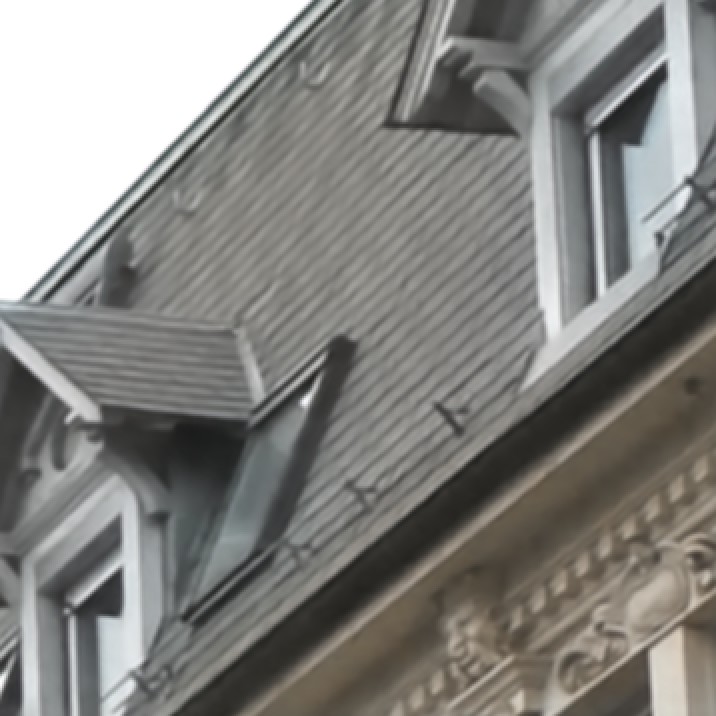}
        \caption{1b) ColorBlur}
        % \label{fig:res_awnet12}
    \end{subfigure}
    \begin{subfigure}[b]{\wid}
        \includegraphics[width=\textwidth]{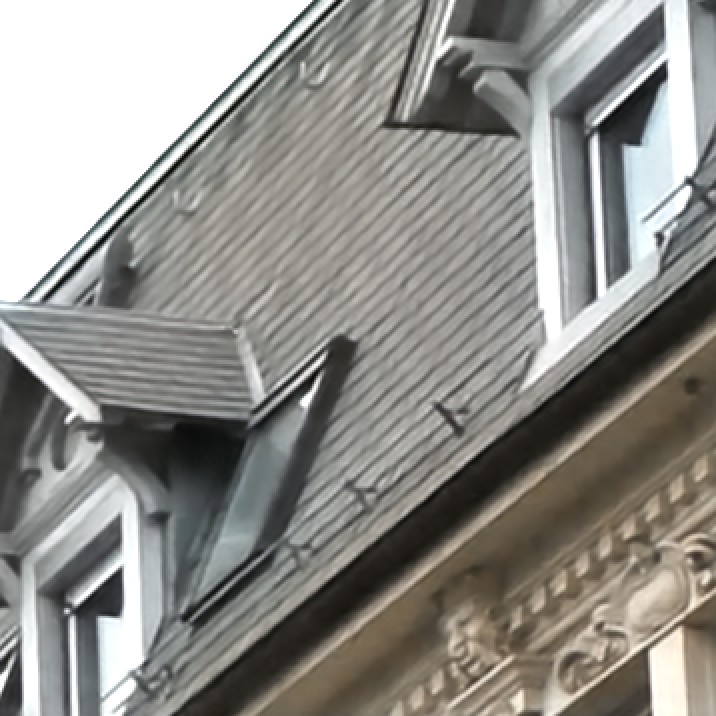}
        \caption{1c) LinearMap}
        % \label{fig:res_liteisp2}
    \end{subfigure}
    \begin{subfigure}[b]{\wid}
        \includegraphics[width=\textwidth]{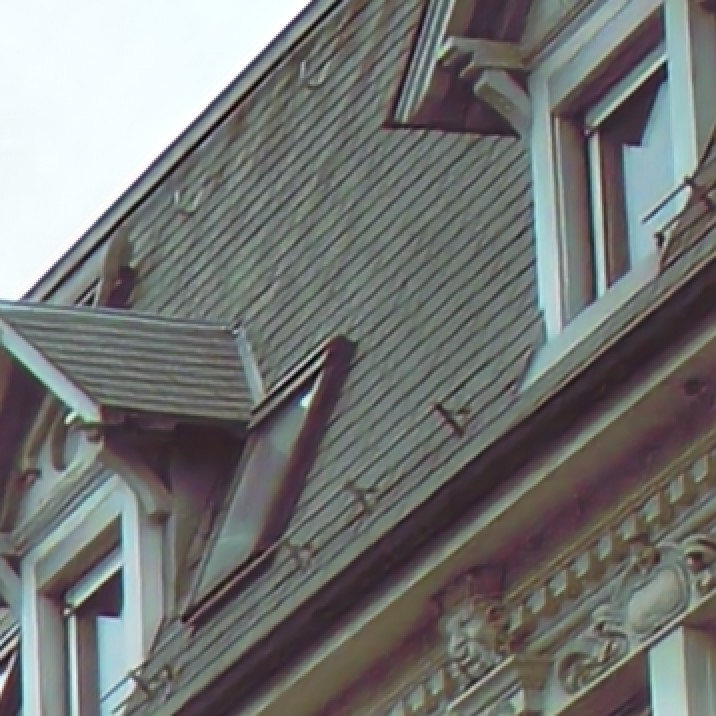}
        \caption{1d) ConstValMap}
        %\label{fig:res_gt2}
    \end{subfigure}
    
    \begin{subfigure}[b]{\wid}
        \includegraphics[width=\textwidth]{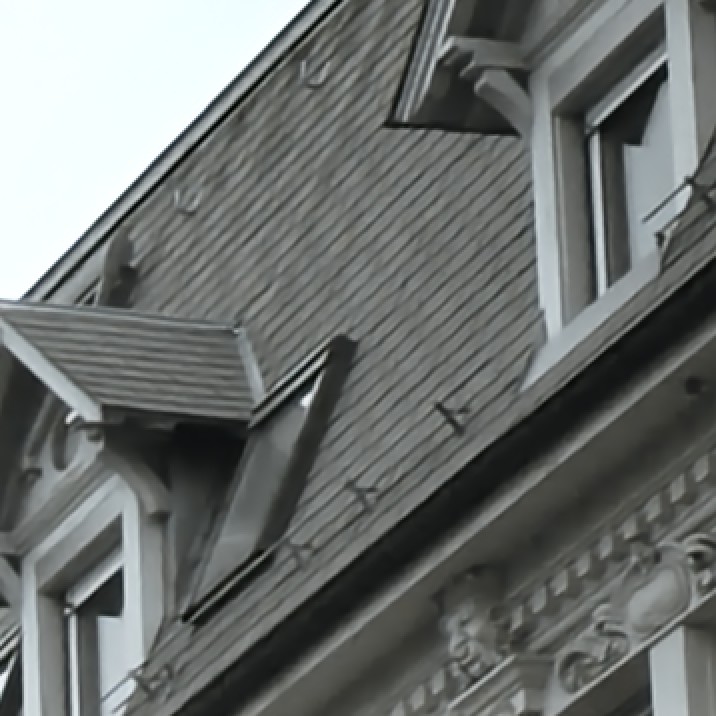}
        \caption{1e) AffineMapIndep}
        % \label{fig:res_mwnet2}
    \end{subfigure}
    \begin{subfigure}[b]{\wid}
        \includegraphics[width=\textwidth]{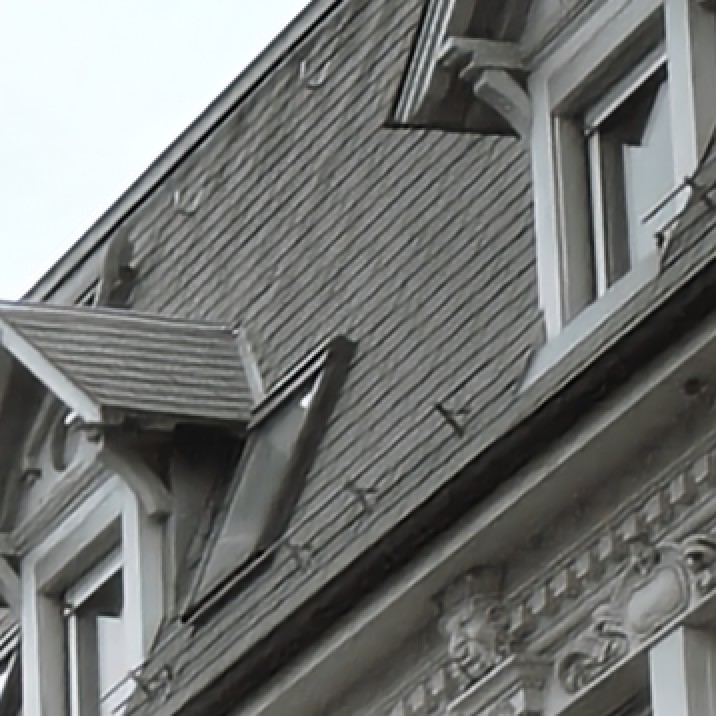}
        \caption{1f) AffineMapDep}
        % \label{fig:res_awnet12}
    \end{subfigure}
    \begin{subfigure}[b]{\wid}
        \includegraphics[width=\textwidth]{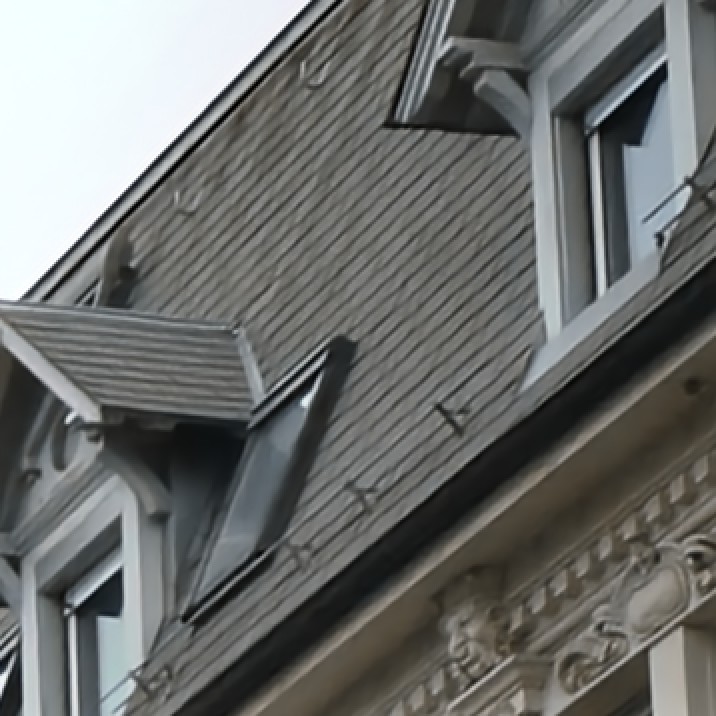}
        \caption{1g) +Preprocess}
        % \label{fig:res_liteisp2}
    \end{subfigure}
    \begin{subfigure}[b]{\wid}
        \includegraphics[width=\textwidth]{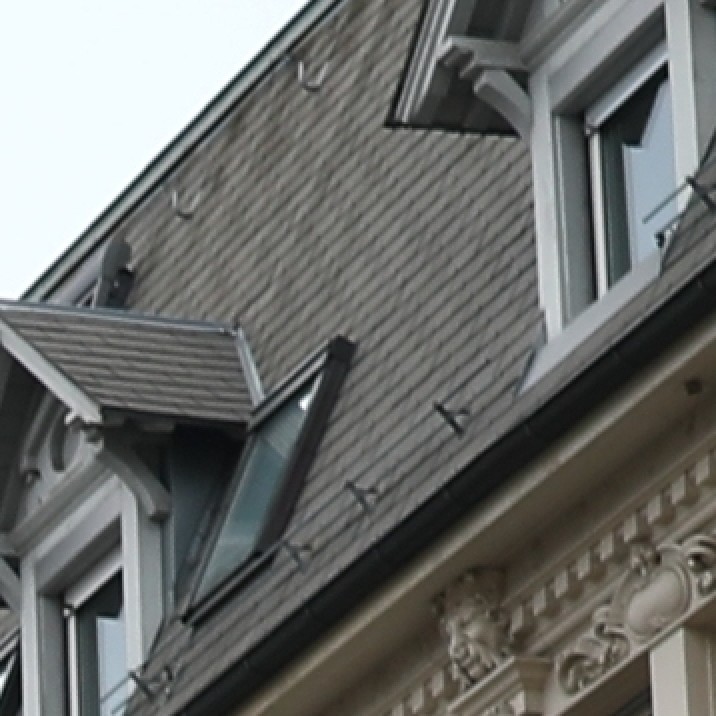}
        \caption{1h) DSLR sRGB}
        %\label{fig:res_gt2}
    \end{subfigure}
    
    \begin{subfigure}[b]{\wid}
        \includegraphics[width=\textwidth]{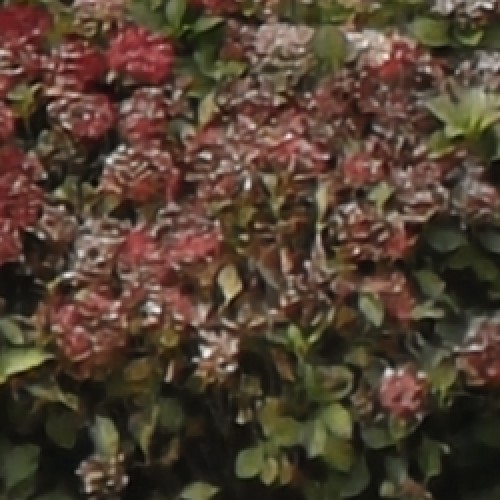}
        \caption{2a) NoColorPred}
        % \label{fig:res_mwnet2}
    \end{subfigure}
    \begin{subfigure}[b]{\wid}
        \includegraphics[width=\textwidth]{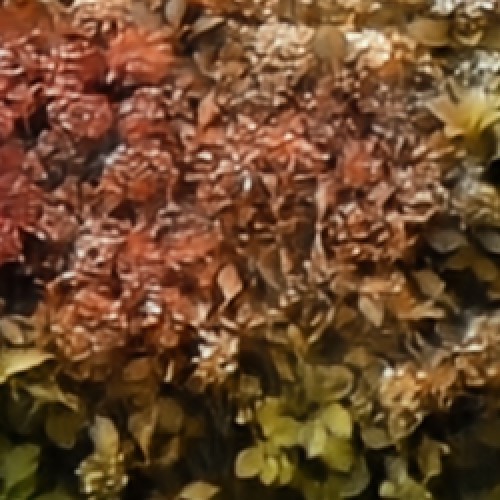}
        \caption{2b) ColorBlur}
        % \label{fig:res_awnet12}
    \end{subfigure}
    \begin{subfigure}[b]{\wid}
        \includegraphics[width=\textwidth]{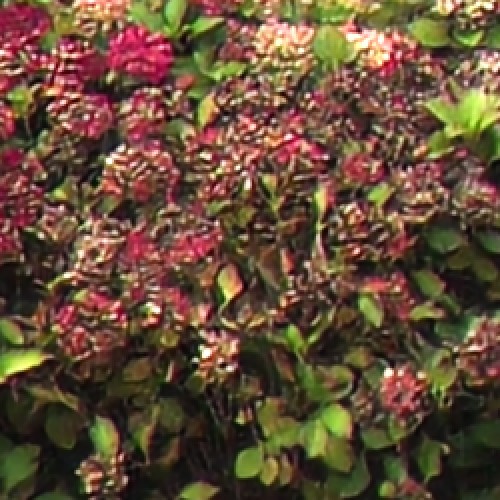}
        \caption{2c) LinearMap}
        % \label{fig:res_liteisp2}
    \end{subfigure}
    \begin{subfigure}[b]{\wid}
        \includegraphics[width=\textwidth]{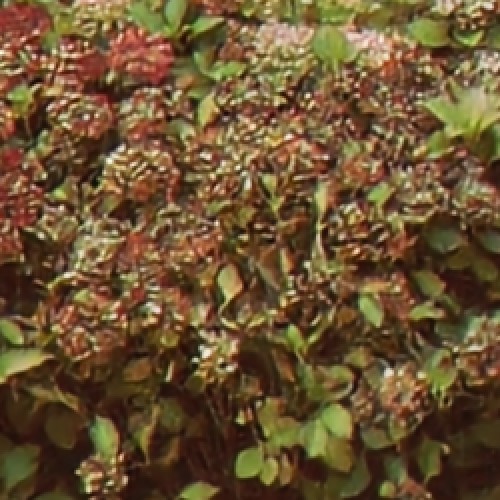}
        \caption{2d) ConstValMap}
        %\label{fig:res_gt2}
    \end{subfigure}
    
    \begin{subfigure}[b]{\wid}
        \includegraphics[width=\textwidth]{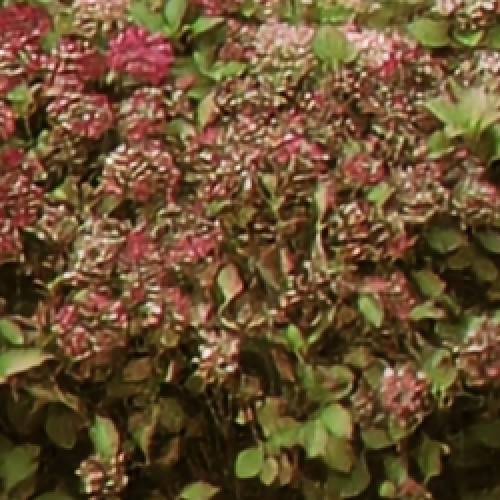}
        \caption{2e) AffineMapIndep}
        % \label{fig:res_mwnet2}
    \end{subfigure}
    \begin{subfigure}[b]{\wid}
        \includegraphics[width=\textwidth]{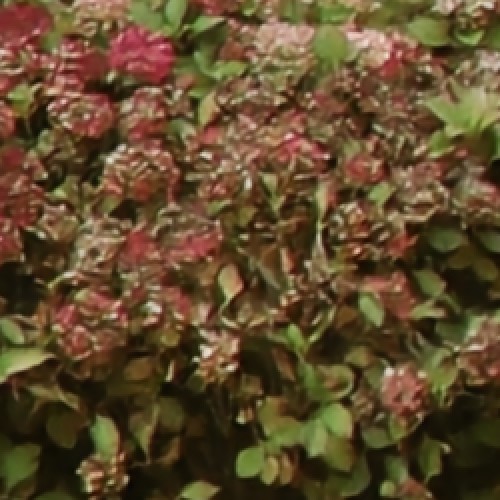}
        \caption{2f) AffineMapDep}
        % \label{fig:res_awnet12}
    \end{subfigure}
    \begin{subfigure}[b]{\wid}
        \includegraphics[width=\textwidth]{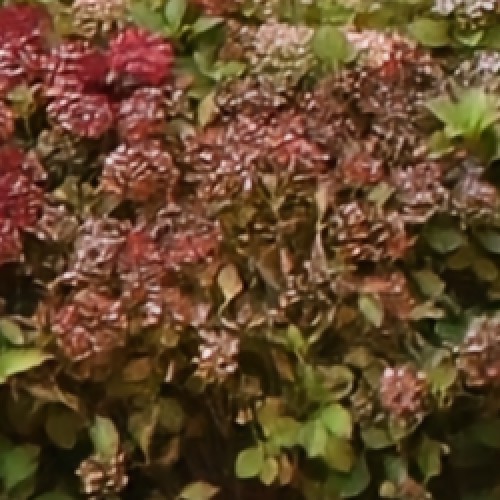}
        \caption{2g) +Preprocess}
        % \label{fig:res_liteisp2}
    \end{subfigure}
    \begin{subfigure}[b]{\wid}
        \includegraphics[width=\textwidth]{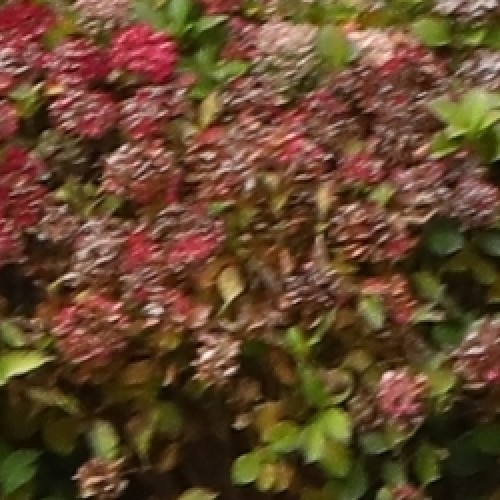}
        \caption{2h) DSLR sRGB}
        %\label{fig:res_gt2}
    \end{subfigure}
    \caption{Qualitative results for the ablation of our color mapping (Sec. ~\ref{sec:color_map} of the manuscript). These results demonstrate qualitatively our ablation study in section ~\ref{sec:ablation_color_map} of the manuscript. The crops are taken from the ZRR dataset. Best viewed with zoom.}\label{fig:ablation_colorMap_supp}
\end{figure*}

Figure~\ref{fig:ablation_colorMap_supp} shows the qualitative results for our ablative study for our proposed flexible soft attention based color mapping scheme (Sec. ~\ref{sec:color_map} of the manuscript). The qualitative results clearly demonstrate that having a more expressive and flexible color mapping scheme like ours is pivotal in capturing accurate colors of the target DSLR. The qualitative results reiterate the trends noticed in the quantitative results presented in the manuscript. A simple feed forward network without a color prediction network (NoColorPred) produces less accurate colors since it does not inherently  capture many other factors like camera parameters and external environmental conditions that effect the color in an image. Incorporating a color prediction network in our DSLR sRGB restoration network provides us with a boost as seen in Fig.~\ref{fig:ablation_colorMap_supp}. Among the various alternatives that were tried, the CycleISP~\cite{cycleisp} inspired ColorBlur version fails to capture the sudden changes of color in the image contour and produces blurry results. On the other hand LinearMap computes a global color correction matrix which produces inaccurately colored images specially in terms of contrast due to its non-local addressing of the problem by LinearMap. 

Among the flexible parametric color mapping based versions of our color-mapping scheme $\mathcal{C}$ (Sec. ~\ref{sec:color_map} of the manuscript), the ConstValMap version that learns a fixed numeric value for each bin centroid is not powerful enough in terms of expressivity and having just 15 bins does not suffice for a reasonable performance. The accuracy in colors predicted by AffineDepMap in comparison to AffineIndepMap clearly demonstrates the the utility of exploiting the dependence between the color channels in an image for our color-mapping. Further, pre-processing the RAW (as discussed in Sec. ~\ref{sec:color_map} of the manuscript) aids our color mapping immensely by getting rid of the noise that is detrimental for color mapping. As seen in the results, our Color conditional RAW-to-sRGB pipeline aided by our color prediction module $\mathcal{G}$ achieves almost identical colors to the target DSLR sRGB  

\subsection{Loss}
\begin{figure*}[h]
\newcommand{\wid}{0.23\linewidth}
    \centering
    \begin{subfigure}[b]{\wid}
        \includegraphics[width=\textwidth]{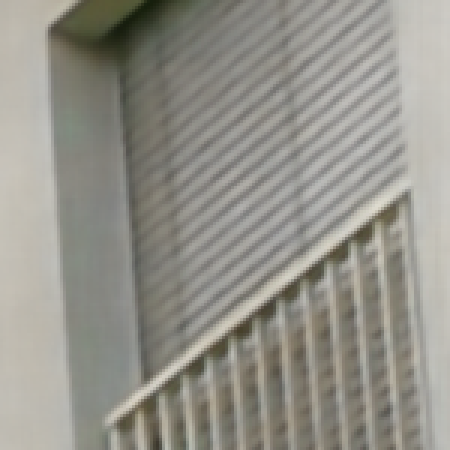}
        \includegraphics[width=\textwidth]{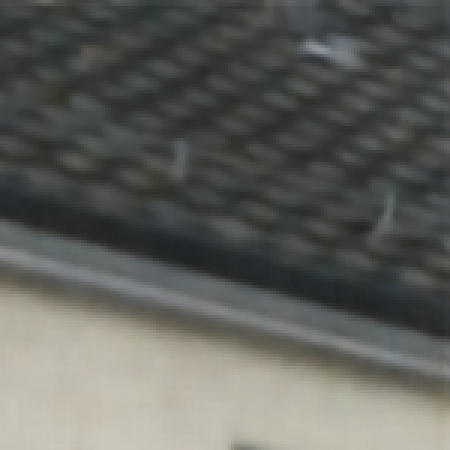}
        \caption{NoAlign}
        % \label{fig:res_mwnet2}
    \end{subfigure}
    \begin{subfigure}[b]{\wid}
        \includegraphics[width=\textwidth]{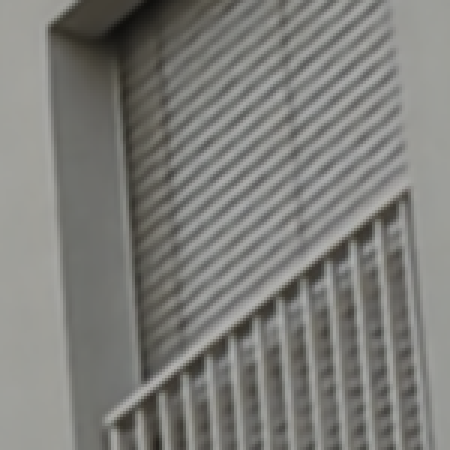}
        \includegraphics[width=\textwidth]{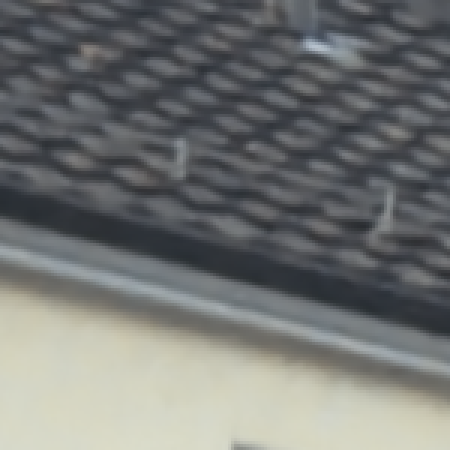}
        \caption{+AlignedLoss}
        % \label{fig:res_awnet12}
    \end{subfigure}
    \begin{subfigure}[b]{\wid}
        \includegraphics[width=\textwidth]{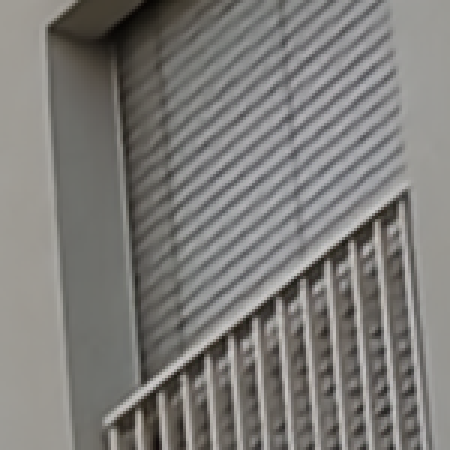}
        \includegraphics[width=\textwidth]{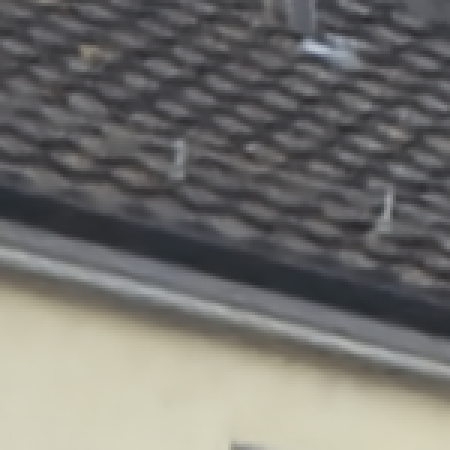}
        \caption{+Mask}
        % \label{fig:res_liteisp2}
    \end{subfigure}
    \begin{subfigure}[b]{\wid}
        \includegraphics[width=\textwidth]{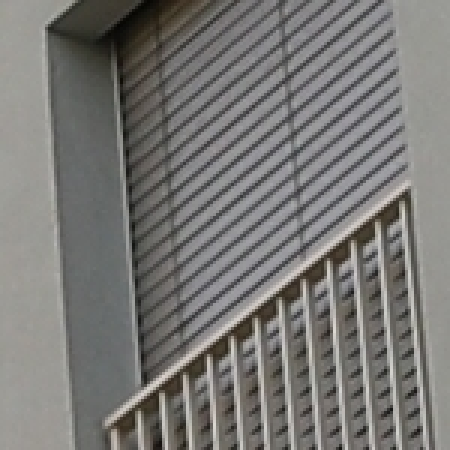}
        \includegraphics[width=\textwidth]{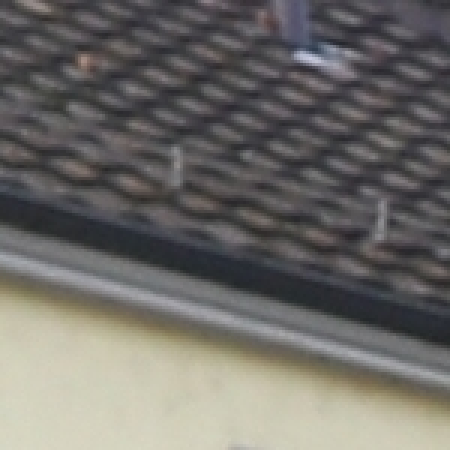}
        \caption{DSLR sRGB}
        %\label{fig:res_gt2}
    \end{subfigure}
    \caption{Qualitative results for the ablation of our robust masked loss (Sec. ~\ref{sec:loss} of the manuscript). These results demonstrate qualitatively our ablation study in section ~\ref{sec:ablation_loss} of the manuscript. The crops are taken from the ZRR dataset. Best viewed with zoom.}\label{fig:ablation_loss_supp}
\end{figure*}

Here, we show qualitatively the effectiveness of using a masked aligned loss for learning accurate RAW-to-sRGB mapping in the wild. Figure~\ref{fig:ablation_loss_supp} shows the visual results for the ablation study of our robust masked aligned loss (refer to Sec. 5.1.2 of the manuscript). The qualitative results show that computing a non-aligned loss (NoAlign) produces a blurry result due to the misalignment between the phone RAW and the corresponding DSLR sRGB during training. Further, aligning the RAW-sRGB pairs (+AlignedLoss) during training by explicit optical flow computations~\cite{pwcnet} improves the results but, the output during inference still remains blurry and is characterized by a noticeable color shift. This is due to the fact that we do not account for the inaccuracies in optical flow computations that may occur due to many reasons such as occlusions and inaccurate flows in homogeneous regions or regions with repeating patterns. To mitigate these inaccuracies in the optical flow computation, employing a forward-backward optical flow consistency  mask (Sec. ~\ref{sec:loss} of the manuscript) to our aligned loss (+Mask) produces a more detailed output with colors consistent with the target DSLR sRGB. This shows that accurate supervision using our masked loss during training provides immense gains to our DSLR sRGB restoration network.  

\subsection{Color Prediction}
\begin{figure*}[h]
\newcommand{\wid}{0.19\linewidth}
    \centering
    \begin{subfigure}[b]{\wid}
        \includegraphics[width=\textwidth]{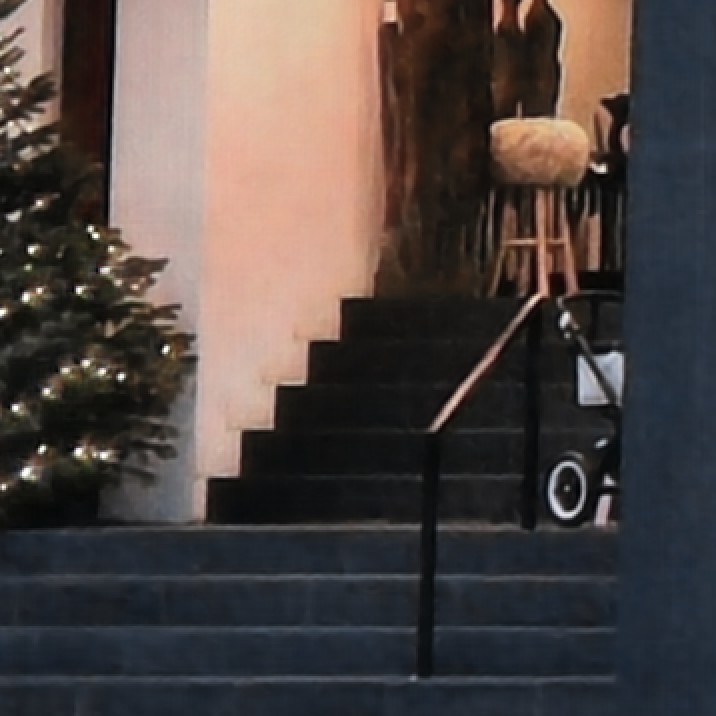}
        \includegraphics[width=\textwidth]{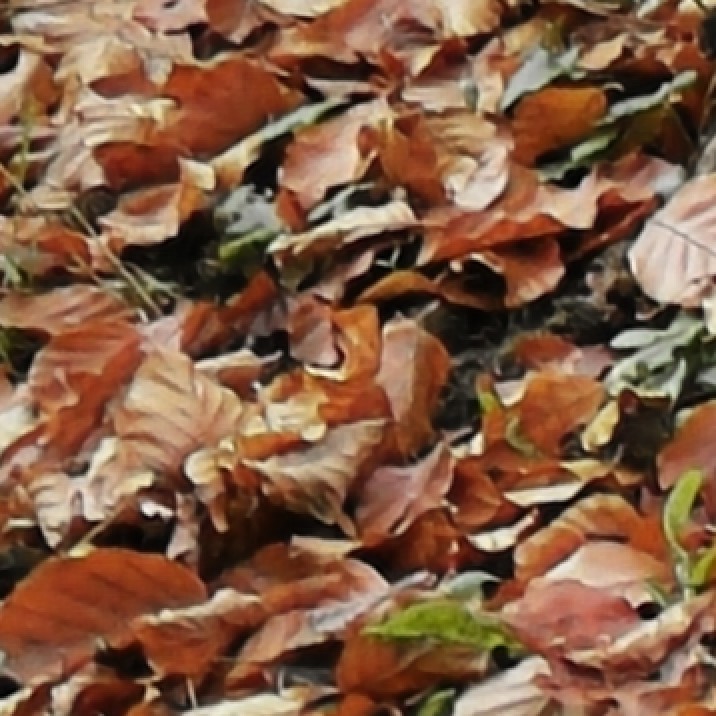}
        \caption{NoColor-\\Pred}
        % \label{fig:res_mwnet2}
    \end{subfigure}
    \begin{subfigure}[b]{\wid}
        \includegraphics[width=\textwidth]{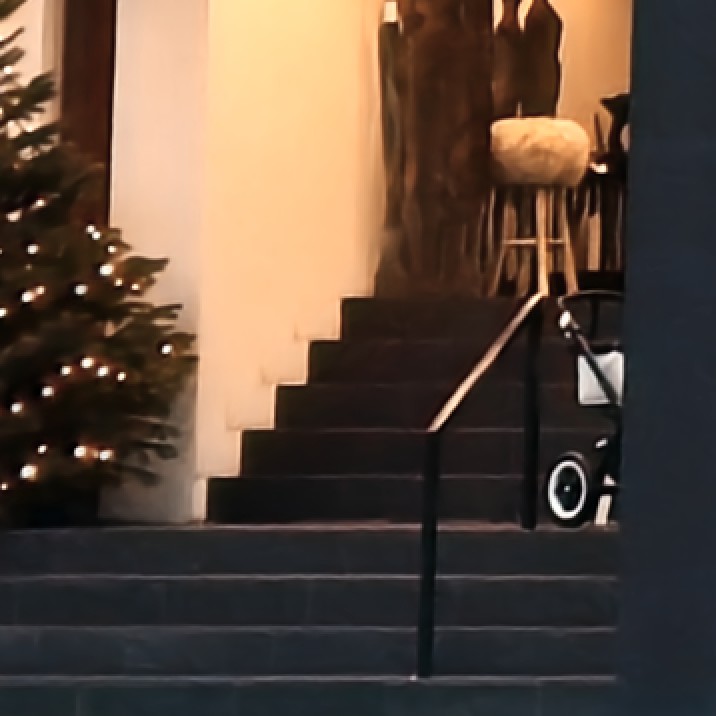}
        \includegraphics[width=\textwidth]{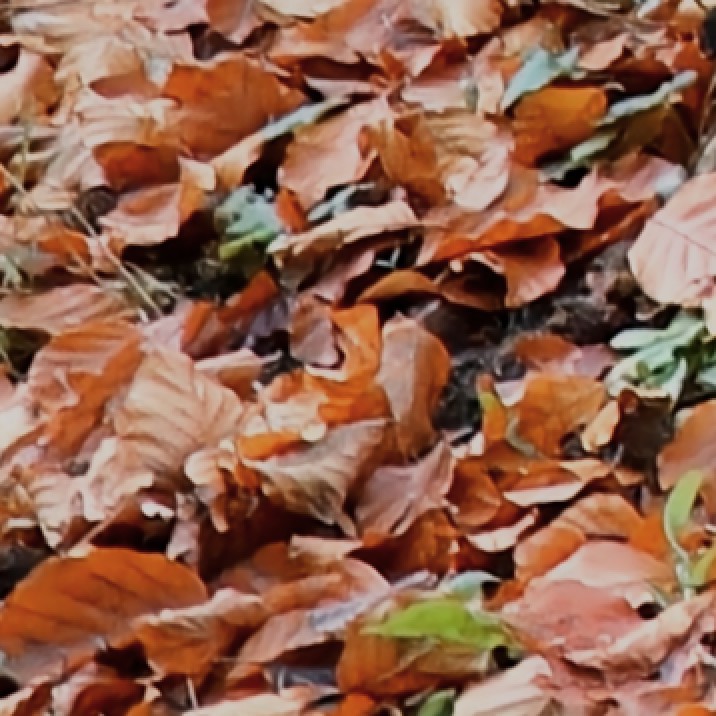}
        \caption{+U-Net\\\hspace{2mm}}
        % \label{fig:res_awnet12}
    \end{subfigure}
    \begin{subfigure}[b]{\wid}
        \includegraphics[width=\textwidth]{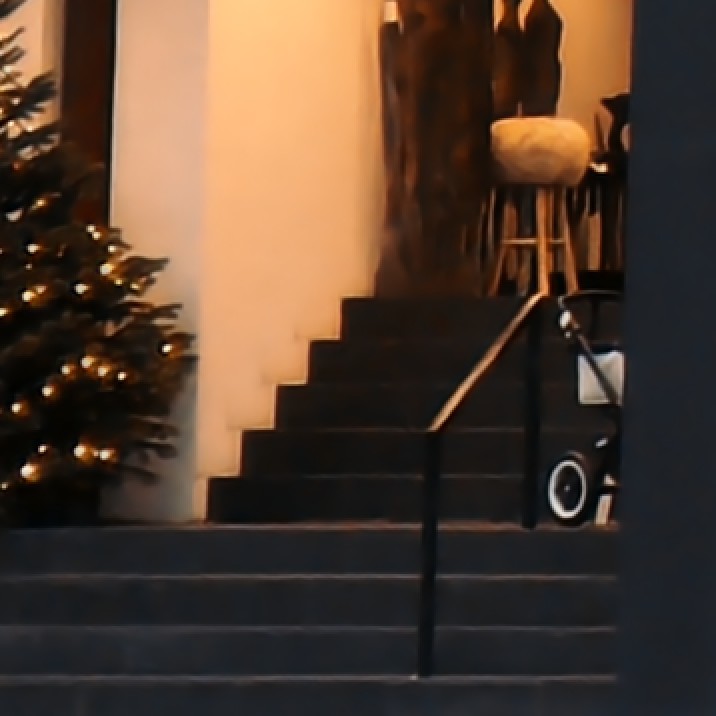}
        \includegraphics[width=\textwidth]{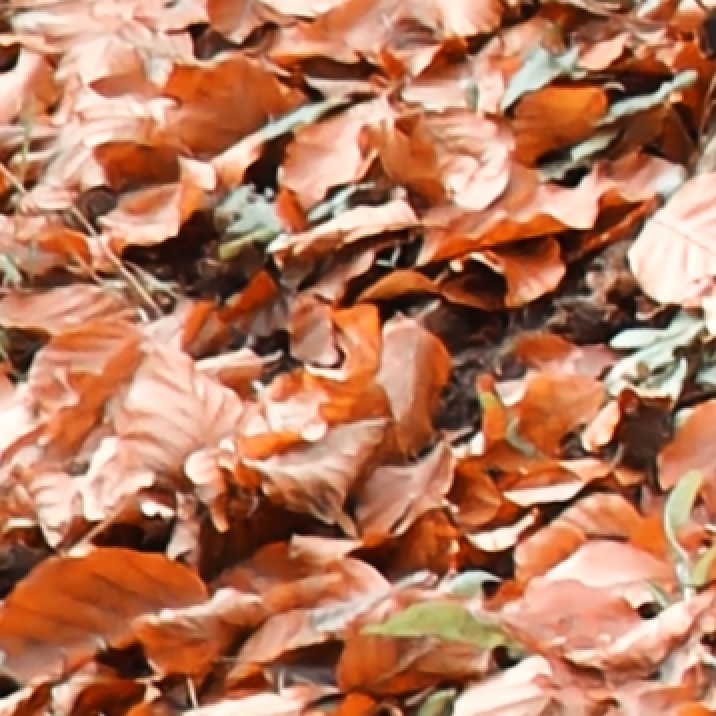}
        \caption{+Reconstruct}
        % \label{fig:res_liteisp2}
    \end{subfigure}
    \begin{subfigure}[b]{\wid}
        \includegraphics[width=\textwidth]{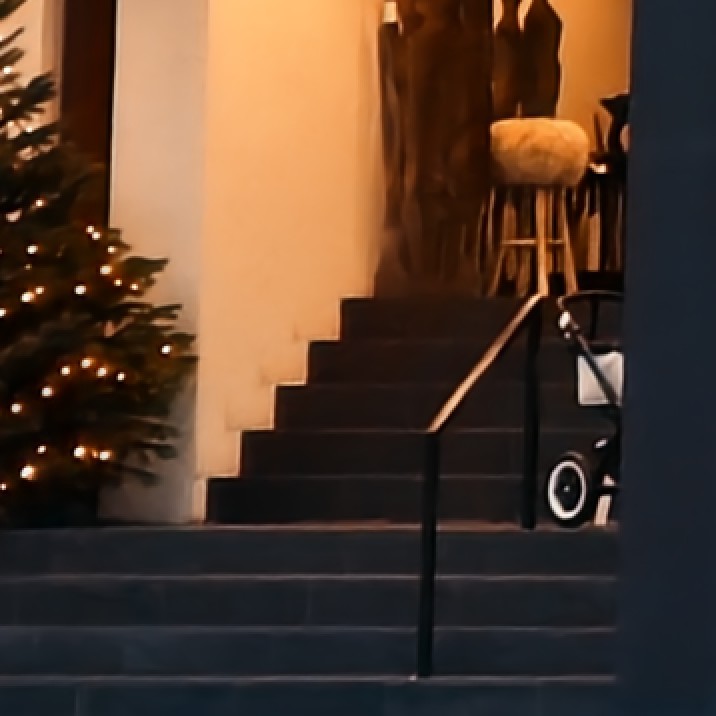}
        \includegraphics[width=\textwidth]{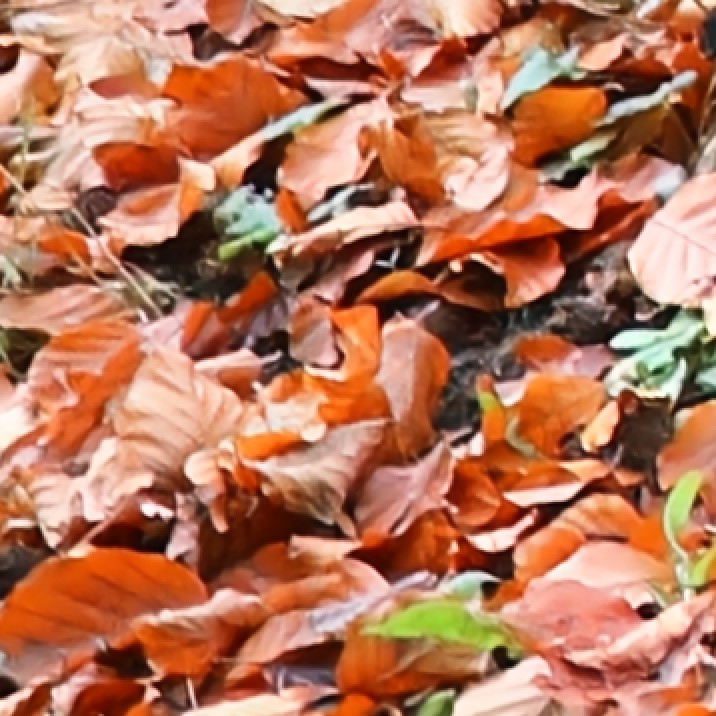}
        \caption{+GlobalContext}
        %\label{fig:res_ours2}
    \end{subfigure}
    \begin{subfigure}[b]{\wid}
        \includegraphics[width=\textwidth]{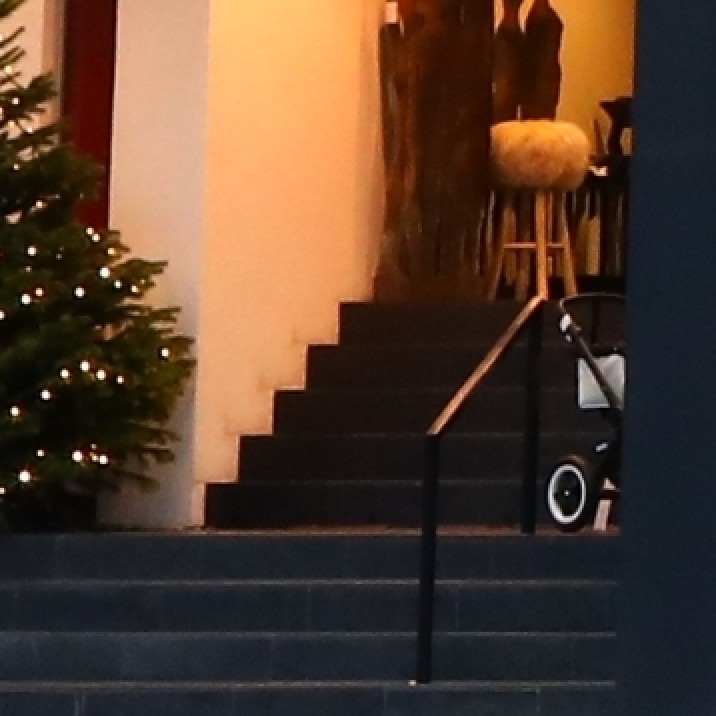}
        \includegraphics[width=\textwidth]{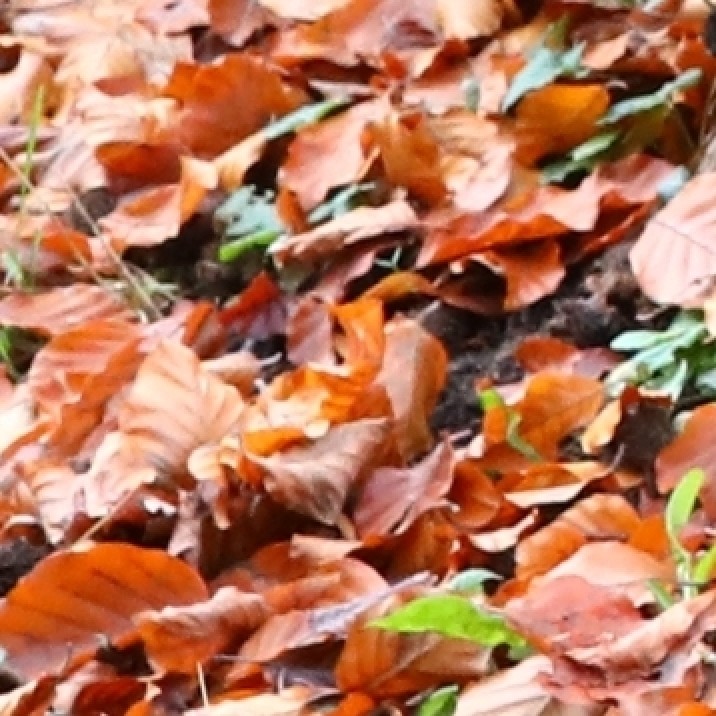}
        \caption{DSLR sRGB\\\hspace{2mm}}
        %\label{fig:res_gt2}
    \end{subfigure}
    \caption{Qualitative results for the ablation of our color prediction network (Sec. ~\ref{sec:colorinference} of the manuscript). These results demonstrate qualitatively our ablation study in section ~\ref{sec:ablation_color_pred} of the manuscript. The crops are taken from the ZRR dataset. Best viewed with zoom.}\label{fig:ablation_colorPred_supp}
\end{figure*}

In this section, we provide the the qualitative results for our color prediction network $\mathcal{G}$. Figure~\ref{fig:ablation_colorPred_supp} shows the qualitative results for the ablative study on our color prediction network. From Fig.~\ref{fig:ablation_colorPred_supp}, it becomes evident that conditioning RAW-to-sRGB pipeline on the color information (+U-Net) is pivotal for RAW-to-sRGB mapping in the wild. Introducing a reconstruction loss (+Reconstruct) on the reconstructed phone RAW, further improves the visual quality. Specifically, we notice that +Reconstruct accurately determines the lighting conditions (and other parameters on which the color in an image depends) at the time of capture. Thus, pointing to the utility of the reconstruction branch that helps our encoder in the color predictor module to encapsulate all the information into the encoding that is necessary for accurate color prediction. Finally, integrating our Global Context Block (+GlobalContext) outputs more coherent and consistent colors with the target DSLR sRGB. For the first example in Fig.~\ref{fig:ablation_colorPred_supp}, exploiting global cues helps our ISP Net to predict a sRGB image more consistent (see top right corner of the image) with the DSLR sRGB. And, in the second example the Global-Context transformer aids in predicting accurate colors for the green leaves in the image. Our final version produces colors almost identical to that of the target DSLR sRGB. 

\section{Results on Full Resolution Images}
\label{sec:full_res_results_supp}
\begin{figure*}[h]
\captionsetup[subfigure]{labelformat=empty}
\newcommand{\wid}{0.41\linewidth}
    \centering
    \begin{subfigure}[b]{\wid}
        \includegraphics[width=\textwidth]{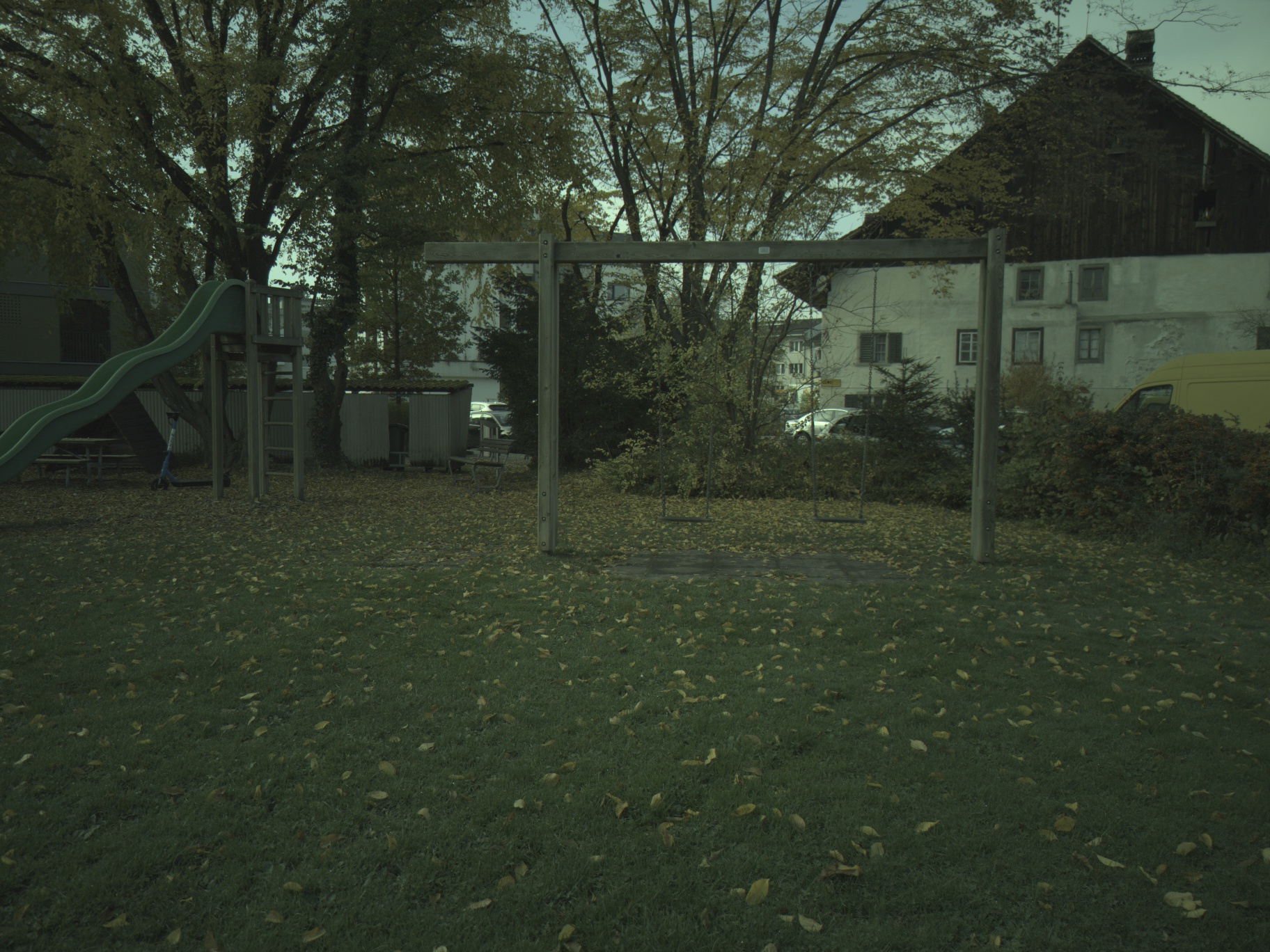}
        \caption{1a) RAW Visualized}
        % \label{fig:res_mwnet2}
    \end{subfigure}
    \begin{subfigure}[b]{\wid}
        \includegraphics[width=\textwidth]{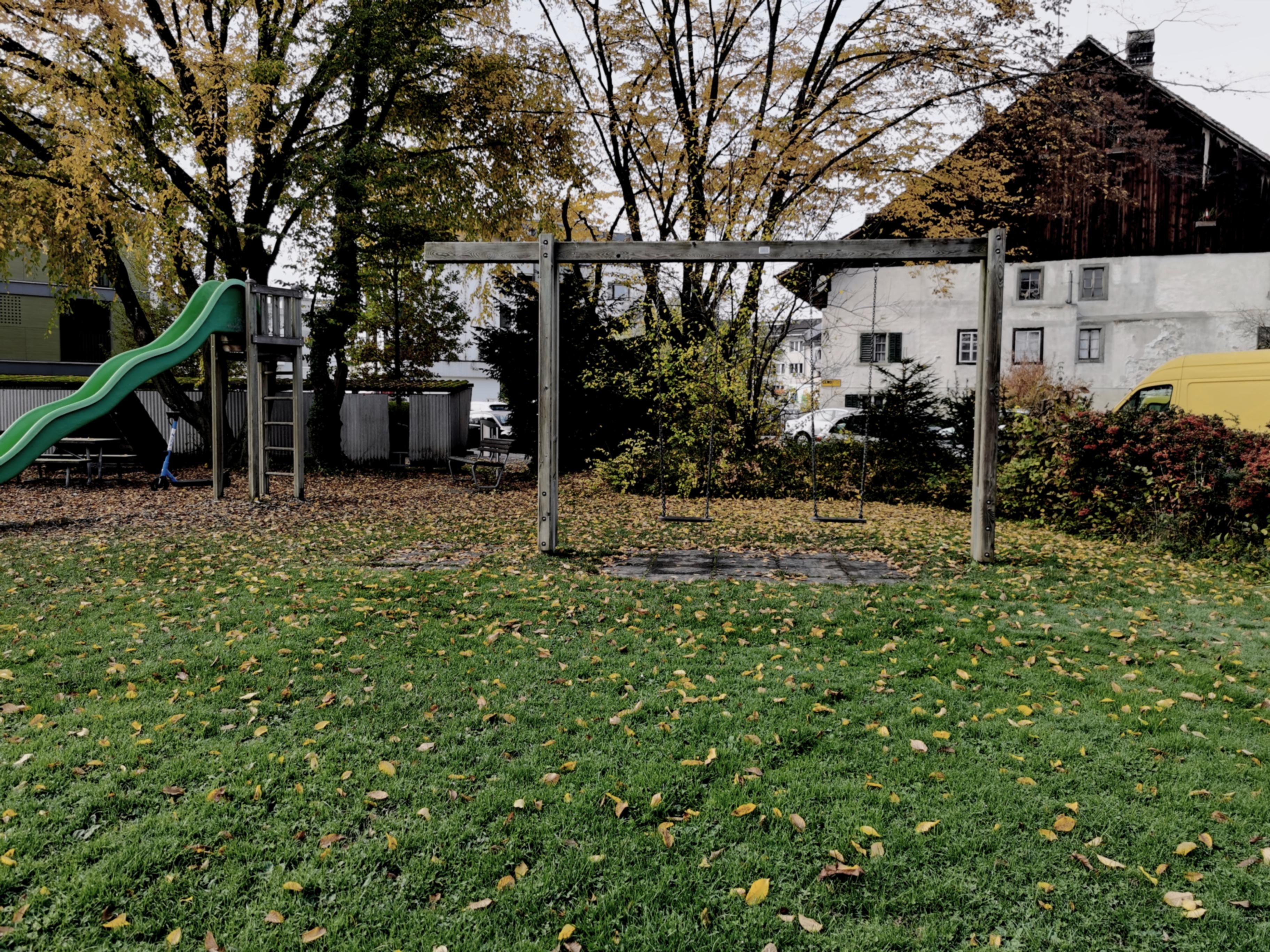}
        \caption{1b) LiteISPNet}
        % \label{fig:res_mwnet2}
    \end{subfigure}
    
    \begin{subfigure}[b]{\wid}
        \includegraphics[width=\textwidth]{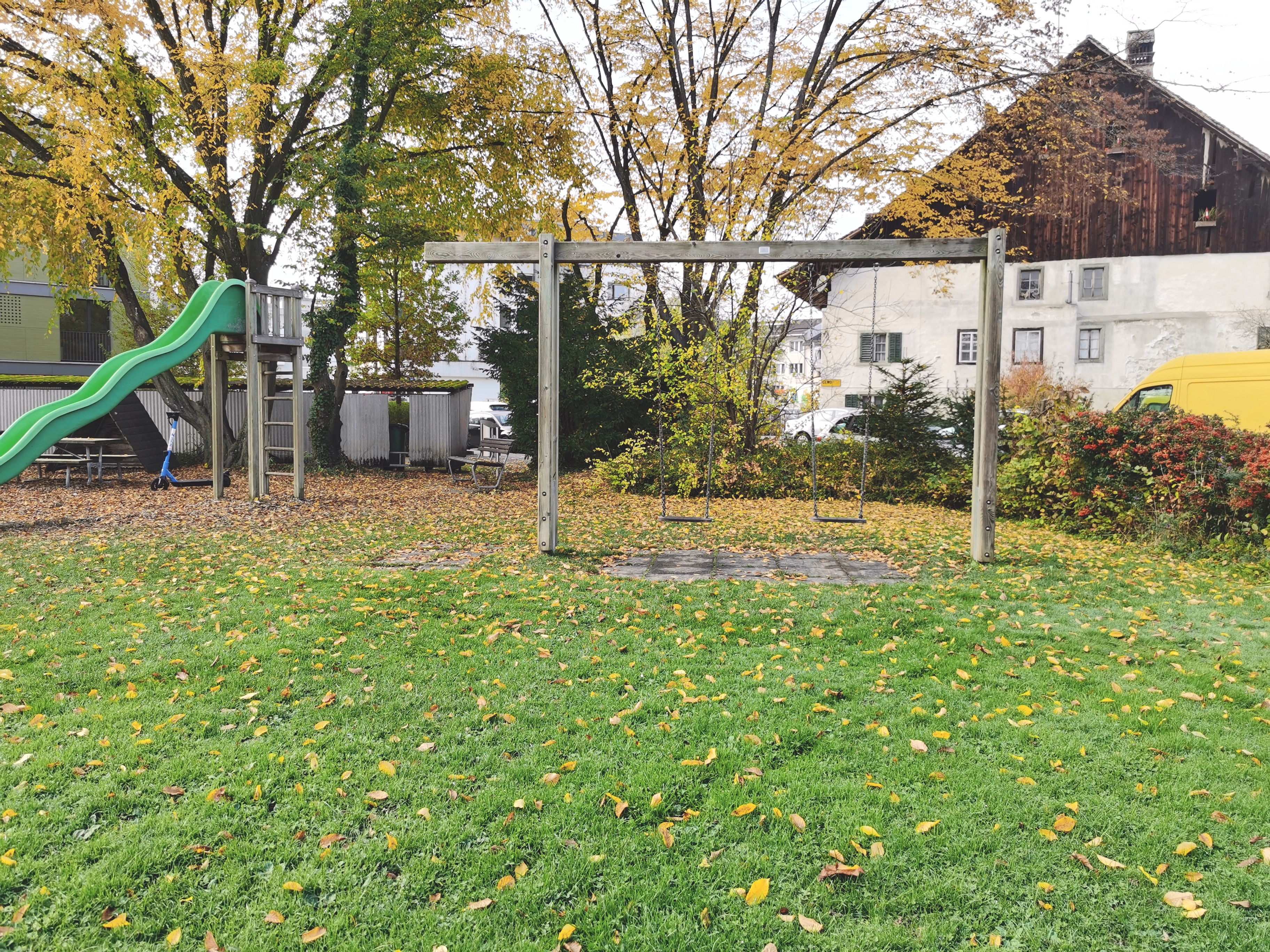}
        \caption{1c) Ours}
        % \label{fig:res_mwnet2}
    \end{subfigure}
    \begin{subfigure}[b]{\wid}
        \includegraphics[width=\textwidth]{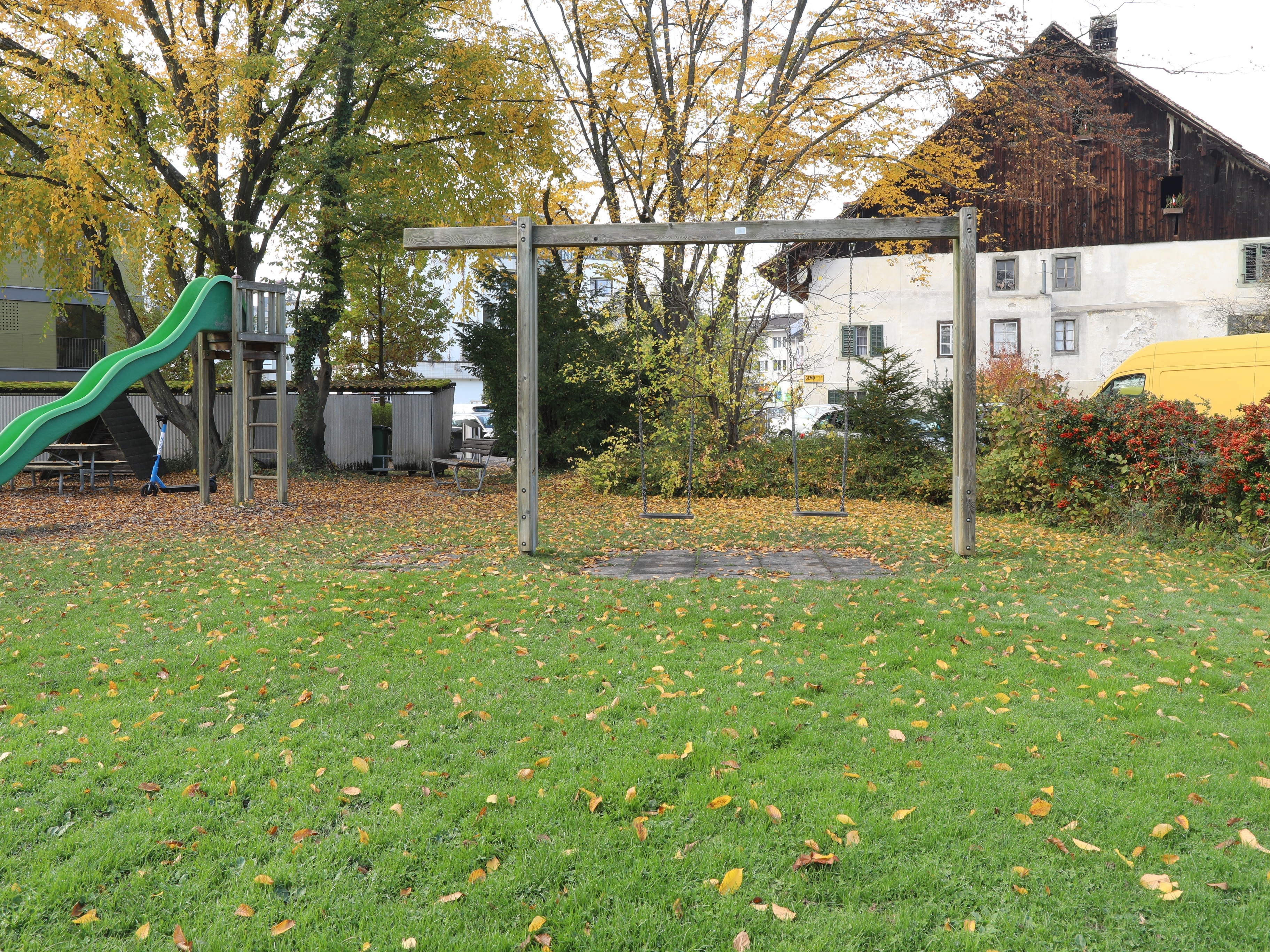}
        \caption{1d) DSLR sRGB}
        % \label{fig:res_mwnet2}
    \end{subfigure}
    
    \begin{subfigure}[b]{\wid}
        \includegraphics[width=\textwidth]{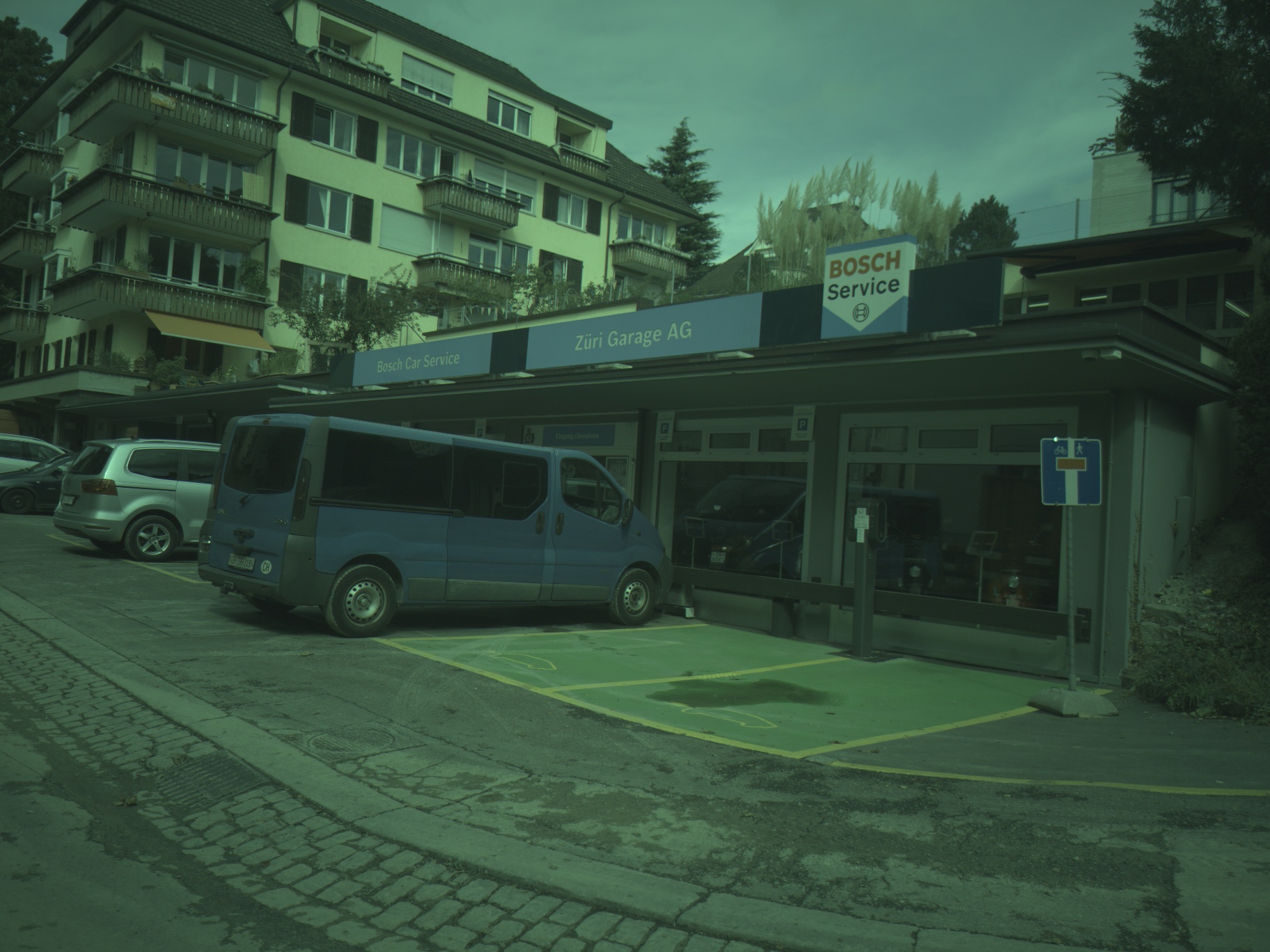}
        \caption{2a) RAW Visualized}
        % \label{fig:res_mwnet2}
    \end{subfigure}
    \begin{subfigure}[b]{\wid}
        \includegraphics[width=\textwidth]{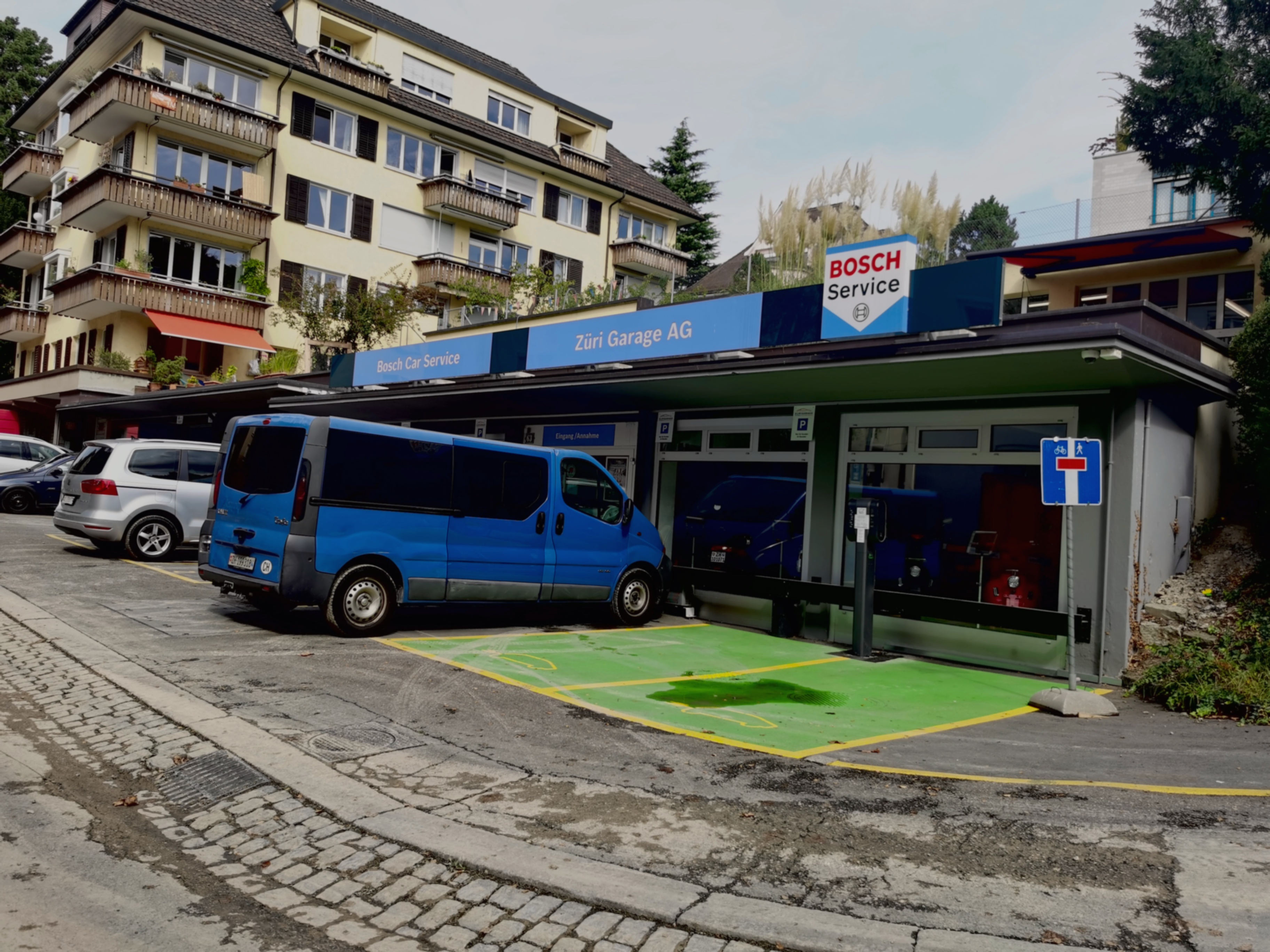}
        \caption{2b) LiteISPNet}
        % \label{fig:res_mwnet2}
    \end{subfigure}
    
    \begin{subfigure}[b]{\wid}
        \includegraphics[width=\textwidth]{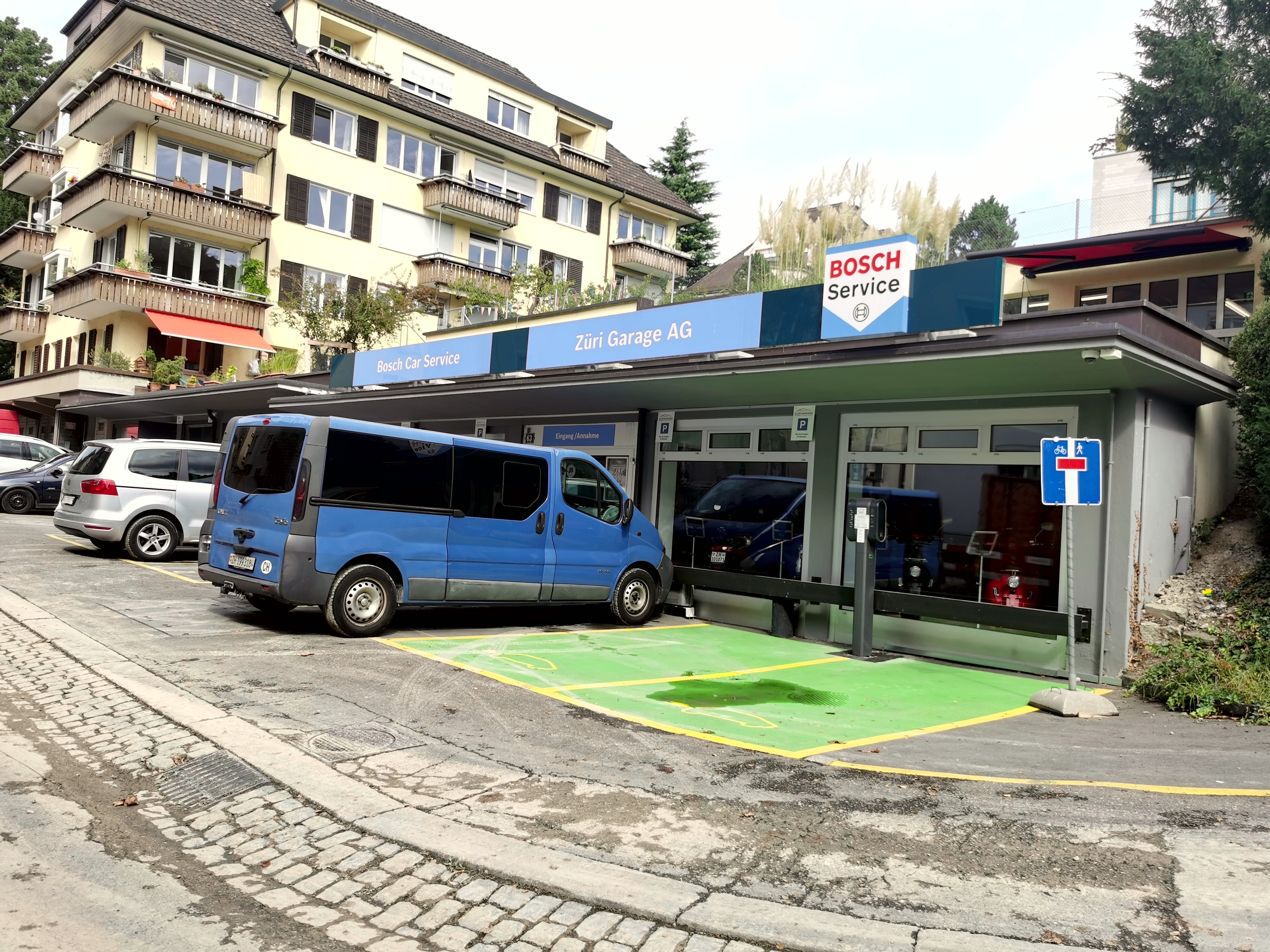}
        \caption{2c) Ours}
        % \label{fig:res_mwnet2}
    \end{subfigure}
    \begin{subfigure}[b]{\wid}
        \includegraphics[width=\textwidth]{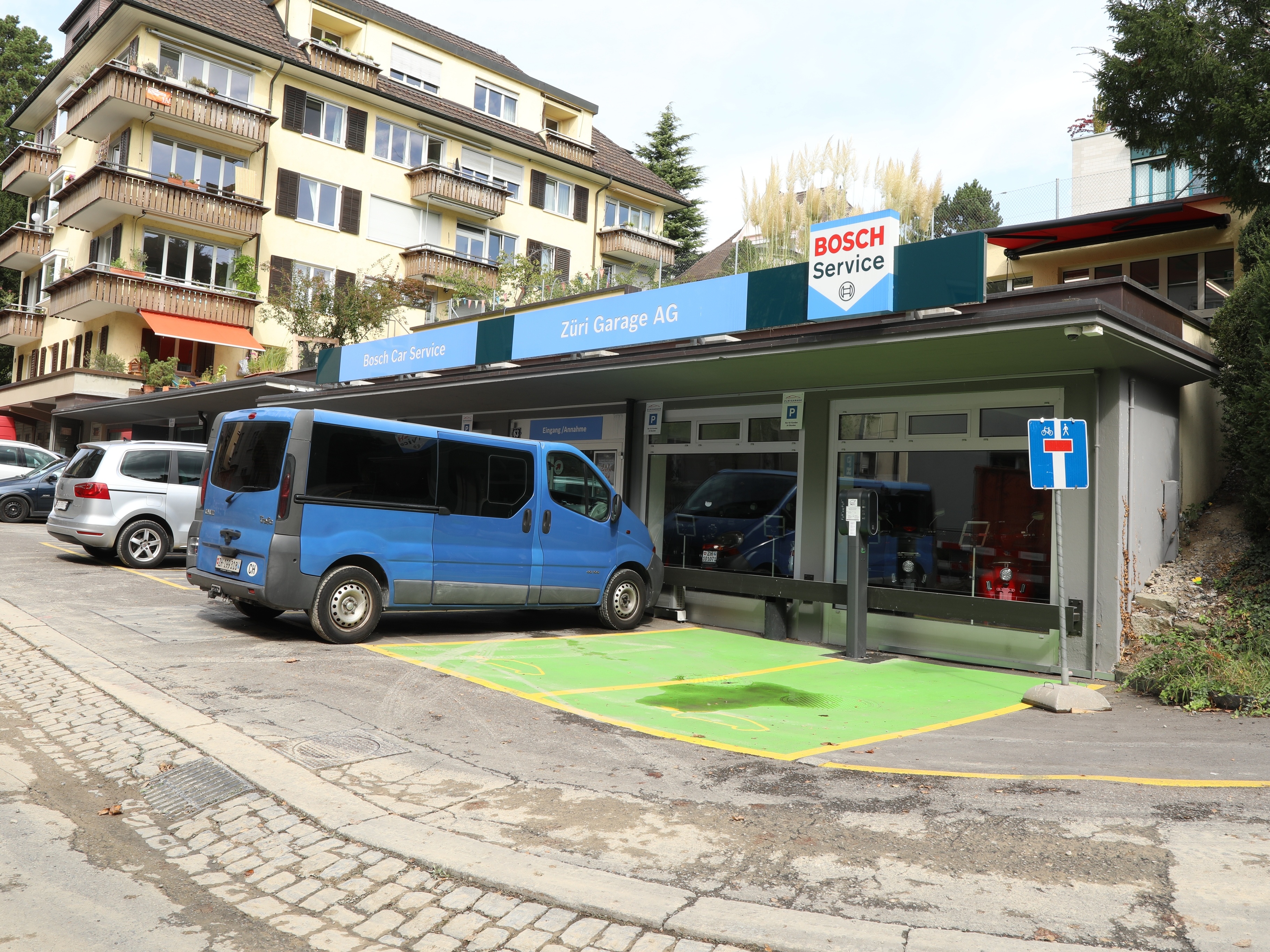}
        \caption{2d) DSLR sRGB}
        % \label{fig:res_mwnet2}
    \end{subfigure}
    
    % \begin{subfigure}[b]{\wid}
    %     \includegraphics[width=\textwidth]{./images/supp/full_res/2/raw_2.jpg}
    %     \caption{2a) RAW Visualized}
    %     % \label{fig:res_mwnet2}
    % \end{subfigure}
    % \begin{subfigure}[b]{\wid}
    %     \includegraphics[width=\textwidth]{./images/supp/full_res/2/litenet_2.jpg}
    %     \caption{2b) LiteISPNet}
    %     % \label{fig:res_mwnet2}
    % \end{subfigure}
    
    % \begin{subfigure}[b]{\wid}
    %     \includegraphics[width=\textwidth]{./images/supp/full_res/2/ours_2.jpg}
    %     \caption{2c) Ours}
    %     % \label{fig:res_mwnet2}
    % \end{subfigure}
    % \begin{subfigure}[b]{\wid}
    %     \includegraphics[width=\textwidth]{./images/supp/full_res/2/dslr_2.jpg}
    %     \caption{2d) DSLR sRGB}
    %     % \label{fig:res_mwnet2}
    % \end{subfigure}
\caption{Full resolution results on our ISPW dataset. We compare our method against the best performing competing method LiteISPNet~\cite{liteispnet}. Our approach captures more details and more accurate colors w.r.t. the DSLR sRGB. On the other hand,  LiteISPNet produces dull colors and results in loss of detail. Best viewed with zoom.}\label{fig:full_res_supp}
\end{figure*}

In this section, we present full resolution results for our approach. Fig. ~\ref{fig:full_res_supp} shows the full resolution (2736x3648) predictions of our approach on the ISPW dataset. Our approach produces accurate globally coherent colors w.r.t. the DSLR sRGB. On the other hand, LiteISPNet~\cite{liteispnet} produces dull inaccurate colors. Thus, underlining the utility of leveraging global context by our color prediction network. Importantly, our efficient fixed size latent-array based global attention aids in applying our models on large images since the computational complexity of our Global Context Transformer layer scales linearly with the image size. Additionally, LiteISPNet results in loss of detail compared to the DSLR quality sRGB images produced by our approach. This shows the effectiveness of employing a masked aligned loss during training.    

% \begin{figure*}[t!]\ContinuedFloat
% \captionsetup[subfigure]{labelformat=empty}
% \newcommand{\wid}{0.41\linewidth}
% \centering
% \begin{subfigure}[b]{\wid}
%         \includegraphics[width=\textwidth]{./images/supp/full_res/3/raw_2.jpg}
%         \caption{3a) RAW Visualized}
%         % \label{fig:res_mwnet2}
%     \end{subfigure}
%     \begin{subfigure}[b]{\wid}
%         \includegraphics[width=\textwidth]{./images/supp/full_res/3/litenet_2.jpg}
%         \caption{3b) LiteISPNet}
%         % \label{fig:res_mwnet2}
%     \end{subfigure}
    
%     \begin{subfigure}[b]{\wid}
%         \includegraphics[width=\textwidth]{./images/supp/full_res/3/ours_2.jpg}
%         \caption{3c) Ours}
%         % \label{fig:res_mwnet2}
%     \end{subfigure}
%     \begin{subfigure}[b]{\wid}
%         \includegraphics[width=\textwidth]{./images/supp/full_res/3/dslr_2.jpg}
%         \caption{3d) DSLR sRGB}
%         % \label{fig:res_mwnet2}
%     \end{subfigure}
%     \caption{Full resolution results on our ISPW dataset. We compare our method against the best performing competing method LiteISPNet~\cite{liteispnet}. Our approach captures more details and more accurate colors w.r.t. the DSLR sRGB. On the other hand,  LiteISPNet produces dull colors and results in loss of detail. Best viewed with zoom.}\label{fig:full_res_supp}
% \end{figure*}

\section{State-of-the-Art Results}
\label{sec:sota_supp}
\begin{figure*}[t]
\newcommand{\wid}{0.19\linewidth}
    \centering
    \begin{subfigure}[b]{\wid}
        \includegraphics[width=\textwidth]{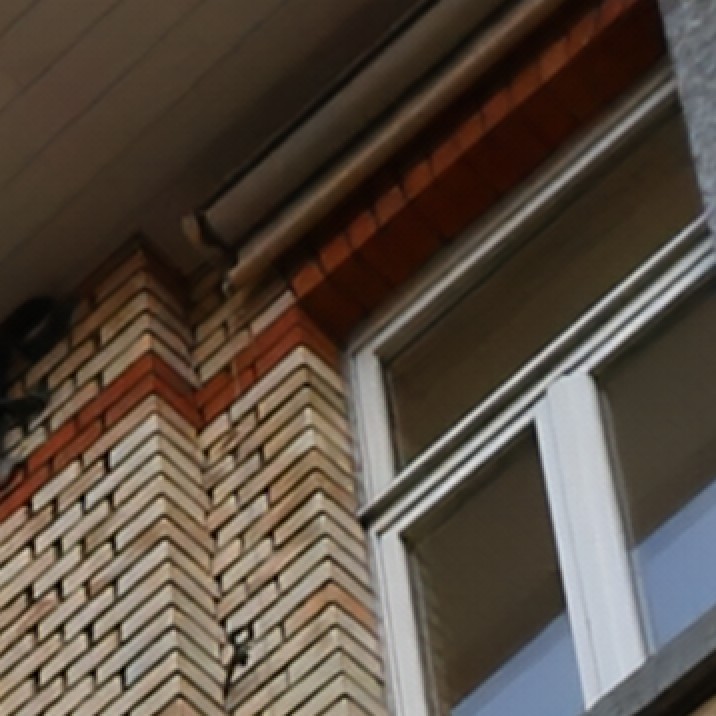}
        \includegraphics[width=\textwidth]{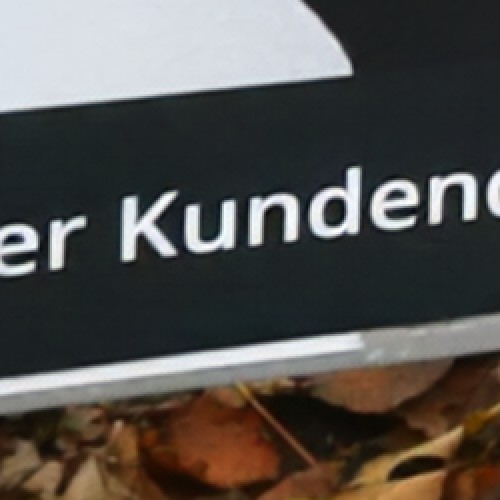}
        \includegraphics[width=\textwidth]{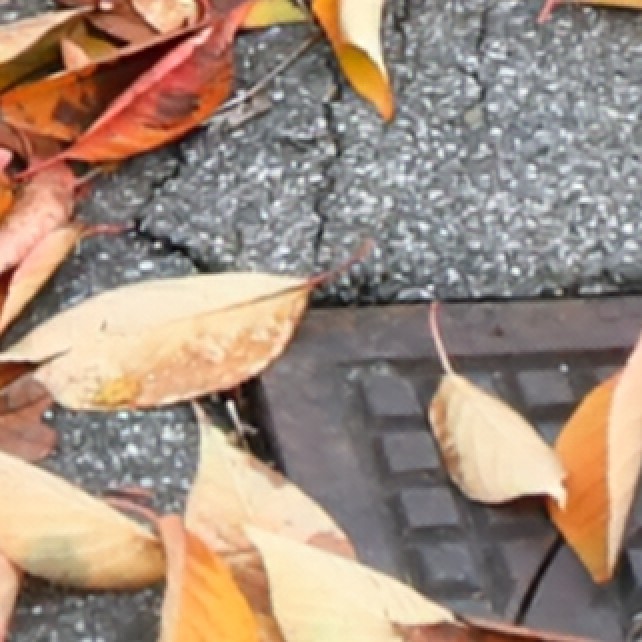}
        \includegraphics[width=\textwidth]{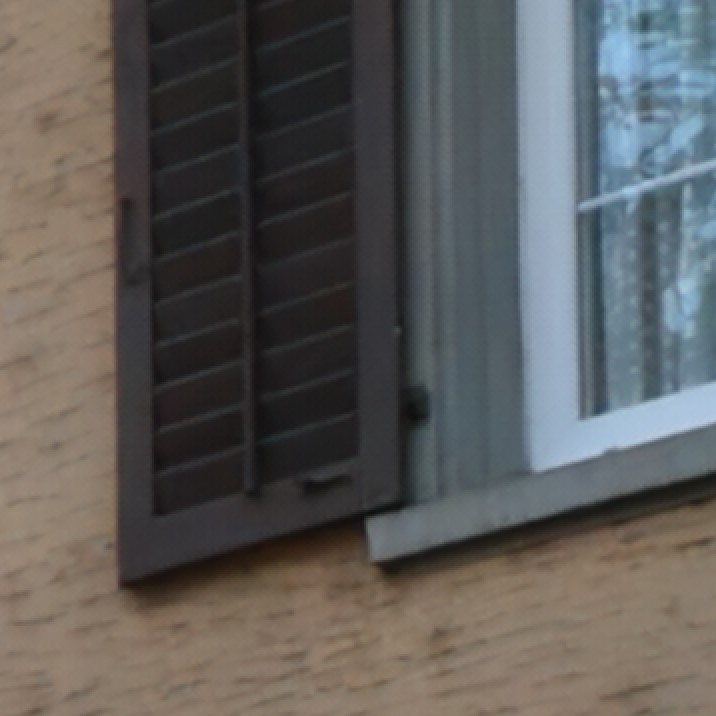}
        \caption{MWISPNet}
        % \label{fig:res_mwnet2}
    \end{subfigure}
    \begin{subfigure}[b]{\wid}
        \includegraphics[width=\textwidth]{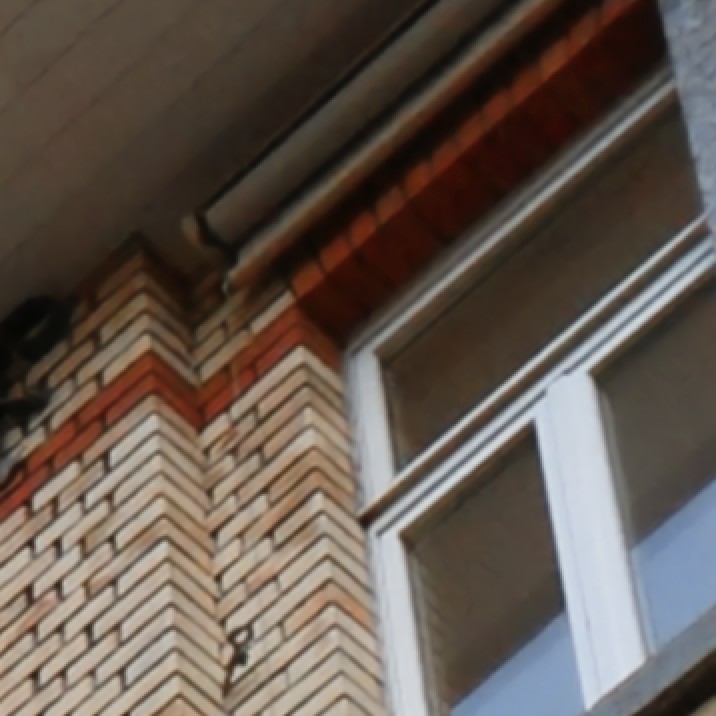}
        \includegraphics[width=\textwidth]{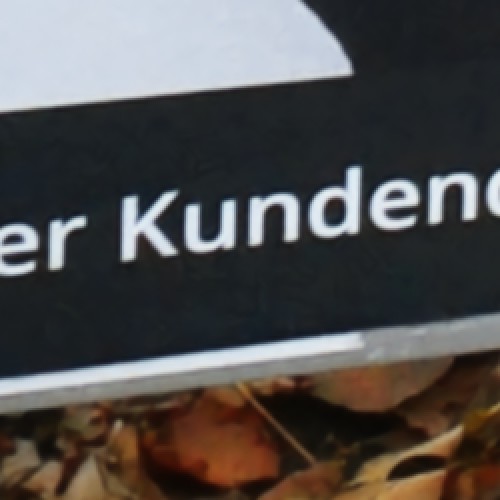}
        \includegraphics[width=\textwidth]{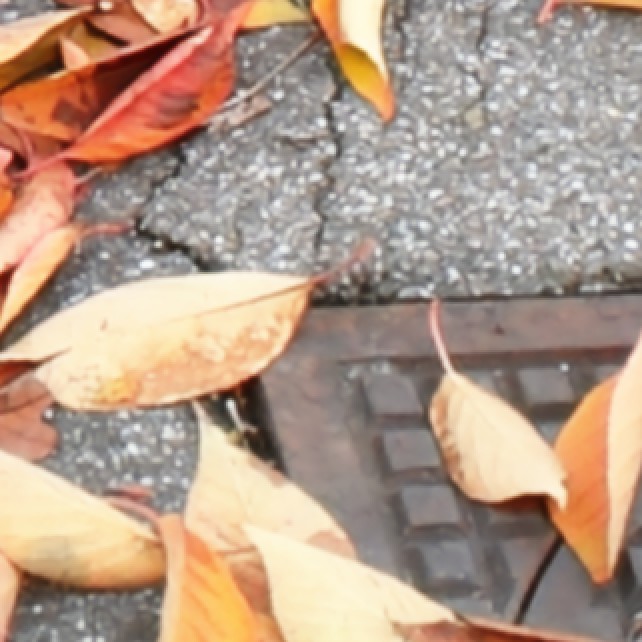}
        \includegraphics[width=\textwidth]{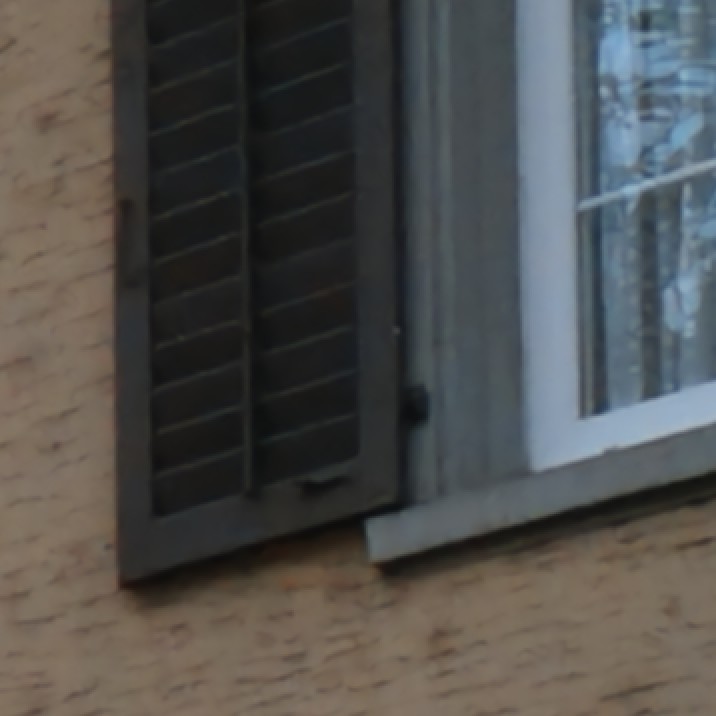}
        \caption{AWNet}
        % \label{fig:res_awnet12}
    \end{subfigure}
    \begin{subfigure}[b]{\wid}
        \includegraphics[width=\textwidth]{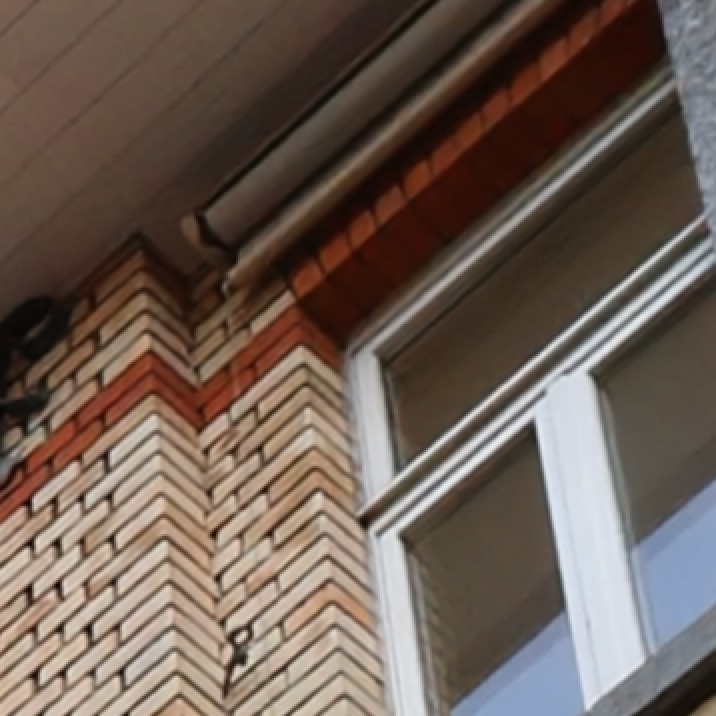}
        \includegraphics[width=\textwidth]{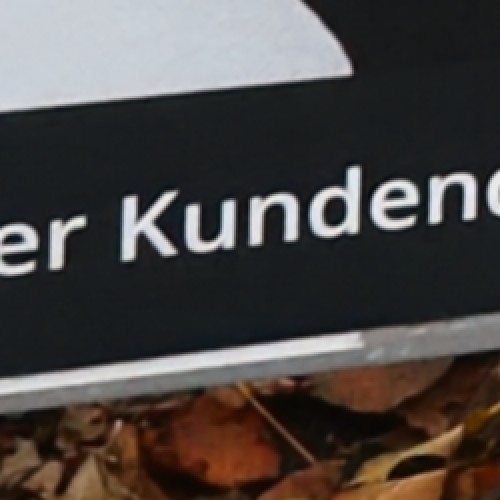}
        \includegraphics[width=\textwidth]{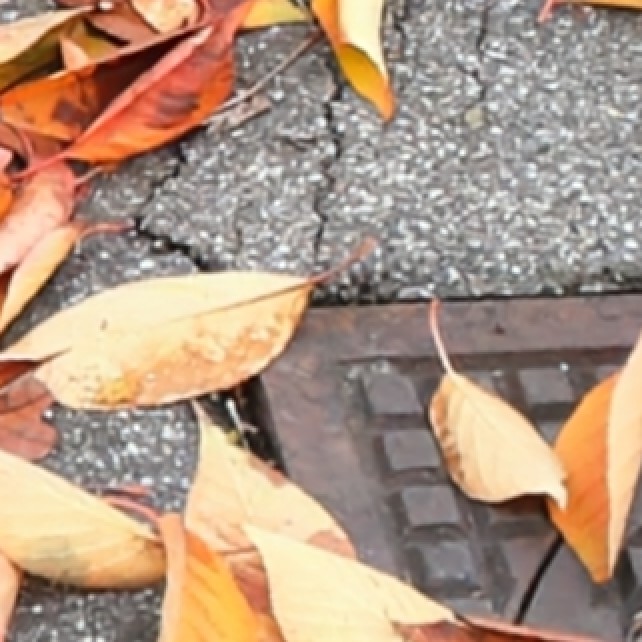}
        \includegraphics[width=\textwidth]{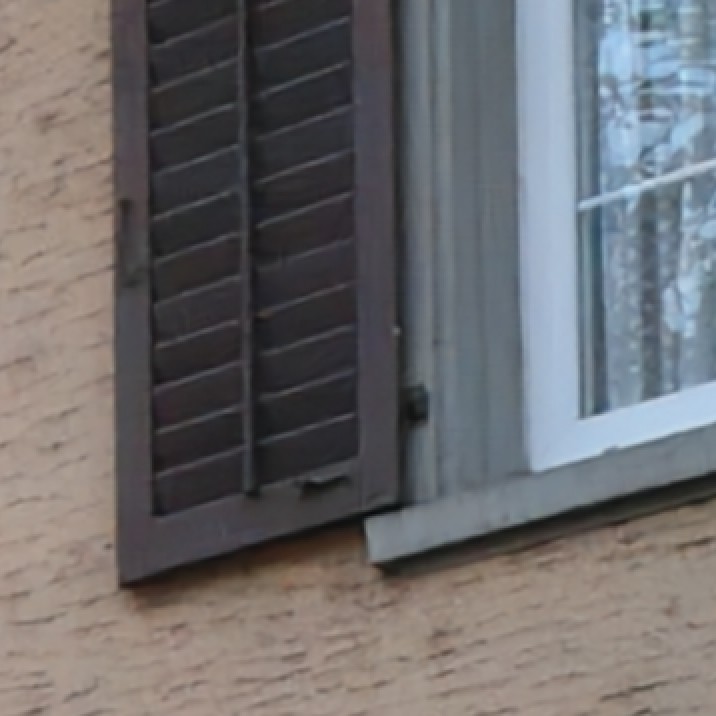}
        \caption{LiteISPNet}
        % \label{fig:res_liteisp2}
    \end{subfigure}
    \begin{subfigure}[b]{\wid}
        \includegraphics[width=\textwidth]{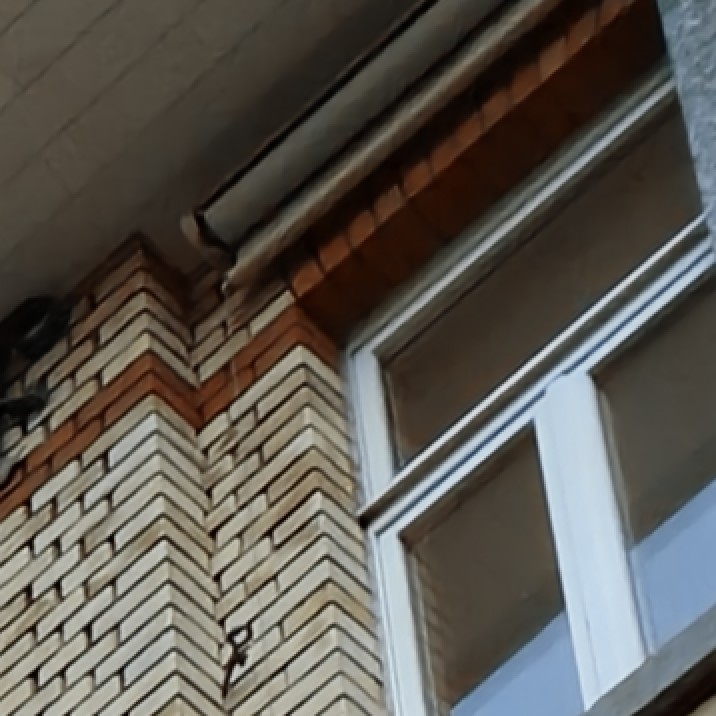}
        \includegraphics[width=\textwidth]{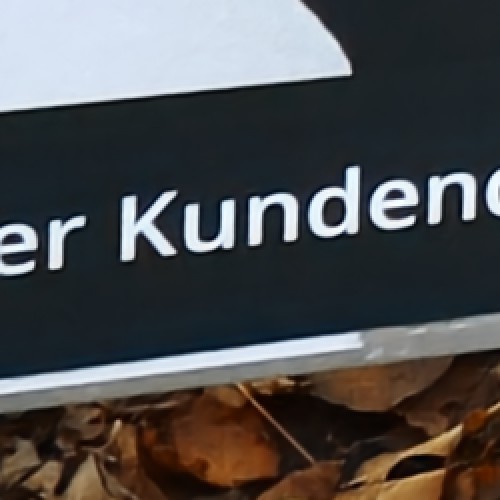}
        \includegraphics[width=\textwidth]{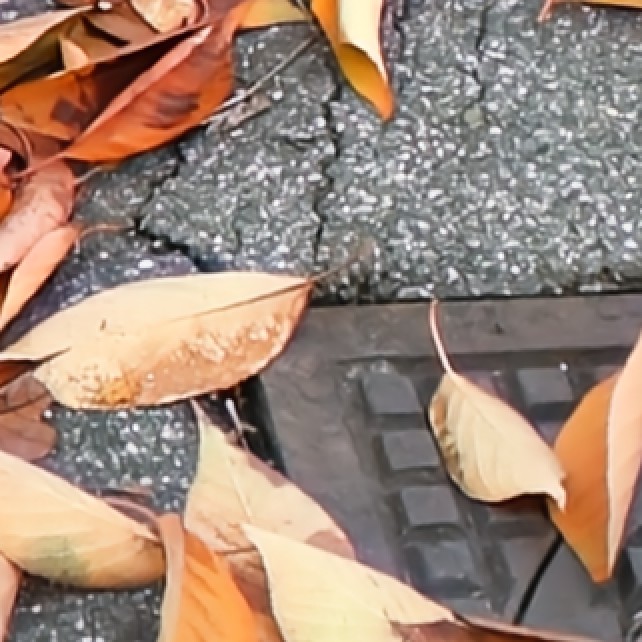}
        \includegraphics[width=\textwidth]{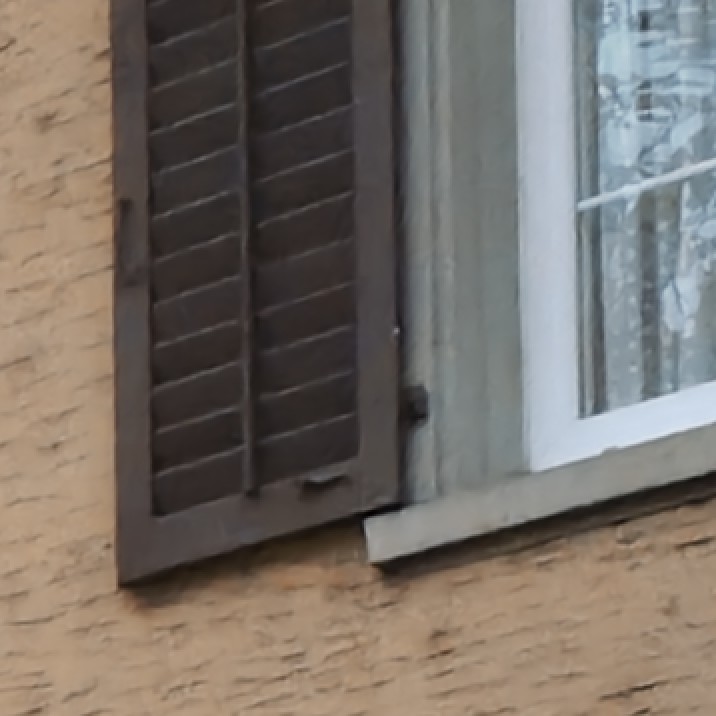}
        \caption{Ours}
        %\label{fig:res_ours2}
    \end{subfigure}
    \begin{subfigure}[b]{\wid}
        \includegraphics[width=\textwidth]{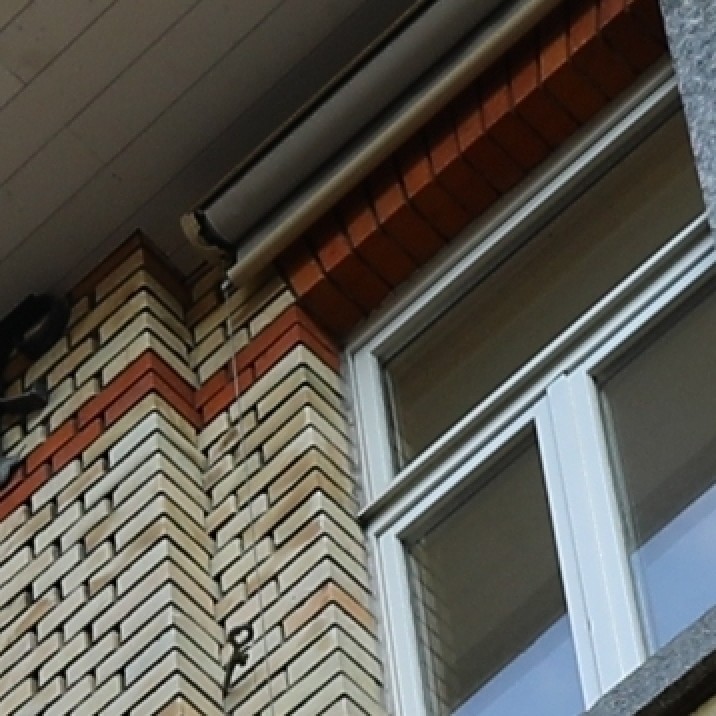}
        \includegraphics[width=\textwidth]{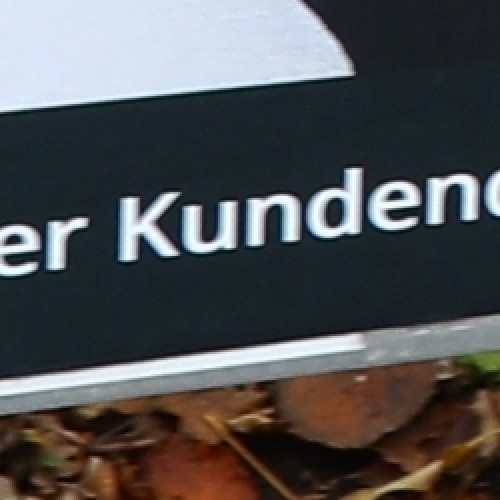}
        \includegraphics[width=\textwidth]{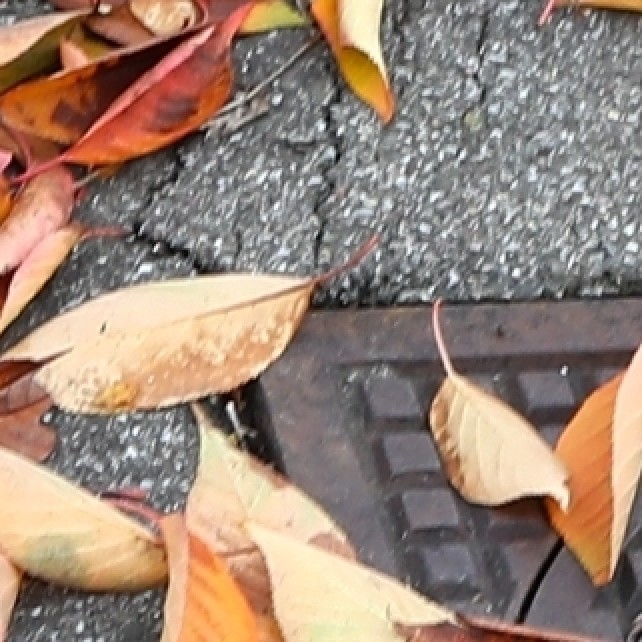}
        \includegraphics[width=\textwidth]{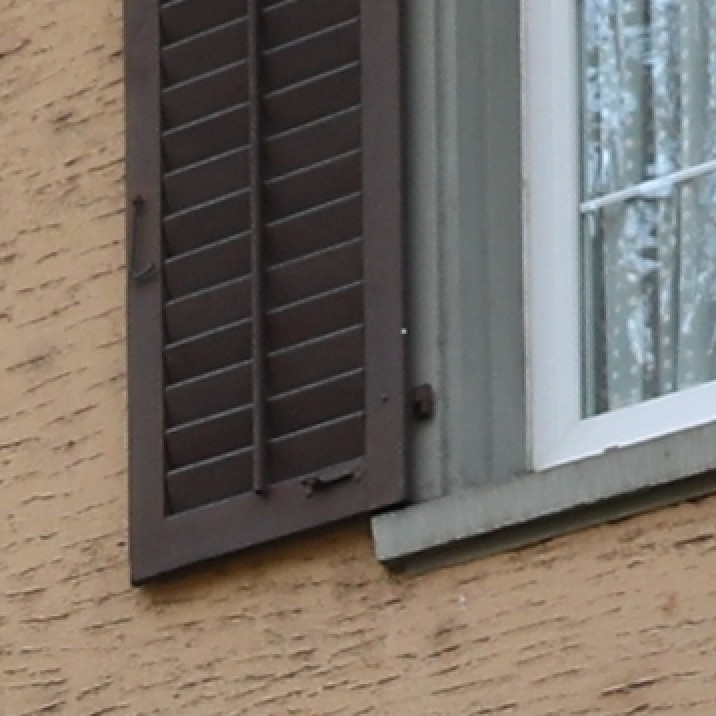}
        \caption{DSLR sRGB}
        %\label{fig:res_gt2}
    \end{subfigure}
    \caption{Some more visual results for state-of-the-art comparison on the ZRR~\cite{ZRR} dataset. Best viewed with zoom.}
    \label{fig:sota_zrr_supp}
\end{figure*}

\begin{figure*}[t]
\newcommand{\wid}{0.19\linewidth}
    \centering
    \begin{subfigure}[b]{\wid}
        \includegraphics[width=\textwidth]{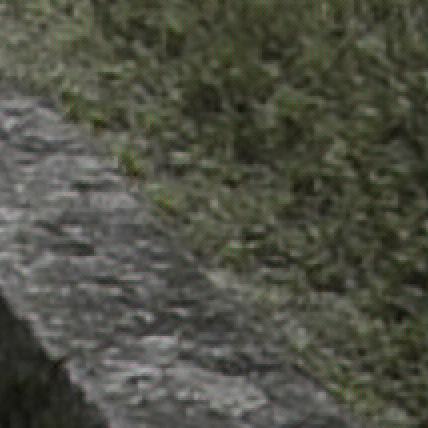}
        \includegraphics[width=\textwidth]{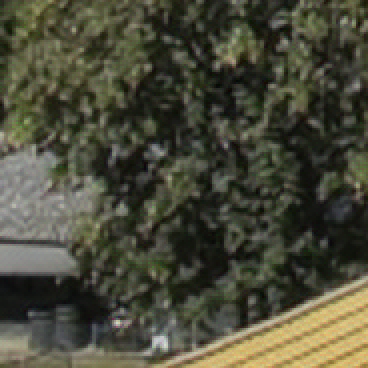}
        \includegraphics[width=\textwidth]{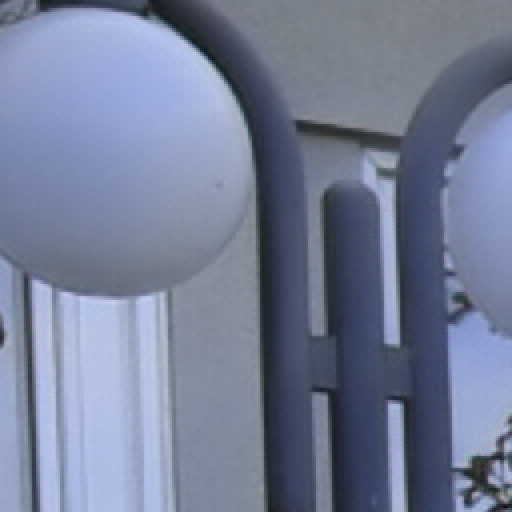}
        \includegraphics[width=\textwidth]{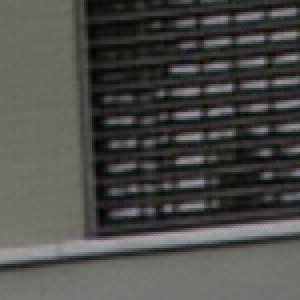}
        \includegraphics[width=\textwidth]{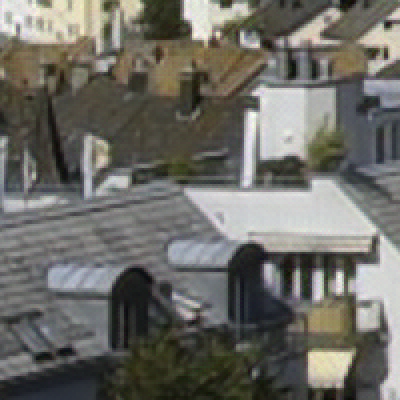}
        \caption{MWISPNet}
        % \label{fig:res_mwnet2}
    \end{subfigure}
    \begin{subfigure}[b]{\wid}
        \includegraphics[width=\textwidth]{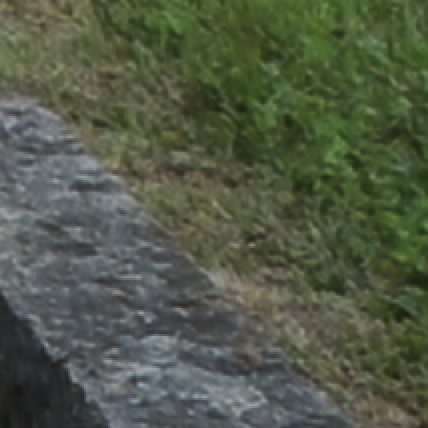}
        \includegraphics[width=\textwidth]{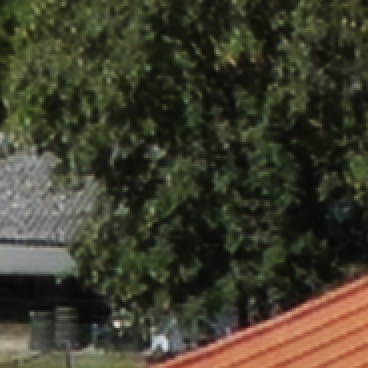}
        \includegraphics[width=\textwidth]{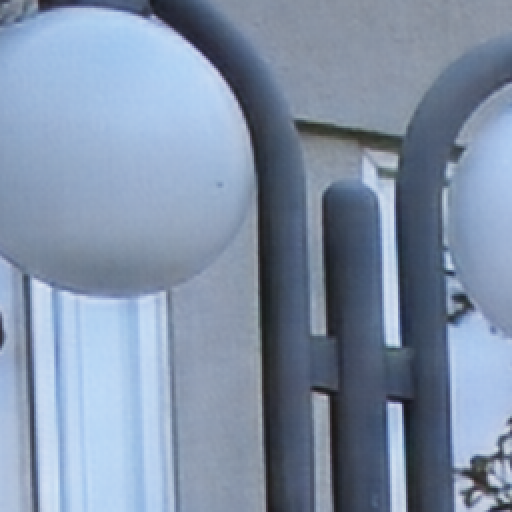}
        \includegraphics[width=\textwidth]{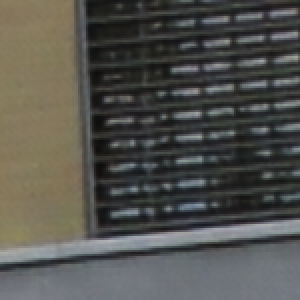}
        \includegraphics[width=\textwidth]{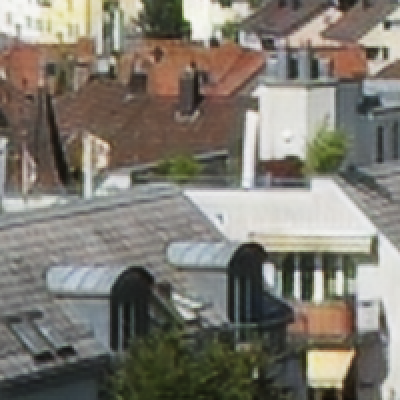}
        \caption{AWNet}
        % \label{fig:res_awnet12}
    \end{subfigure}
    \begin{subfigure}[b]{\wid}
        \includegraphics[width=\textwidth]{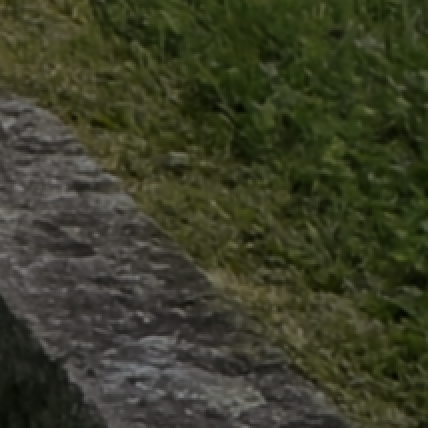}
        \includegraphics[width=\textwidth]{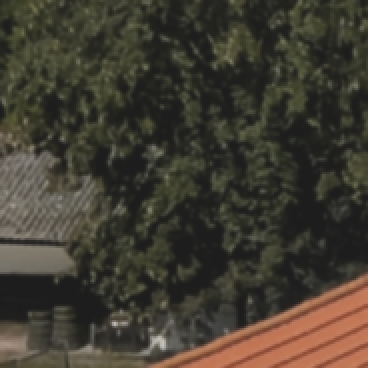}
        \includegraphics[width=\textwidth]{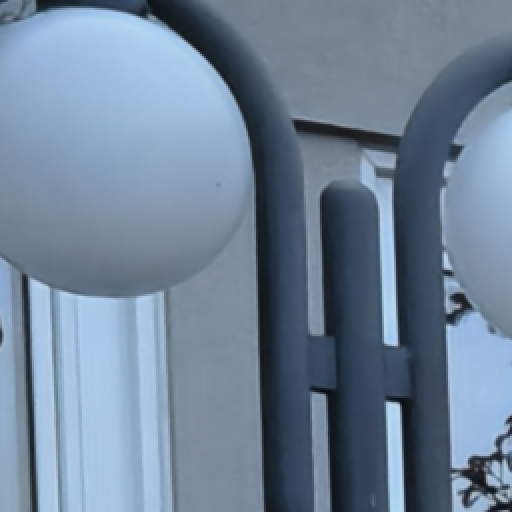}
        \includegraphics[width=\textwidth]{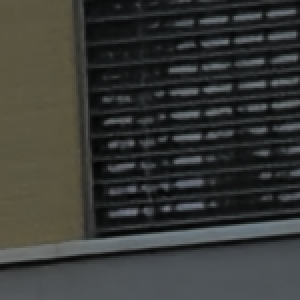}
        \includegraphics[width=\textwidth]{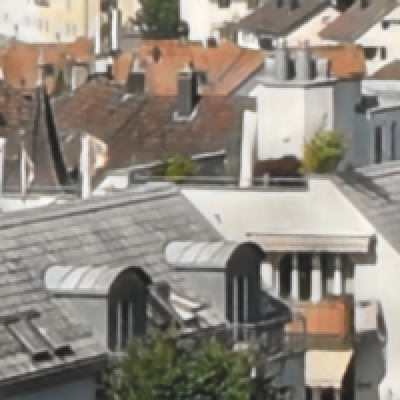}
        \caption{LiteISPNet}
        % \label{fig:res_liteisp2}
    \end{subfigure}
    \begin{subfigure}[b]{\wid}
        \includegraphics[width=\textwidth]{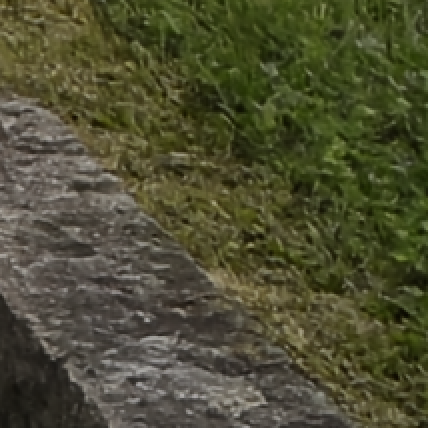}
        \includegraphics[width=\textwidth]{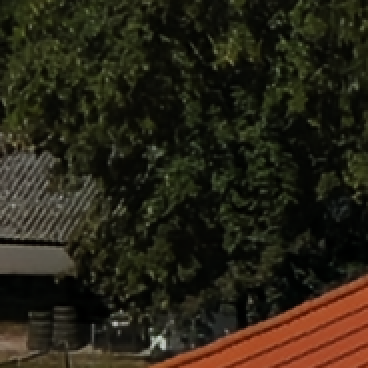}
        \includegraphics[width=\textwidth]{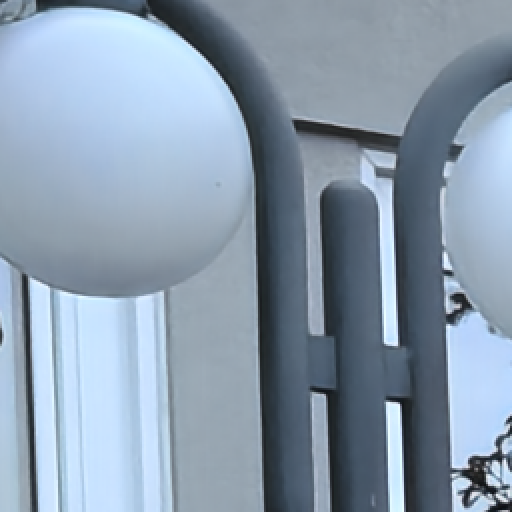}
        \includegraphics[width=\textwidth]{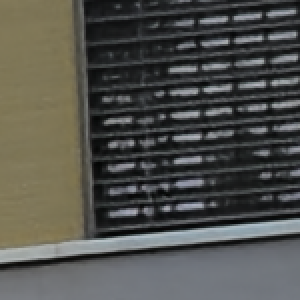}
        \includegraphics[width=\textwidth]{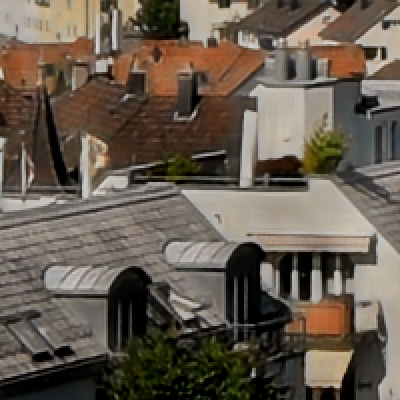}
        \caption{Ours}
        %\label{fig:res_ours2}
    \end{subfigure}
    \begin{subfigure}[b]{\wid}
        \includegraphics[width=\textwidth]{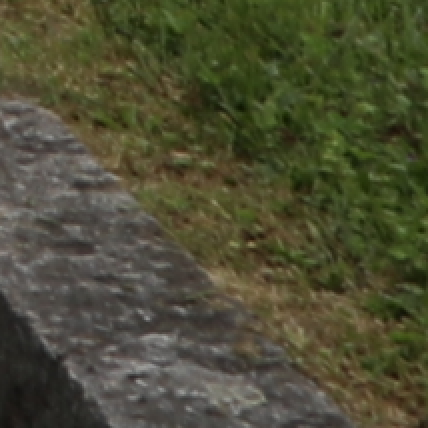}
        \includegraphics[width=\textwidth]{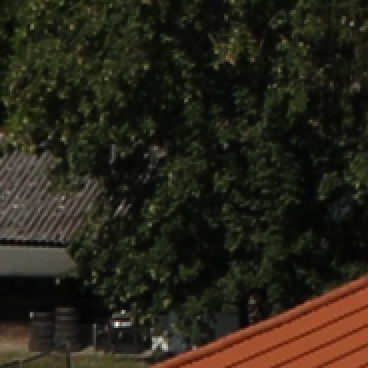}
        \includegraphics[width=\textwidth]{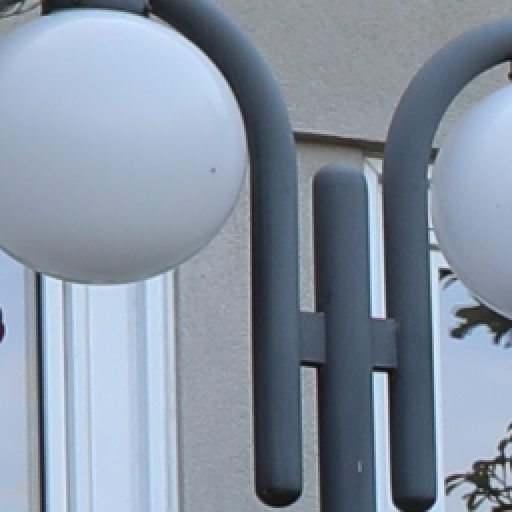}
        \includegraphics[width=\textwidth]{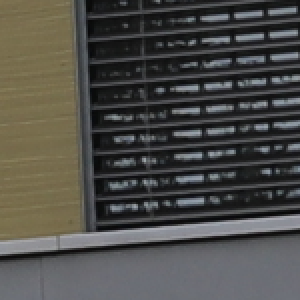}
        \includegraphics[width=\textwidth]{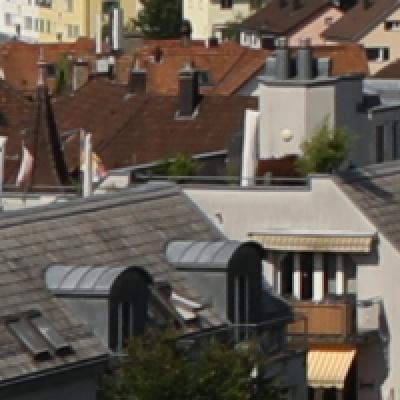}
        \caption{DSLR sRGB}
        %\label{fig:res_gt2}
    \end{subfigure}
    \caption{Some more visual results for state-of-the-art comparison on our ISPW dataset. Best viewed with zoom.}\label{fig:sota_mrr_supp}
\end{figure*}

In this section, we exhibit our results qualitatively in comparison to other existing methods on the test sets of ZRR dataset~\cite{ZRR} and our ISPW dataset. Figures ~\ref{fig:sota_zrr_supp} and ~\ref{fig:sota_mrr_supp} show the state-of-the-art comparison of our approach with other existing approaches on the ZRR and the ISPW datasets, respectively. The visual results clearly show the supremacy of our method in comparison to previous methods. In particular MW-ISPNet~\cite{ZRR} and AWNet~\cite{awnet} produce blurry results hence demonstrating their ineffectiveness in handling misalignment between the phone RAW and the DSLR sRGB pairs during training. The effect is more adverse in case of the ISPW dataset where the degree of the aforementioned pairwise misalignment is worse as compared to the ZRR dataset. Further, the LiteISPNet~\cite{liteispnet} uses an aligned loss for learning a mapping between the phone RAW and the DSLR sRGB. Though, this reduces the blur (does not completely get rid of it) in the results as in previously mentioned methods, it lacks detail and suffers a significant color shift. Our approach on the other hand leapfrogs LiteISPNet significantly by providing very crisp results capturing rich details and accurate colors. This is clearly evident from our visual results. Further, in Fig. ~\ref{fig:sota_mrr_supp} we also show the results from the phone ISP. We notice that in many cases our results are richer in detail as compared to the target DSLR sRGB and the resulting sRGB from the phone ISP. This underlines the effectiveness of our approach for RAW-to-sRGB mapping in the wild. 

%%%%%%%%%%%%%%%%%%%%%%%%%%%%%%%%%%%%%%%%%%%%%%
%%%%%%%%%%%%%%%%%%%%%%%%%%%%%%%%%%%%%%%%%%%%%%%

\section{ISP in the Wild (ISPW) dataset}
\label{sec:dataset_supp}
\begin{figure}[h]
\centering
\includegraphics[width=0.8\columnwidth]{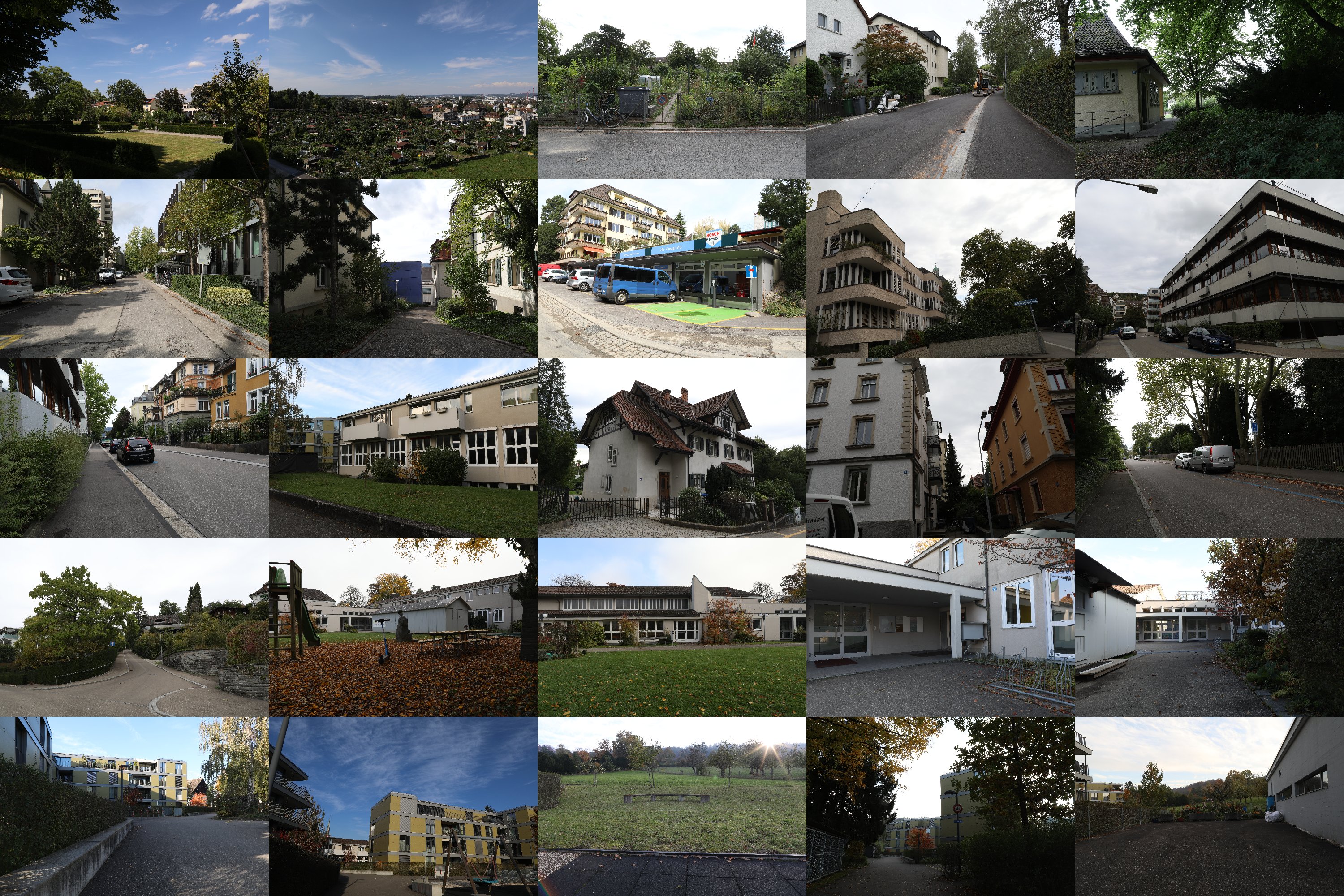}
\caption{Example captures from our ISPW dataset. We show some example captures from the DSLR camera. As demonstrated, the ISPW dataset is collected in various lighting and weather conditions which makes it a very challenging dataset for learning and benchmarking the full ISP pipeline in the wild.}
\label{fig:collage_supp}
\end{figure}

Here, we demonstrate a few example images captured in our ISP in the Wild (ISPW) dataset. Fig. ~\ref{fig:collage_supp} demonstrates that our ISPW dataset is captured in varying lighting and weather conditions. Thus making ISPW a very challenging dataset for training and benchmarking ISP pipelines in the wild.

\begin{figure*}[h]
    \centering
    \begin{subfigure}[b]{0.20\textwidth}
        \includegraphics[width=\textwidth]{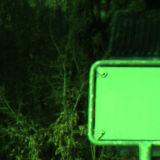}
        \includegraphics[width=\textwidth]{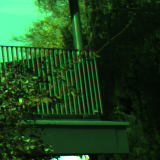}
        \caption{Phone RAW\\\hspace{2mm}}
        \label{fig:raw_vis}
    \end{subfigure}
    \begin{subfigure}[b]{0.20\textwidth}
        \includegraphics[width=\textwidth]{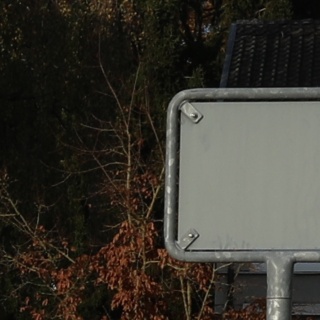}
        \includegraphics[width=\textwidth]{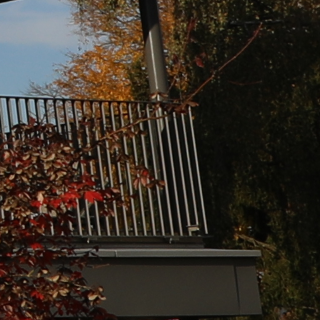}
        \caption{DSLR sRGB:\\EV = -1}
        \label{fig:tiger}
    \end{subfigure}
    \begin{subfigure}[b]{0.20\textwidth}
        \includegraphics[width=\textwidth]{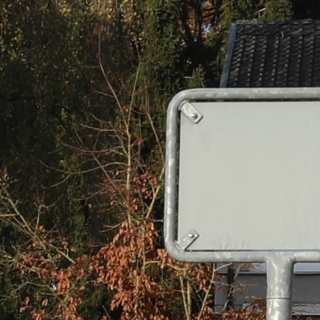}
        \includegraphics[width=\textwidth]{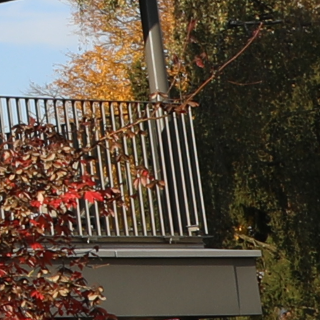}
        \caption{DSLR sRGB:\\EV = 0}
        \label{fig:ev2}
    \end{subfigure}
    \begin{subfigure}[b]{0.20\textwidth}
        \includegraphics[width=\textwidth]{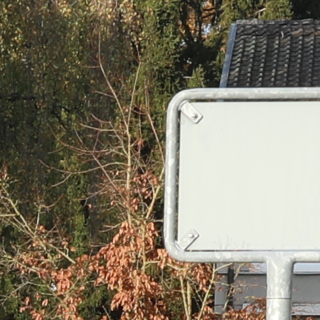}
        \includegraphics[width=\textwidth]{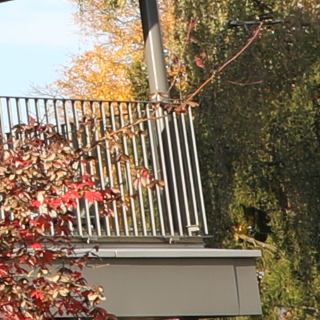}
        \caption{DSLR sRGB:\\EV = 1}
        \label{fig:ev3}
    \end{subfigure}
    \caption{Example crops from our ISPW dataset. We collect DSLR sRGB's at three different exposure settings. Note that we use the DSLR sRGB at EV setting of 0 for training our Color conditional DSLR sRGB restoration network.}\label{fig:ourdata}
\end{figure*}

Further, we provide a few example crops from our ISPW dataset after data processing (Sec. ~\ref{sec:dataset} of the manuscript). We capture the DSLR sRGB at 3 different exposures for the same phone RAW (Fig.~\ref{fig:ourdata}). We consider the DSLR sRGB captured with an EV setting 0 as the target for our RAW-to-sRGB mapping in the wild. Apart from providing various additional metadata that can further aid RAW-to-sRGB mapping in the wild research, we also provide the DSLR sRGB at 2 additional exposure settings which can be further used by the community for research directions such as automatic exposure correction~\cite{exposure_afifi} and various other avenues.

\section{Visual results for various components in our ISP pipeline}
\label{sec:intermediate_results_supp}

\begin{figure*}[h]
\newcommand{\wid}{0.136\linewidth}
    \centering
    \begin{subfigure}[b]{\wid}
        \includegraphics[width=\textwidth]{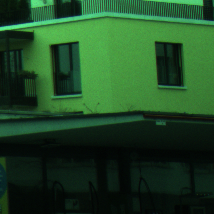}
        \includegraphics[width=\textwidth]{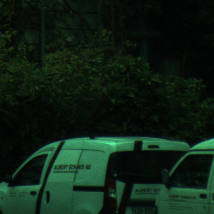}
        \includegraphics[width=\textwidth]{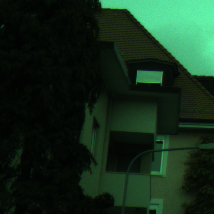}
        \includegraphics[width=\textwidth]{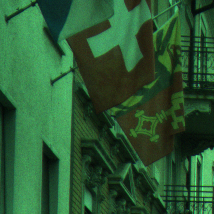}
        \includegraphics[width=\textwidth]{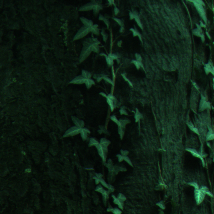}
        \includegraphics[width=\textwidth]{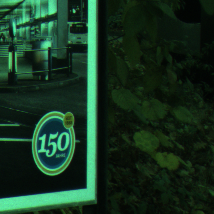}
        \caption{$x$}
        % \label{fig:res_mwnet2}
    \end{subfigure}
    \begin{subfigure}[b]{\wid}
        \includegraphics[width=\textwidth]{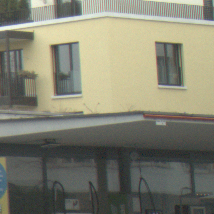}
        \includegraphics[width=\textwidth]{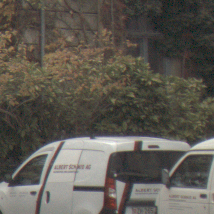}
        \includegraphics[width=\textwidth]{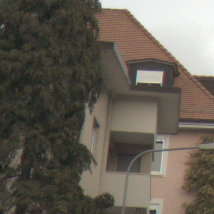}
        \includegraphics[width=\textwidth]{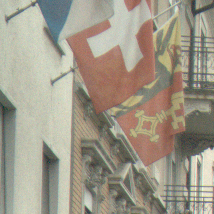}
        \includegraphics[width=\textwidth]{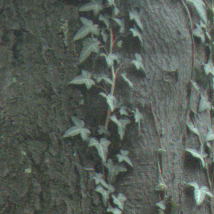}
        \includegraphics[width=\textwidth]{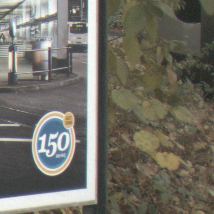}
        \caption{$x'$}
        % \label{fig:res_mwnet2}
    \end{subfigure}
    \begin{subfigure}[b]{\wid}
        \includegraphics[width=\textwidth]{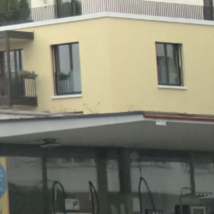}
        \includegraphics[width=\textwidth]{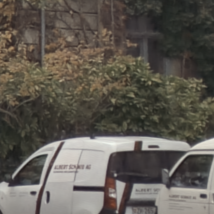}
        \includegraphics[width=\textwidth]{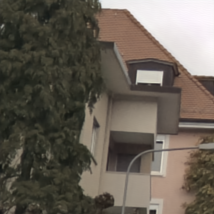}
        \includegraphics[width=\textwidth]{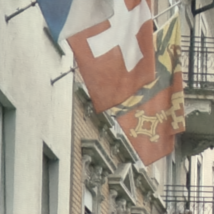}
        \includegraphics[width=\textwidth]{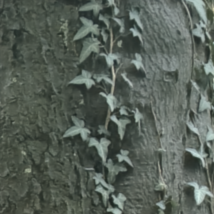}
        \includegraphics[width=\textwidth]{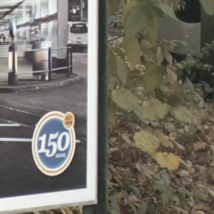}
        \caption{$\Tilde{x}$}
        % \label{fig:res_mwnet2}
    \end{subfigure}
    \begin{subfigure}[b]{\wid}
        \includegraphics[width=\textwidth]{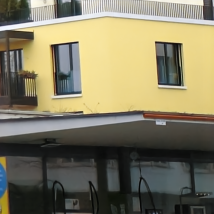}
        \includegraphics[width=\textwidth]{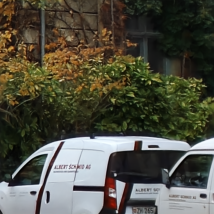}
        \includegraphics[width=\textwidth]{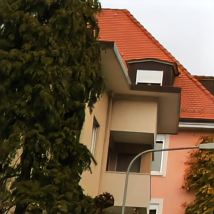}
        \includegraphics[width=\textwidth]{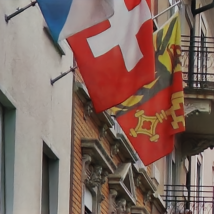}
        \includegraphics[width=\textwidth]{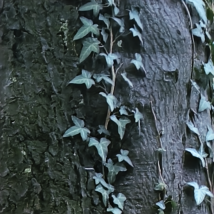}
        \includegraphics[width=\textwidth]{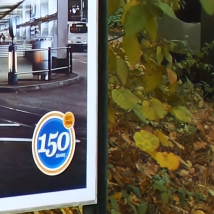}
        \caption{$c$}
        % \label{fig:res_mwnet2}
    \end{subfigure}
    \begin{subfigure}[b]{\wid}
        \includegraphics[width=\textwidth]{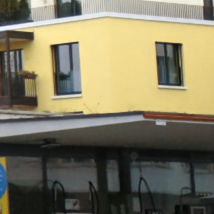}
        \includegraphics[width=\textwidth]{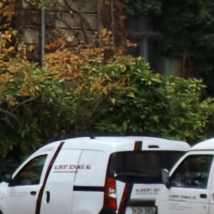}
        \includegraphics[width=\textwidth]{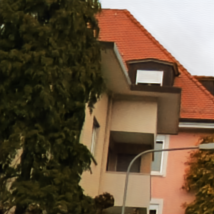}
        \includegraphics[width=\textwidth]{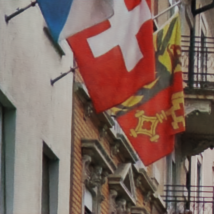}
        \includegraphics[width=\textwidth]{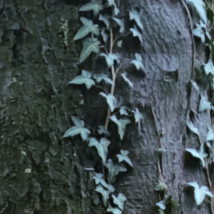}
        \includegraphics[width=\textwidth]{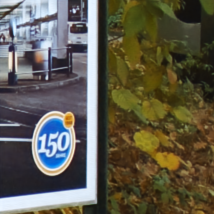}
        \caption{$\hat{c}$}
        % \label{fig:res_mwnet2}
    \end{subfigure}
    \begin{subfigure}[b]{\wid}
        \includegraphics[width=\textwidth]{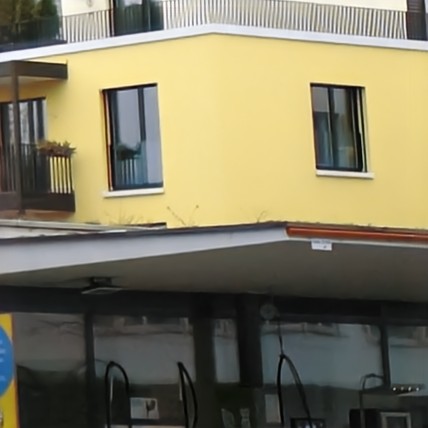}
        \includegraphics[width=\textwidth]{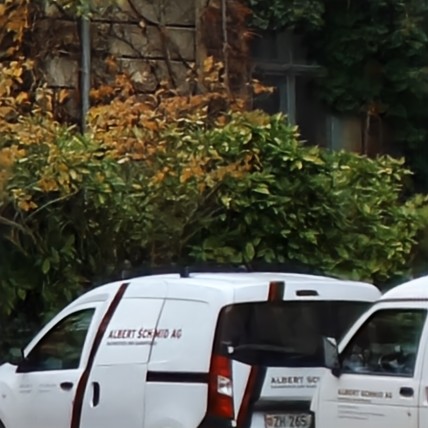}
        \includegraphics[width=\textwidth]{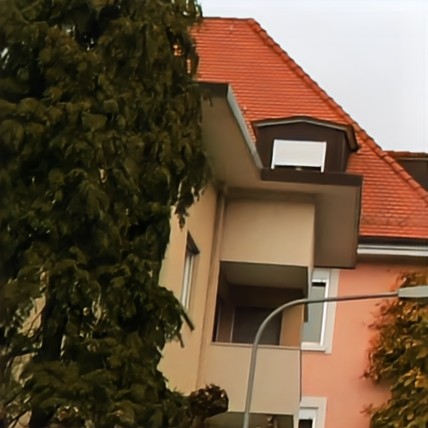}
        \includegraphics[width=\textwidth]{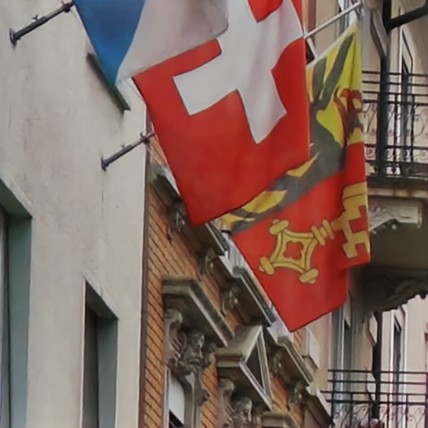}
        \includegraphics[width=\textwidth]{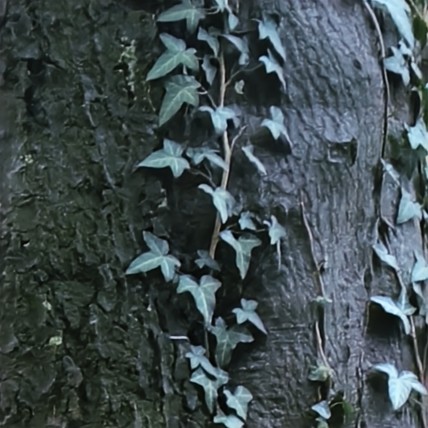}
        \includegraphics[width=\textwidth]{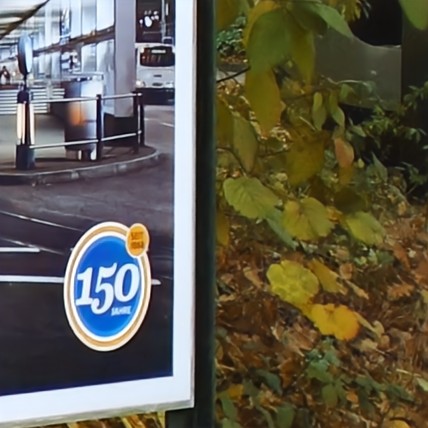}
        \caption{$\hat{y}$}
        % \label{fig:res_mwnet2}
    \end{subfigure}
    \begin{subfigure}[b]{\wid}
        \includegraphics[width=\textwidth]{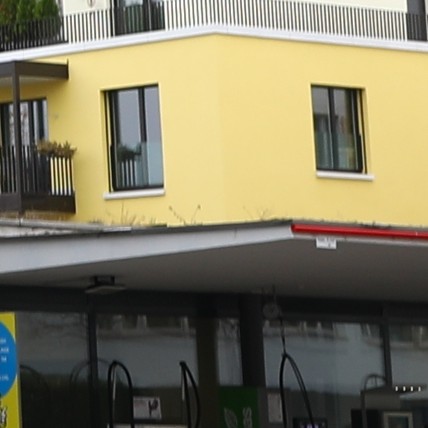}
        \includegraphics[width=\textwidth]{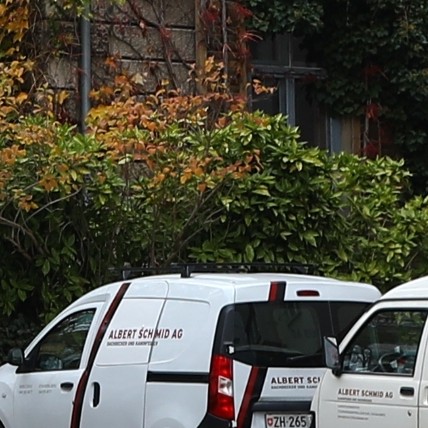}
        \includegraphics[width=\textwidth]{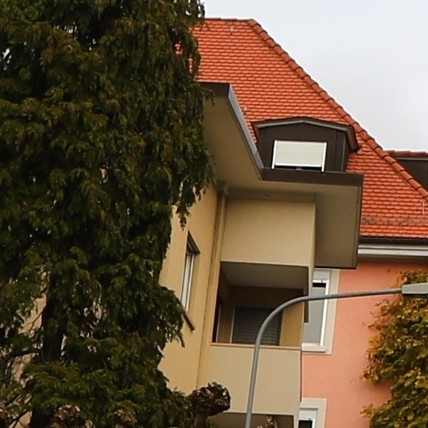}
        \includegraphics[width=\textwidth]{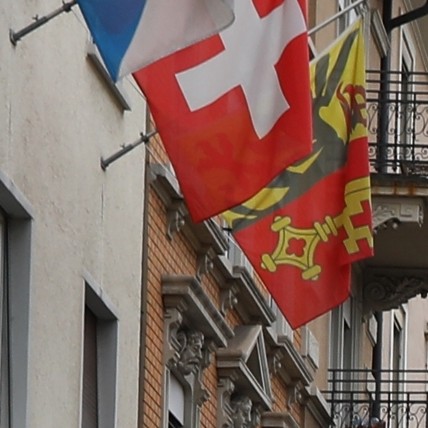}
        \includegraphics[width=\textwidth]{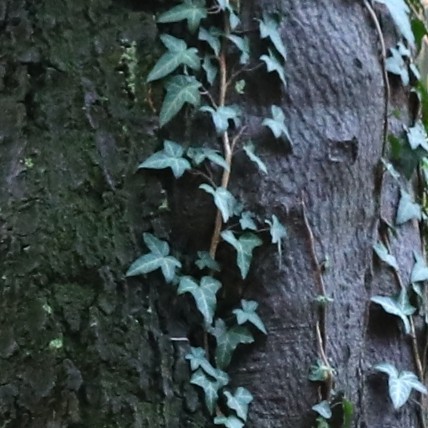}
        \includegraphics[width=\textwidth]{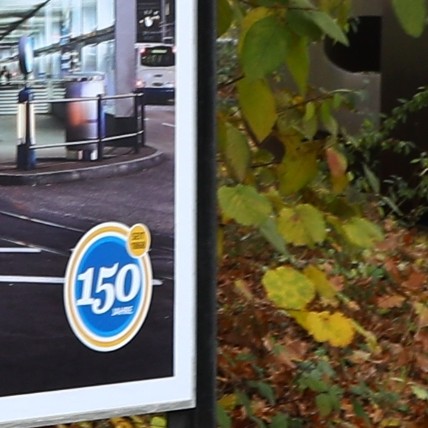}
        \caption{$y$}
        % \label{fig:intermediate_supp}
    \end{subfigure}
    \caption{We show the intermediate predictions in our framework for a few examples in the ZRR dataset. In the figure, $x$ is the visualized RAW from the phone and $x'=\Gamma(x)$ (Eq.~\ref{eq:process}). The output of the Pre-processing network (Sec. ~\ref{sec:color_map} of the manuscript) $\Tilde{x}$ is shown in column 3. Further, $c=\mathcal{G}(x)$ is the predicted low-resolution color image by our color prediction network (Sec. ~\ref{sec:colorinference} of the manuscript) that integrates a global context transformer to integrate global cues for predicting accurate colors. The pre-processed RAW $\Tilde{x}$ is then color mapped to $c$ using our parametric color mapping formulation (Sec. ~\ref{sec:color_map} of the manuscript). The color mapped image $\hat{c}=\mathcal{C}(\Tilde{x}, c)$. During inference the parametric color mapping $\mathcal{C}$ aids in smoothing out the spurious color predictions that may occur in $c$. Finally, our ISP network predicts the final DSLR quality $\hat{y}=\mathcal{F}(x, \hat{c})$. The last column shows the DSLR sRGB ($y$) crop. Best viewed with zoom.}\label{fig:intermediate_supp}
\end{figure*}

In this section, we show the visual results for different components in our RAW-to-sRGB mapping in the wild pipeline. Figure ~\ref{fig:intermediate_supp} shows the intermediate results for our ISP Network. We show that $x'=\Gamma(x)$ (Eq.~\ref{eq:process}) provides our pipeline with a rough visualization for the phone RAW $x$. This processed RAW $x'$ aids in creating a mask for regions where alignment is difficult leading to a more accurate training supervision. We also see that, our Global-Context transformer based color predictor predicts a color image $c=\mathcal{G}(x)$ that is consistent with the colors in the target DSLR sRGB $y$. Our flexible parametric color mapping scheme is powerful enough to color-map the pre-processed RAW $\Tilde{x}$ to the predicted color image $c=\mathcal{G}(x)$ very accurately with just 15 bins. Finally, our RAW-to-sRGB restoration network predicts the DSLR quality sRGB image $y=\mathcal{F}(x, \hat{c})$.   

\section{Additional Experiments}
\label{sec:additional_exp}
\begin{figure*}[t]
\captionsetup[subfigure]{labelformat=empty}
\newcommand{\wid}{0.18\linewidth}
\centering
\begin{subfigure}[b]{\wid}
        \includegraphics[width=\textwidth]{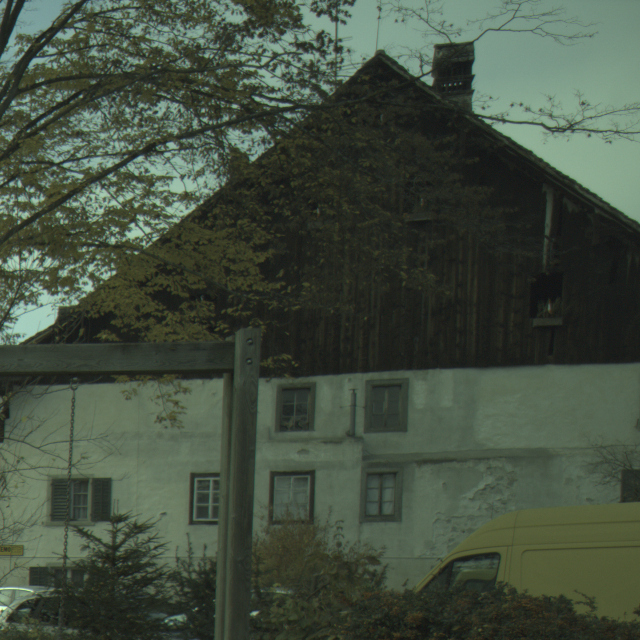}
        \caption{1a) RAW}
        % \label{fig:res_mwnet2}
    \end{subfigure}
    \begin{subfigure}[b]{\wid}
        \includegraphics[width=\textwidth]{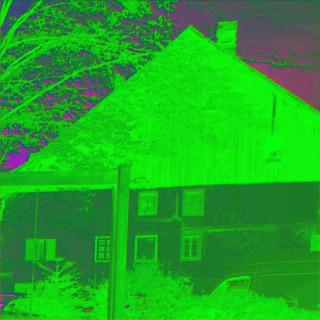}
        \caption{1b) encode-l1}
        % \label{fig:res_mwnet2}
    \end{subfigure}
    \begin{subfigure}[b]{\wid}
        \includegraphics[width=\textwidth]{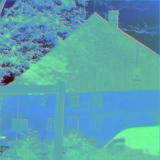}
        \caption{1c) encode-l2}
        % \label{fig:res_mwnet2}
    \end{subfigure}
    \begin{subfigure}[b]{\wid}
        \includegraphics[width=\textwidth]{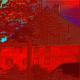}
        \caption{1d) encode-l3}
        % \label{fig:res_mwnet2}
    \end{subfigure}
    \begin{subfigure}[b]{\wid}
        \includegraphics[width=\textwidth]{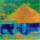}
        \caption{1e) encode-l4}
        % \label{fig:res_mwnet2}
    \end{subfigure}
    
    \begin{subfigure}[b]{\wid}
        \includegraphics[width=\textwidth]{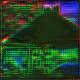}
        \caption{1f) decode-l1}
        % \label{fig:res_mwnet2}
    \end{subfigure}
    \begin{subfigure}[b]{\wid}
        \includegraphics[width=\textwidth]{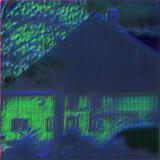}
        \caption{1g) decode-l2}
        % \label{fig:res_mwnet2}
    \end{subfigure}
    \begin{subfigure}[b]{\wid}
        \includegraphics[width=\textwidth]{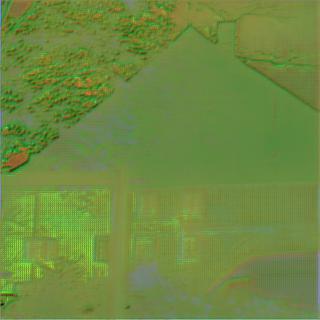}
        \caption{1h) decode-l3}
        % \label{fig:res_mwnet2}
    \end{subfigure}
    \begin{subfigure}[b]{\wid}
        \includegraphics[width=\textwidth]{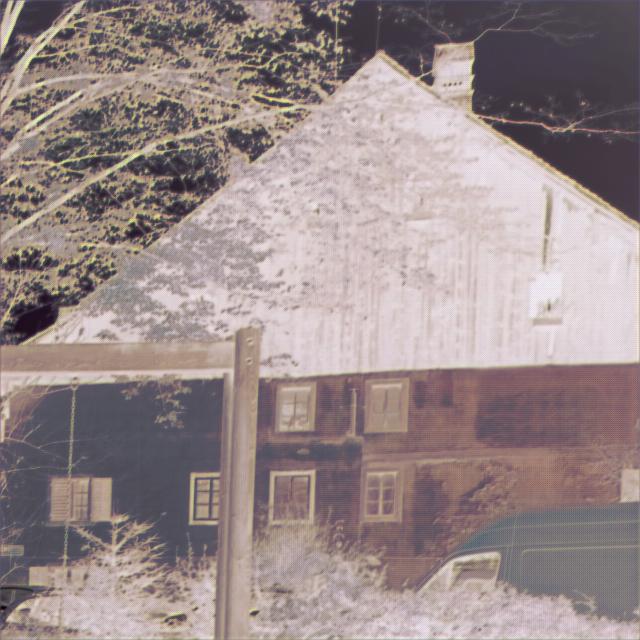}
        \caption{1i) decode-l4}
        % \label{fig:res_mwnet2}
    \end{subfigure}
    \begin{subfigure}[b]{\wid}
        \includegraphics[width=\textwidth]{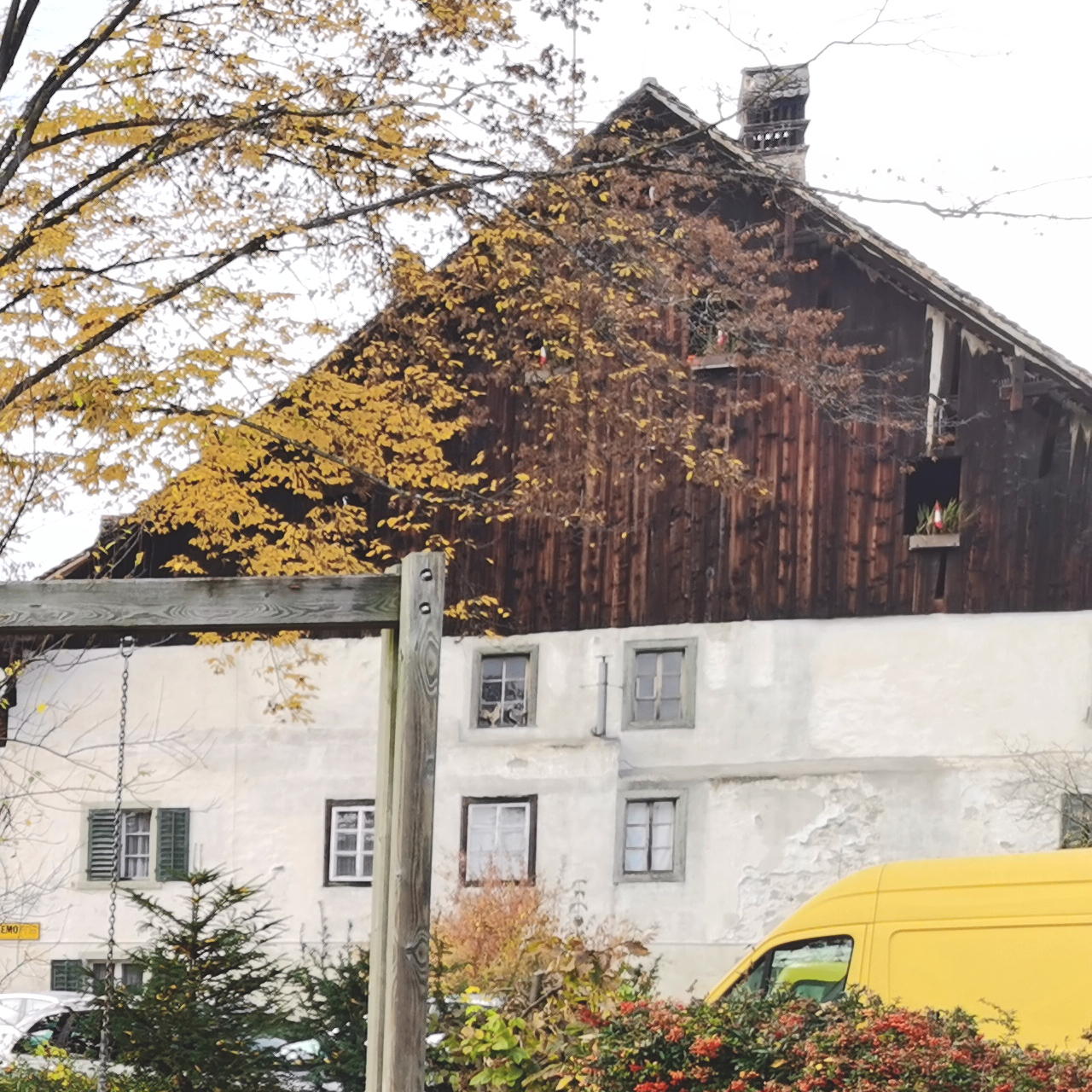}
        \caption{1j) $c$}
        % \label{fig:res_mwnet2}
    \end{subfigure}
    \caption{We show the visualized (by taking the first 3 channels) resulting feature maps at each U-Net level (both encoder and the DSLR decoder) for an example crop from our ISPW dataset. Here, encode-ln signifies the feature map output from our encoder block at level n. Similarly, decode-ln is the feature map output from our encoder block at level n. Best viewed with zoom.}\label{fig:feature_supp}
\end{figure*}

\parsection{Feature maps from our Color-Prediction Net} Figure~\ref{fig:feature_supp} shows the feature maps from different encoder decoder levels in our U-Net color predictor network $\mathcal{G}$. The network captures detailed image information at different levels.

\begin{figure*}[h]
\newcommand{\wid}{0.26\linewidth}
    \centering
    \begin{subfigure}[b]{\wid}
        \includegraphics[width=\textwidth]{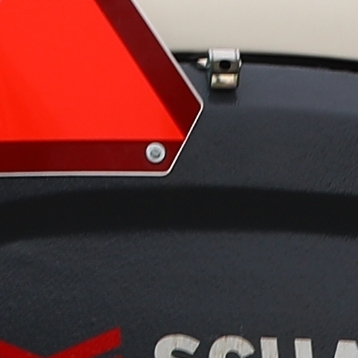}
        \includegraphics[width=\textwidth]{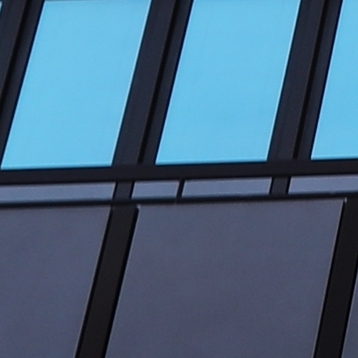}
        \caption{DSLR sRGB (ZRR dataset)}
        % \label{fig:res_mwnet2}
    \end{subfigure}
    \begin{subfigure}[b]{\wid}
        \includegraphics[width=\textwidth]{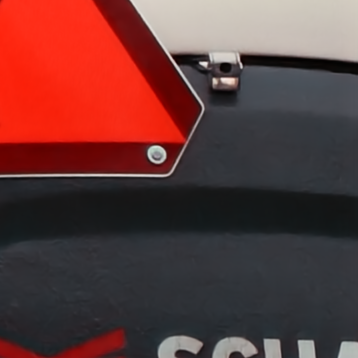}
        \includegraphics[width=\textwidth]{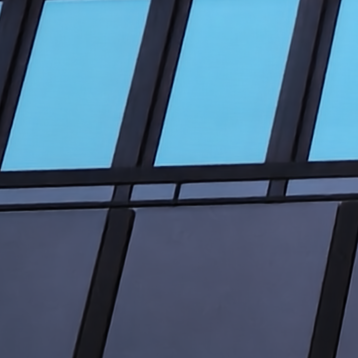}
        \caption{Ours-ZRR\\\hspace{2mm}}
        % \label{fig:res_mwnet2}
    \end{subfigure}
    \begin{subfigure}[b]{\wid}
        \includegraphics[width=\textwidth]{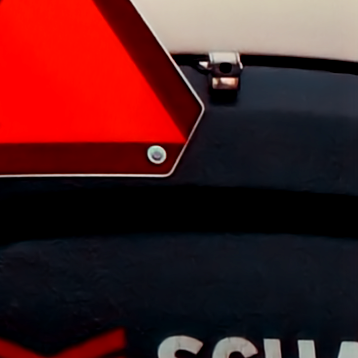}
        \includegraphics[width=\textwidth]{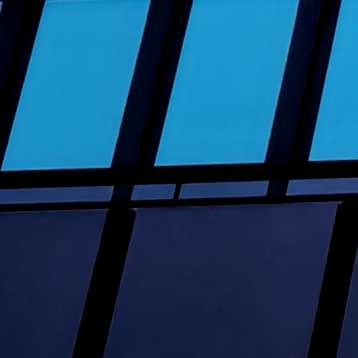}
        \caption{Ours-ISPW\\\hspace{2mm}}
        % \label{fig:res_mwnet2}
    \end{subfigure}
    \caption{Testing our model trained on the ISPW dataset on two example crops from the ZRR dataset. Ours-ISPW shows the results for the model trained on our ISPW dataset. Ours-ZRR is the result of the model trained on the ZRR dataset.  produces}\label{fig:crossval_zrr_supp}
\end{figure*}

\begin{figure*}[t!]
\newcommand{\wid}{0.26\linewidth}
    \centering
    \begin{subfigure}[b]{\wid}
        \includegraphics[width=\textwidth]{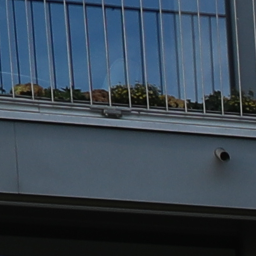}
        \includegraphics[width=\textwidth]{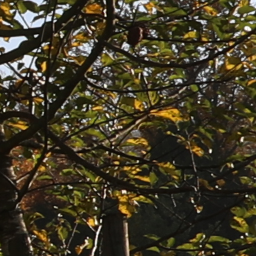}
        \caption{DSLR sRGB\\ (ISPW dataset)}
        % \label{fig:res_mwnet2}
    \end{subfigure}
    \begin{subfigure}[b]{\wid}
        \includegraphics[width=\textwidth]{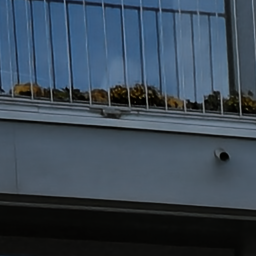}
        \includegraphics[width=\textwidth]{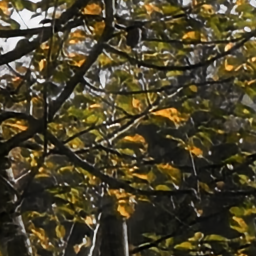}
        \caption{Ours-ISPW\\\hspace{2mm}}
        % \label{fig:res_mwnet2}
    \end{subfigure}
    \begin{subfigure}[b]{\wid}
        \includegraphics[width=\textwidth]{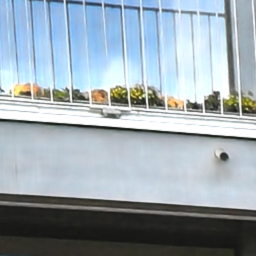}
        \includegraphics[width=\textwidth]{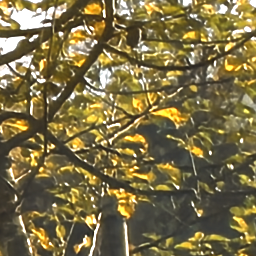}
        \caption{Ours-ZRR\\\hspace{2mm}}
        % \label{fig:res_mwnet2}
    \end{subfigure}
    \caption{Testing our model trained on the ZRR dataset on two example crops from the ISPW dataset. Ours-ISPW shows the results for the model trained on our ISPW dataset. Ours-ZRR is the result of the model trained on the ZRR dataset.  }\label{fig:crossval_ispw_supp}
\end{figure*}

\parsection{Cross-dataset experiment} Next, to check how our models perform on datasets they are not trained on. We do inference on the ISPW dataset using the model trained on the ZRR dataset and vice versa. Figures~\ref{fig:crossval_zrr_supp} and~\ref{fig:crossval_ispw_supp} show the visual results on example crops from both the datasets. It is evident from the qualitative results that our framework is able to produce feasable DSLR quality sRGB's even when it is run on a dataset it is not trained on.

\clearpage
\bibliographystyle{splncs04}
%\bibliography{egbib}
\bibliography{main}
\end{document}